\documentclass[12pt]{article}

\usepackage{array,arydshln}
\usepackage{algorithm}
\usepackage{algorithmicx}
\usepackage{amsmath}
\usepackage{amssymb}
\usepackage{caption}
\usepackage{xcolor}
\usepackage{enumerate}
\usepackage{float}
\usepackage{graphicx}
\usepackage{hhline}
\usepackage{hyperref}
\usepackage[utf8]{inputenc}
\usepackage{makecell}
\usepackage{multirow}
\usepackage{natbib}
\usepackage{siunitx} 
\usepackage{verbatim} 

\def\aln{\begin{align*}}
\def\eln{\end{align*}}
\def\be{\begin{equation}}
\def\ee{\end{equation}}
\def\bse{\begin{eqnarray*}}
\def\ese{\end{eqnarray*}}
\def\bea{\begin{eqnarray}}
\def\eea{\end{eqnarray}}

\newtheorem{proposition}{Proposition}[section]

\begin{document}

\thispagestyle{empty}

\title{\Large  Likelihood-guided Regularization in Attention Based Models}
\author{Mohamed Salem$^1$ and Inyoung Kim$^{1*}$}
\date{}
\maketitle

\noindent 1 Department of Statistics, Virginia Polytechnic Institute and State University, Blacksburg, VA 24061, USA.

\vspace{1in}
\noindent $\ast$ To whom correspondence should be addressed:\\
Inyoung Kim, Ph.D.\\
Department of Statistics, Virginia Tech., Blacksburg, VA 24061, USA.\\
Tel: (540) 231-5366\\
Fax: (540) 231-3863\\
E-Mail: \texttt{inyoungk{@}vt.edu}\\

\newpage
\begin{center}
\textbf{Abstract}
\end{center}

The transformer architecture has demonstrated strong performance in classification tasks involving structured and high-dimensional data. However, its success often hinges on large-scale training data and careful regularization to prevent overfitting. In this paper, we introduce a novel likelihood-guided variational Ising-based regularization framework for Vision Transformers (ViTs), which simultaneously enhances model generalization and dynamically prunes redundant parameters. The proposed variational Ising-based regularization approach leverages Bayesian sparsification techniques to impose structured sparsity on model weights, allowing for adaptive architecture search during training. Unlike traditional dropout-based methods, which enforce fixed sparsity patterns, the variational Ising-based regularization method learns task-adaptive regularization, improving both efficiency and interpretability. We evaluate our approach on benchmark vision datasets, including MNIST, Fashion-MNIST, CIFAR-10, and CIFAR-100, demonstrating improved generalization under sparse, complex data and allowing for principled uncertainty quantification on both weights and selection parameters. Additionally, we show that the Ising regularizer leads to better-calibrated probability estimates and structured feature selection through uncertainty-aware attention mechanisms. Our results highlight the effectiveness of structured Bayesian sparsification in enhancing transformer-based architectures, offering a principled alternative to standard regularization techniques.

\vspace*{.3in}


\newpage
\section{Introduction}\label{int}

The success of deep learning in various domains has driven the development of increasingly complex architectures, often at the expense of interpretability and efficiency. While deep neural networks excel at feature extraction, they frequently suffer from overparameterization, making them prone to poor generalization and inefficient computation. Regularization techniques are essential to mitigate these issues by enforcing sparsity, improving robustness, and dynamically adapting model complexity to task requirements.

Transformers, initially introduced by \cite{attntn}, have emerged as a powerful alternative to convolutional neural networks (CNNs) in computer vision. Vision Transformers (ViTs) \cite{vit} leverage self-attention mechanisms to model long-range dependencies, allowing them to capture spatial relationships in images more effectively than traditional CNNs. However, ViTs typically require large-scale datasets and strong regularization to prevent overfitting, highlighting the need for structured sparsification techniques.

\subsection{Likelihood-guided Regularization Approach}
In this paper, we introduce a novel likelihood-guided variational Ising-based regularization framework designed to simultaneously regularize model complexity and dynamically prune redundant parameters. Unlike conventional dropout-based methods \cite{zoubingal}, which enforce fixed sparsity patterns, our approach learns structured sparsity patterns that adapt to the data, enhancing both model efficiency and interpretability.

Building on prior work in Bayesian sparsification \cite{Graves2011, Titsias2014} and Ising-based pruning \cite{Salehinejad2019IsingdropoutAR, Salehinejad2021PruningOC}, our method imposes an adaptive, structured prior on model weights. Existing spike-and-slab priors \cite{McCulloch1993} often assume fixed mixture parameters, limiting flexibility. In contrast, our variational formulation introduces learnable sparsity patterns, optimizing both the model architecture and its predictive uncertainty.

By integrating variational Ising regularization within a transformer-based architecture, we provide a principled approach for uncertainty-aware structured pruning that maintains high model performance while reducing computational complexity. This framework offers advantages in uncertainty calibration, structured feature selection, and interpretability.

\subsection{Main Contributions}
Our key contributions are as follows:
\begin{enumerate}[(i)]
    \item We propose a novel variational Ising-based regularization approach that simultaneously performs structured pruning and model selection.
    \item Our method extends uncertainty quantification to attention structures, enabling uncertainty-aware feature selection and enhancing interpretability.
    \item We develop an efficient implementation that scales linearly with the number of model parameters, integrating seamlessly with standard transformer training routines.
\end{enumerate}

\subsection{Outline of the Paper}
This paper is organized as follows: In Section \ref{L2}, we describe both the typical variational inference approach in attention-based models and our proposed method. In Section \ref{alg}, we provide our variational Ising-based algorithm.  Section \ref{stat_props} presents the statistical properties of our approach. Section \ref{exp_dat} evaluates our method on multiple benchmark datasets, comparing it against alternative sparsification techniques. 
Finally, Section \ref{conc} concludes with a summary of findings and potential future directions.

\section{Likelihood-guided Regularization in Attention-Based Models}\label{L2}


In this section, following \cite{sankararaman2022bayesformer},  we show how to implement Bayesian inference using our proposed new regularizer in  Attention-based models. 

\subsection{Classification model}
Let $\mathcal{D} = \{y_{n},\mathbf{X}_{n}\}_{n=1}^{N}$ be a finite dataset, where $N$ is the number of observations. Let $y_{n}$ be a binary response variable indicating the class label associated with observation $n$, where $\mathbf{X}_{n}$ represents the functional covariate (i.e. image) associated with observation $n$, and $\mathbf{X} = \big(\mathbf{X}_{1},\dots, \mathbf{X}_{N}\big{)}^{T}$ is our full covariate tensor. Then, an attention-based model can be expressed as
\begin{equation*}
\begin{aligned}
Pr(y_n=1| \mathbf{X}, \mathbf{W}, \boldsymbol{\xi})=f_{\mathbf{y,W, \boldsymbol{\xi}}}(\mathbf{X})
\end{aligned} 
\end{equation*}
where $f_{\mathbf{y,W, \boldsymbol{\xi}}}(\mathbf{X})$ is the output of an attention based model on a finite dataset $\mathcal{D} = \{y_{n},\mathbf{X}_{n}\}_{n=1}^{N}$, $\mathbf{W}$ represents the weights of all layers in our model, and finally $\boldsymbol{\xi}$ represents a matrix of binary dropout variables acting as a dropout mask for weights in the model. The dropout scheme proposed here is weight-specific, in the essence of DropConnect \cite{dropconnect}, rather than the node-based dropout of \cite{sankararaman2022bayesformer}. 

\subsection{Variational Bayes for Weights}
Adopting a full Bayesian approach for our neural network model would involve placing a prior distribution, $p(\cdot)$, on weights ($W$) of all linear transformations in the model such that:
\begin{equation*}
\begin{aligned}
\mathbf{W} \sim p(\mathbf{W}).
\end{aligned} 
\end{equation*}
Typically, a $\mathcal{N}(\mathbf{W}; 0, \sigma^{2})$ prior is placed on the weights with a well chosen $\sigma^{2}$. \cite{blundell2015weight} propose a singular-spike slab prior for all $\mathbf{W}$ as,
\begin{equation*}
\begin{aligned}
p(\mathbf{W}) = \prod \pi\mathcal{N}(\mathbf{W}; 0, \sigma^{2}_{1}) + (1-\pi)\mathcal{N}(\mathbf{W}; 0, \sigma^{2}_{2}),
\end{aligned} 
\end{equation*}
where $\pi$ is a mixing probability of two normal distributions and $\sigma^{2}_{1} < \sigma^{2}_{2}$ causing some weights to cluster around zero. 
%
The formulation in which the prior shares its hyperparameters across all weights offers greater computational efficiency for optimization. However, this contrasts with the method of assigning hierarchical priors and learning separate hyperparameters for each weight, which, according to \cite{blundell2015weight}, tends to be ineffective in practice.

While we propose a weight-specific hyperparameter, we will show how to efficiently perform the computation using stochastic gradient descent with minimal computational cost. Our proposed prior is
\begin{equation*}
\begin{aligned}
p(\mathbf{W}| \boldsymbol{\xi}) &= \prod_{j,j'} \xi_{j,j'}\mathcal{N}(\mathbf{W}; 0, \sigma^{2}_{1}) + (1-\xi_{j,j'})\mathcal{N}(\mathbf{W}; 0, \sigma^{2}_{2}), \quad\quad \xi_{j}\;\in\;\{0,1\} \quad \forall\;j\\
\boldsymbol{\xi} &\sim p(\boldsymbol{\xi}),
\end{aligned} 
\end{equation*}
where $p(\boldsymbol{\xi})$ is a prior distribution.
We are therefore interested in obtaining a posterior on the joint distribution $p(\mathbf{W}, \boldsymbol{\xi}|\mathbf{X, y}) = p(\mathbf{W|\boldsymbol{\xi},X,y})\cdot p(\mathbf{\boldsymbol{\xi}|X,y})$. Recovery of  $p(\mathbf{W|\boldsymbol{\xi},X,y})$ allows for quantifying both uncertainty about the weight estimates as well as uncertainty about the prediction through the posterior predictive distribution as,
\begin{equation*}
\begin{aligned}
p(\mathbf{y|X}) = \sum_{\boldsymbol{\xi}}\int  p(\mathbf{y|X,W, \boldsymbol{\xi}})p(\mathbf{W|\boldsymbol{\xi},X,y})p(\mathbf{\boldsymbol{\xi}|X,y})\;d\mathbf{W}.
\end{aligned} 
\end{equation*}
However, given the dimensionality of $\mathbf{W}$ - which is a function of the number and size of the model layers - as well as the discrete nature of $\boldsymbol{\xi}$ this becomes an intractable task, particularly in a high-dimensional setting. 

As such, we resort to an approximation of this posterior via Variational Inference (VI) whereby we approximate $p(\mathbf{W, \boldsymbol{\xi}|X,y})$ using some distribution $q(\mathbf{W},\boldsymbol{\xi})$ that is easier to compute. We then attempt to minimize the KL divergence between our proposed approximation and the true posterior distribution. To formulate our objective, we  consider the function $L(q)$,
\begin{equation*}
\begin{aligned}
L(q) = \log p(\mathbf{y|\boldsymbol{\xi},X,W}) - \mathbb{KL}(q(\mathbf{W},\boldsymbol{\xi})||p(\mathbf{W,\boldsymbol{\xi}|X,y})) \leq \log p(\mathbf{y|\boldsymbol{\xi},X,W}),
\end{aligned} 
\end{equation*}
where $\mathbb{KL}$ is the Kullback-Leibler divergence; which being a non-negative quantity, implies the inequality. Maximizing this lower bound on the marginal log-likelihood thereby indirectly maximizes the marginal log-likelihood of the data, and as our function of choice $q$ only appears in the KL-divergence term, this is equivalent to minimizing KL-divergence, which, in turn, is equivalent to the following optimization objective, 

\begin{equation*}
\begin{aligned}
\min_{q} - \sum_{\boldsymbol{\xi}} q(\boldsymbol{\xi})\bigg{(}\int  q(\mathbf{W}| \boldsymbol{\xi})\cdot\log [p(\mathbf{y|X,W, \boldsymbol{\xi}})]\;d\mathbf{W} + \mathbb{KL}(q(\mathbf{W}|\boldsymbol{\xi})||p(\mathbf{W}|\boldsymbol{\xi}))\bigg{)} + \mathbb{KL}(q(\boldsymbol{\xi})||p(\boldsymbol{\xi})).
\end{aligned} 
\end{equation*}
The integral in this expression computes the expectation of $\log p(\mathbf{y_{n}|X_{n},W, \boldsymbol{\xi}})$ over all $\mathbf{W}$ assuming $\mathbf{W}\sim q(\mathbf{W}|\boldsymbol{\xi})$. Therefore, we can obtain an unbiased estimate for the integral using Monte Carlo integration \citep{zoubingal} such that,

\begin{equation*}
\begin{aligned}
\int  q(\mathbf{W}| \boldsymbol{\xi})\cdot\log [p(\mathbf{y_{n}|X_{n},W, \boldsymbol{\xi}})]\;d\mathbf{W} \approx p(\mathbf{y_{n}|X_{n},\widehat{W}, \boldsymbol{\xi}}).
\end{aligned} 
\end{equation*}
Employing our sampling of $\widehat{\mathbf{W}}$ within our stochastic gradient descent leads to the following minimization over this quantity across our observed dataset, which leads to the following form for empirical risk minimization:
\begin{equation*}
\begin{aligned}
\min_{q} - \sum_{n}\sum_{\boldsymbol{\xi}} q(\boldsymbol{\xi})\bigg{(}\log [p(\mathbf{y_{n}|X_{n},\widehat{W}, \boldsymbol{\xi}})] + \mathbb{KL}(q(\mathbf{W}|\boldsymbol{\xi})||p(\mathbf{W}|\boldsymbol{\xi}))\bigg{)} + \mathbb{KL}(q(\boldsymbol{\xi})||p(\boldsymbol{\xi})).
\end{aligned} 
\end{equation*}
We then employ a similar Monte Carlo approximation for $\boldsymbol{\xi}$ which has the same dimensionality as $\mathbf{W}$,
\begin{equation*}
\begin{aligned}
\sum_{\boldsymbol{\xi}} q(\boldsymbol{\xi})\bigg{(}\log [p(\mathbf{y_{n}|X_{n},\widehat{W}, \boldsymbol{\xi}})] + \mathbb{KL}(q(\mathbf{W}|\boldsymbol{\xi})||p(\mathbf{W}|\boldsymbol{\xi}))\bigg{)} \approx \bigg{(}p(\mathbf{y_{n}|X_{n},\widehat{W}, \boldsymbol{\hat \xi}}) + \mathbb{KL}(q(\mathbf{W}|\boldsymbol{\hat \xi})||p(\mathbf{W}|\boldsymbol{\xi}))\bigg{)}.
\end{aligned} 
\end{equation*}
Therefore, our objective function is
\begin{equation}\label{ourobj}
\begin{aligned}
\min_{q} - \sum_{n} \log [p(\mathbf{y_{n}|X_{n},\widehat{W}, \boldsymbol{\hat \xi}})] + \mathbb{KL}(q(\mathbf{W}|\boldsymbol{\hat \xi})||p(\mathbf{W}|\boldsymbol{\xi})) + \mathbb{KL}(q(\boldsymbol{\xi})||p(\boldsymbol{\xi})).
\end{aligned} 
\end{equation}
The above derivations (\ref{ourobj}) express our problem in the most general terms. For the binary classification case we can express it as the following,
\begin{equation*}
\begin{aligned}
\log [p(\mathbf{y_{n}|X_{n},\widehat{W}, \boldsymbol{\hat \xi}})] = \mathbf{y_{n}}\log\bigg{(}f_{\widehat{\mathbf{W}}, \hat{\boldsymbol{\xi}}}(\mathbf{X_{n}})\bigg{)} + (1-\mathbf{y_{n}})\log\bigg{(}1-(f_{\widehat{\mathbf{W}}, \hat{\boldsymbol{\xi}}}(\mathbf{X_{n}}))\bigg{)}.
\end{aligned} 
\end{equation*}

Next, for any weight matrix from within our architecture, denoted by $\mathbf{W}$, we posit the following variational distribution, $q_{\mathbf{M}}(\mathbf{W}|\boldsymbol{\xi})$, parametrized by a matrix $\mathbf{M}$ of hyperparameters, where $\mathbf{M} = \bigg{(}m_{1,1}, m_{1,2},\dots, m_{J^{(L-1)},J^{(L)}}\bigg{)}$,
%
\begin{equation}
\begin{aligned}
q_{\mathbf{M}}(\mathbf{W}|\boldsymbol{\xi}) = \prod_{j,j'} \xi_{j,j'}\mathcal{N}(\mathbf{W}; 0, \sigma^{2})+(1-\xi_{j,j'})\mathcal{N}(\mathbf{W}; m_{j,j'}, \sigma^{2}).
\end{aligned} 
\end{equation}
For our selection variables, we propose the following variational distribution, 
\begin{equation}
\begin{aligned}
q(\xi_{j', \; j}^{(l)}=1) = \bigg{[}1+\exp\bigg{\{}-2\frac{\sum_{j''\in\;(l+1)}w_{j'',j'}^{2}E_{q}[\xi_{j'',j'}^{(l+1)}]}{\sum_{j''\in\;(l+1)}w_{j'',j'}^{2}}-(L_{j}^{+}-L_{j}^{-})\bigg{\}}\bigg{]}^{-1},
\end{aligned} 
\end{equation}
where $\xi_{j',\; j}^{(l)}$ is the binary dropout variable for the weight connecting node $j$ in layer $l$ to node $j'$ in layer $l+1$ (before application of the chosen activation function), $L_{j}^{+}$ and $L_{j}^{-}$ denote \\
$\sum_{n} \log [p(\mathbf{y_{n}|X_{n},W, E_{q}[\boldsymbol{\xi}_{-(j',j)}]}, \xi_{j',j}^{(l)}=1)]$ and $\sum_{n} \log [p(\mathbf{y_{n}|X_{n},W, E_{q}[\boldsymbol{\xi}_{-(j',j)}]}, \xi_{j',j}^{(l)}=0)]$, respectively. $E_{q}[\xi_{j',j}^{(l+1)}]$ denotes the expectation over binary dropout variables in the following layer $(l+1)$, and $w_{j',j}$ denotes the weights originating at the output of the activation of the $j^{th}$ node from layer $l$ and going into the $j'$ node in layer $l+1$. 

The dependence on $\mathbf{W}$ may seem contradictory at first glance. However, there are no outbound weights at the last layer of the network, $L$, causing the first term to vanish. As such, for any layer $l$, $q(\xi_{j',j}^{(l)})$ only depends on $\mathbf{W}^{(l+1)}$ which has already been updated by backpropagation independently of $\xi_{j',j}^{(l)}$. We note that the Ising prior has been applied for neural network regularization before \citep{Salehinejad2019IsingdropoutAR, Salehinejad2021PruningOC}, but to our knowledge never in a backward fashion, as all previous applications followed a forward looking scheme based on previous node activation.

\begin{figure}[htbp]	
\centering
\includegraphics[scale=0.65]{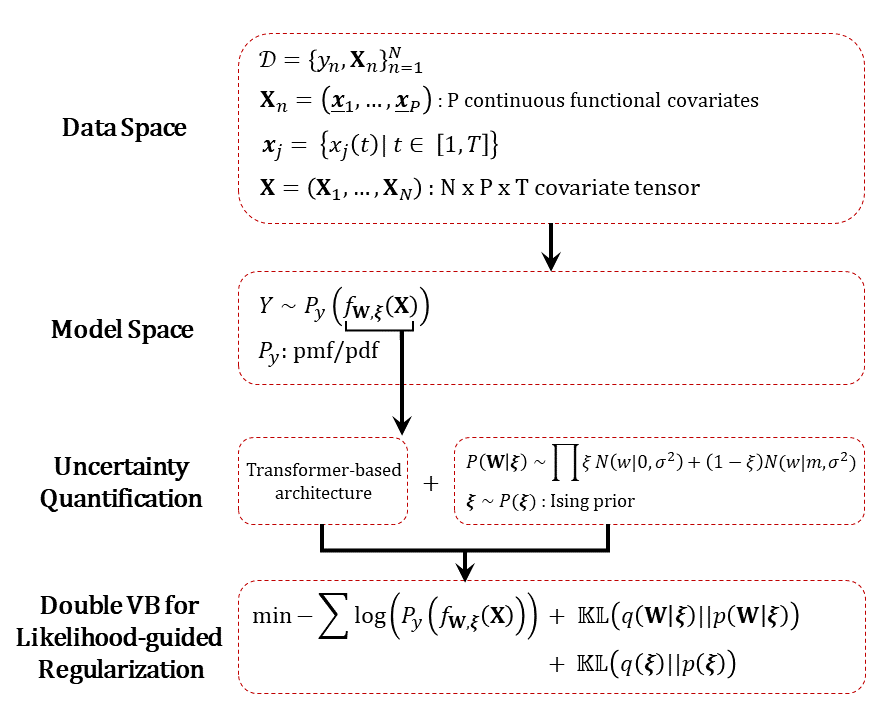}
	\caption{The figure shows the components in the likelihood-guided regularization method. The selection variables $\xi$ determine which distribution in the spike-slab prior a weight, $w$, is sampled from.} 
    \label{flow_diag}
\end{figure}

The estimation of this probability directly is computationally intractable, as it requires implementing a forward pass for each proposed value of $\xi_{j',j}$, leading to a total of $2^{|\boldsymbol{\xi}|}$ required evaluations per batch. To overcome this computational bottleneck, we leverage the approximation utilized in \cite{OBD} to compute saliency scores by using a Taylor expansion to approximate the log-likelihood around a previously obtained local optimum. This leads to the following approximation involving terms of the Hessian obtained from the proposed likelihood function,

\begin{equation}\label{taylor}
\begin{aligned}
L_{j}^{+}-L_{j}^{-} \approx \frac{\partial^{2}L(q)}{\partial w_{j',j}^{2}} = \frac{\partial^{2}L(q)}{\partial a_{j'}^{2}}x_{j}^{2}.
\end{aligned} 
\end{equation}
This can be made even more computationally efficient using the Levenberg-Marquardt approximation to the Hessian, which requires only the entries of the gradient, where for all layers $l \neq L$,

\begin{equation}\label{backprop_taylor}
\begin{aligned}
\frac{\partial^{2}L(q)}{\partial a_{j'}^{2}} = f'(a_{j'})^{2}\sum_{j''}w_{j'',j'}^{2}\frac{\partial^{2}L(q)}{\partial a_{j''}^{2}} - f''(a_{j'})\frac{\partial L(q)}{\partial x_{j'}} 
\end{aligned} 
\end{equation}
and for the final layer, $L$ we have the boundary condition,
\begin{equation}\label{backprop_taylor_last}
\begin{aligned}
\frac{\partial^{2}L(q)}{\partial a_{j'}^{2}} = 2f'(a_{j'})^{2} -2(d_{j'}-x_{j'})f''(a_{j'}),
\end{aligned} 
\end{equation}
where $x_{j'}$ is the output of the $j'^{\;th}$ neuron in layer $L$, $f'(a_{j'})$ is the gradient with respect to the activation function at node $j'$, $f'(a_{j''})$ is the diagonal component of the Hessian at the node $j'$,  and $d_{j'}$ is the ground truth or desired output of the node. The Levenberg-Marquardt approximation proceeds by eliminating all $f''(\cdot)$ terms such that the quantity $L_{j}^{+}-L_{j}^{-}$ can be efficiently approximated simultaneously with the typical backpropagation training procedure, thereby requiring no additional computational overhead whereas computing the diagonal hessian requires one additional pass per parameter. Regardless of the choice of utilizing the Levenberg-Marquardt approximation, the algorithm remains the same.The overall structure of our proposed approach is summarized in Figure \ref{flow_diag}.

\newpage
\section{Algorithm}\label{alg}
We present the likelihood-guided regularization method fitting procedure in Algorithm \ref{alg2}.

\begin{algorithm}[htbp]
    \footnotesize
    \caption{Model fitting via Likelihood-guided Regularization}
    \label{alg2}
    \begin{algorithmic}[1]
    \State Given the architecture and its initializing hyperparameters
    \State Train the model by backpropagating the negative log likelihood, 
    $$-\log [p(\mathbf{y_{n}|X_{n},\widehat{W}, \boldsymbol{\hat \xi}}))]$$
    onto the means, $\mathbf{M}$ of the proposed variational posterior on $\mathbf{W}$ while setting $\boldsymbol{\xi}=\mathbf{1}$ until a reasonable minimum of the loss function is obtained 
    \State For each new epoch compute the diagonal of the Hessian for each element of $\mathbf{M}$
    \State Recover the saliency approximation for each parameter in $\mathbf{M}$ as

    \begin{equation*}
    \begin{aligned}
    L_{j}^{+}-L_{j}^{-} \approx \frac{\partial^{2}L(q)}{\partial w_{j',j}^{2}}, 
    \end{aligned} 
    \end{equation*}
    
    \State Sample $\boldsymbol{\xi}$ from the proposed variational posterior

    \begin{equation*}
    \begin{aligned}
    q(\xi_{j', \; j}^{(l)}=1) = \bigg{[}1+\exp\bigg{\{}-2\frac{\sum_{j''\in\;(l+1)}w_{j'',j'}^{2}E_{q}[\xi_{j'',j'}^{(l+1)}]}{\sum_{j''\in\;(l+1)}w_{j'',j'}^{2}}-(L_{j}^{+}-L_{j}^{-})\bigg{\}}\bigg{]}^{-1},
    \end{aligned} 
    \end{equation*}
        
    \State Apply the sampled mask, do a forward pass then backpropagate the error
    \State Sample $\mathbf{W}$ from the updated variational posterior

    \begin{equation*}
    \begin{aligned}
    q_{\mathbf{M}}(\mathbf{W}|\boldsymbol{\xi}) = \prod_{j,j'} \xi_{j,j'}\mathcal{N}(\mathbf{W}; 0, \sigma^{2})+(1-\xi_{j,j'})\mathcal{N}(\mathbf{W}; m_{j,j'}, \sigma^{2}),
    \end{aligned} 
    \end{equation*}
    
    \State Iterate through steps 3-6 until a reasonable minimizer is attained
    \end{algorithmic}
\end{algorithm}

\begin{figure}[htbp]
\centering
\hspace*{-1cm}
\includegraphics[scale=0.6]{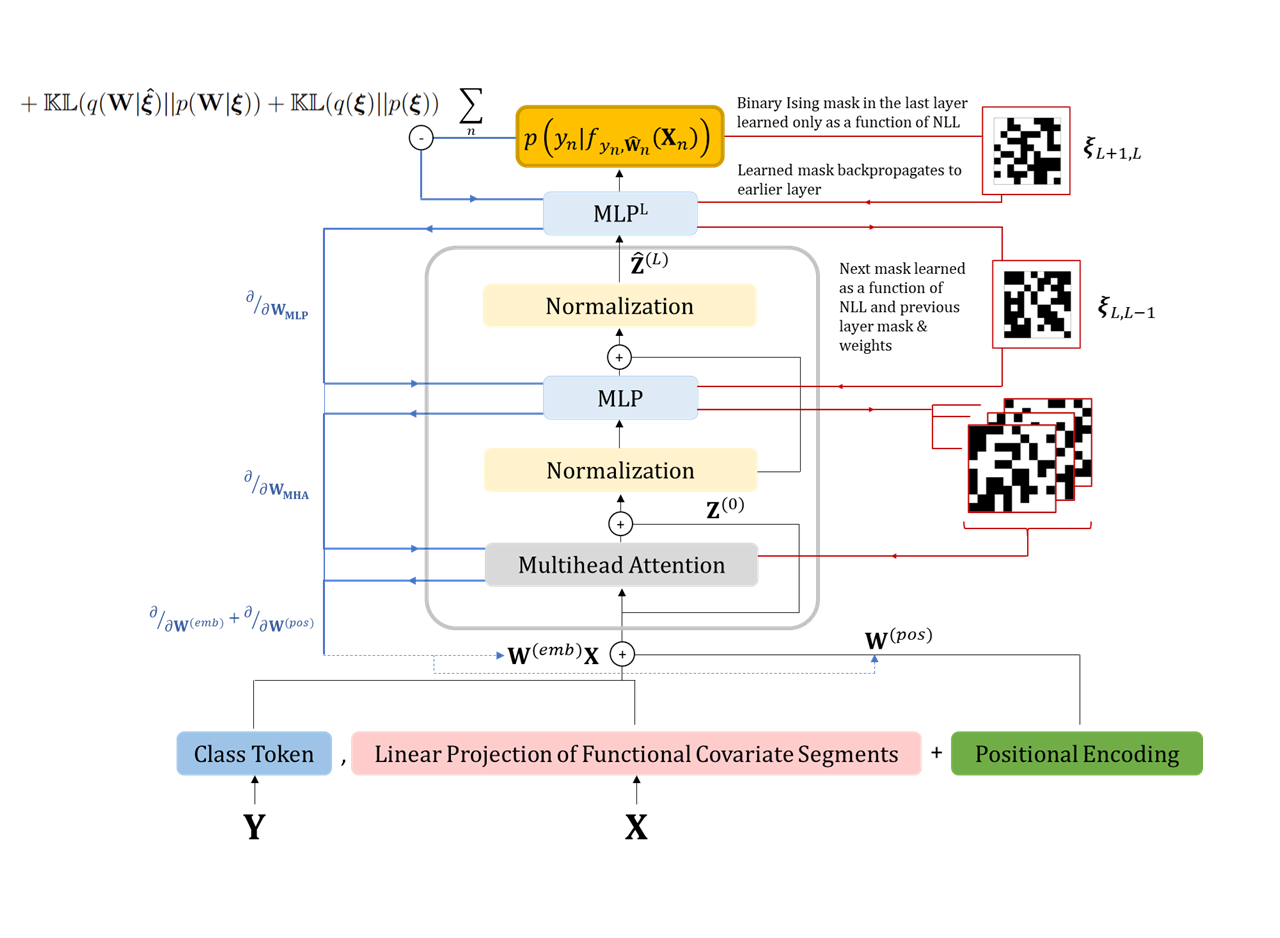}	
\caption{The figure displays the backpropagation of masks alongside the backpropagation of weights in an alternating process illustrated on a simplified vision transformer.}
\label{illust_model}
\end{figure}

\newpage
\section{Statistical properties}\label{stat_props}

Given the variational distributions $q(\mathbf{W}|\boldsymbol{\xi})$ and $q(\boldsymbol{\xi})$, we have the following statistical properties.
\begin{proposition}
For any input $\mathbf{X}$, we have the following properties of our likelihood-guided regularization approach:
\begin{enumerate}[(a)]
    \item Computing the model output $f_{\mathbf{y,W, \boldsymbol{\xi}}}(\mathbf{X})$ is equivalent to performing a forward pass on the input $\mathbf{X}$ and applying the drop with probability $pr(\xi_{j,j'}=1)$ in likelihood-guided regularization of attention based model.
    \item The posterior $p(y,\mathbf{X})\approx \frac{1}{T} \sum_{t=1}^T f_{\mathbf{y,\widehat{W}_t, \hat{\boldsymbol{\xi}}_t}}(\mathbf{X})$, where $f_{\mathbf{y,\widehat{W}_t, \hat{\boldsymbol{\xi}}_t}}(\mathbf{X})$ is the output of the $t^{th}$ forward pass through the likelihood-guided regularization approach.
    \item The prediction uncertainty over the prediction $p(\mathbf{y}^*|\mathbf{X}^*)$ can be approximated by the Bayesian lower and upper credible intervals of the $T$ forward passes.
\end{enumerate}
\end{proposition}
This proposition describes how our approach can provide uncertain quantification effectively.

\begin{proposition}
If $pr(\xi_{j,j'}=1)=\pi$ for all $(j, j')$ and $q(\boldsymbol{\xi})=O(1)$, our likelihood-guided regularization of attention based model aligns with BayesFormer\citep{sankararaman2022bayesformer}. 
\end{proposition}
This property describes our approach is a generalization of existing BayesFormer.

\section{Experiments}\label{exp_dat}

\subsection{Experimental Setup}
We conduct experiments to evaluate the performance of our proposed Ising-regularized Vision Transformer (Ising-ViT) against two comparison methods: 
(1) a Bayesian ViT with fixed dropout probabilities, and 
(2) a Bayesian ViT with fixed dropconnect probabilities; { both of which are regularization methods that involve stochastic removal of network components.}


We test these methods across multiple benchmark datasets: MNIST, FashionMNIST, CIFAR-10, and CIFAR-100 (sample visualizations shown in \ref{MNIST_viz}-\ref{Cifar100_viz}), with the order of the datasets reflecting the increasing complexity of the features in the images. Evaluating each at 3 different training data sizes and two settings for the regularization hyperparameter (0.1 and 0.5). For our proposed method, we alter the baseline probability by setting $\delta$ in $q(\xi_{j', \; j}^{(l)}=1)$, where
\begin{equation*}
\begin{aligned}
q(\xi_{j', \; j}^{(l)}=1) = \bigg{[}1+\exp\bigg{\{}-2\frac{\sum_{j''\in\;(l+1)}w_{j'',j'}^{2}E_{q}[\xi_{j'',j'}^{(l+1)}]}{\sum_{j''\in\;(l+1)}w_{j'',j'}^{2}}-(L_{j}^{+}-L_{j}^{-}) - \log \bigg{(} \frac{\delta}{1-\delta}\bigg{)}\bigg{\}}\bigg{]}^{-1}.
\end{aligned} 
\end{equation*}
Note that $\delta =0.5$ leaves the term unbiased, such that the default dropout probability occurring at $\exp{\{}0{\}}$ is $0.5$ while a smaller delta reduces this probability ($\delta=0.1$ in the absence of other input leads to a keep probability of $0.9$).

To evaluate the performance of the models under examination, we observe five metrics: Accuracy, Recall, Precision, False Positive Rate, and F1 Score. Accuracy provides an overall measure of correctly classified samples across all classes. For the remaining four metrics, we compute each metric by assessing the performance of the model in terms of one-vs-all for each class, treating one class at a time as the positive class. The final metrics are produced by applying a simple average across all class metrics.

\subsection{Results}
\begin{figure}[htbp]
\centering
\includegraphics[scale=0.3]{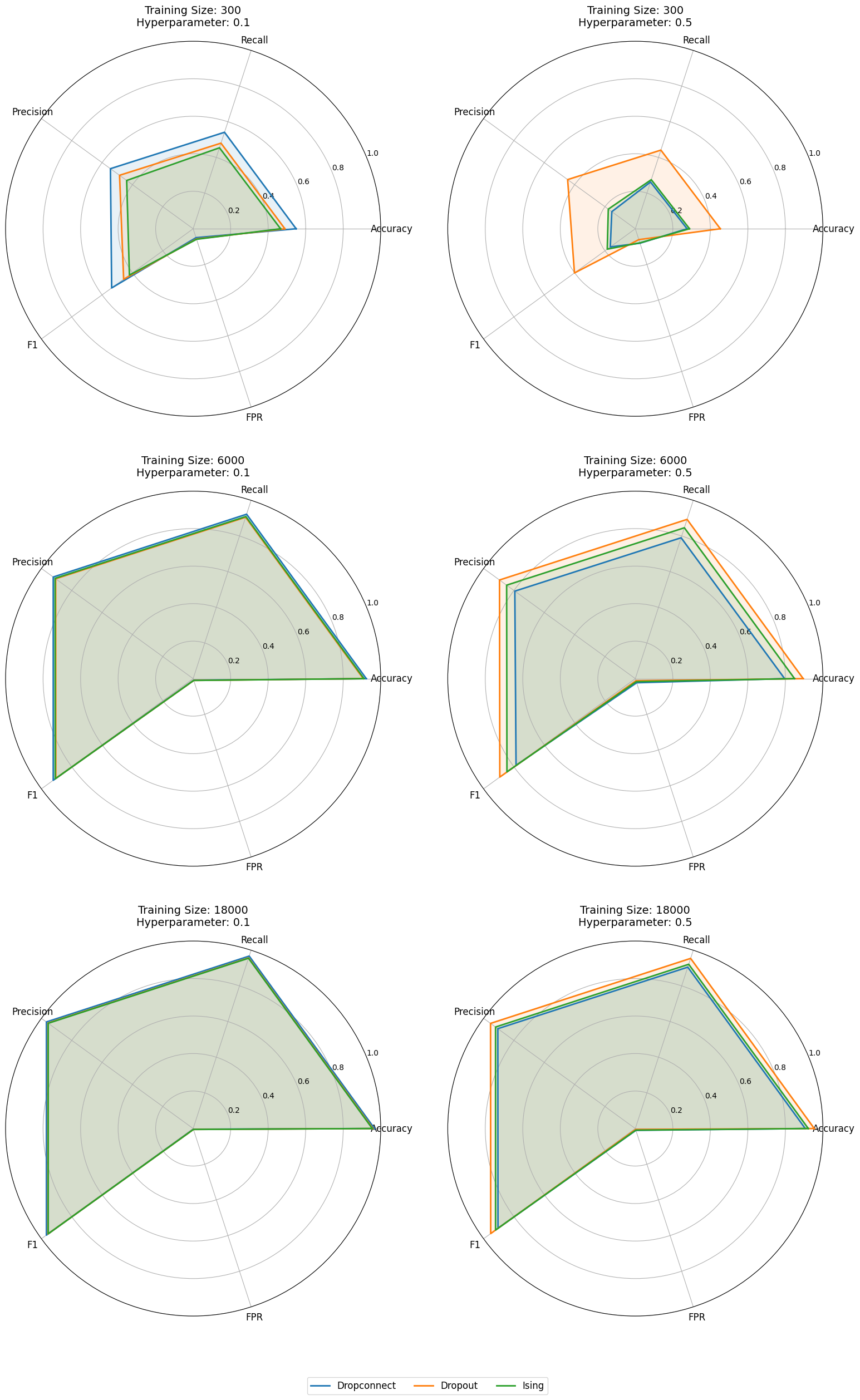}	
\caption{Visualization of accuracy, recall, precision, F1-score, and false positive rate obtained using the MNIST dataset.}
\label{tbl_MNIST}
\end{figure}

\begin{figure}[htbp]
\centering
\includegraphics[scale=0.3]{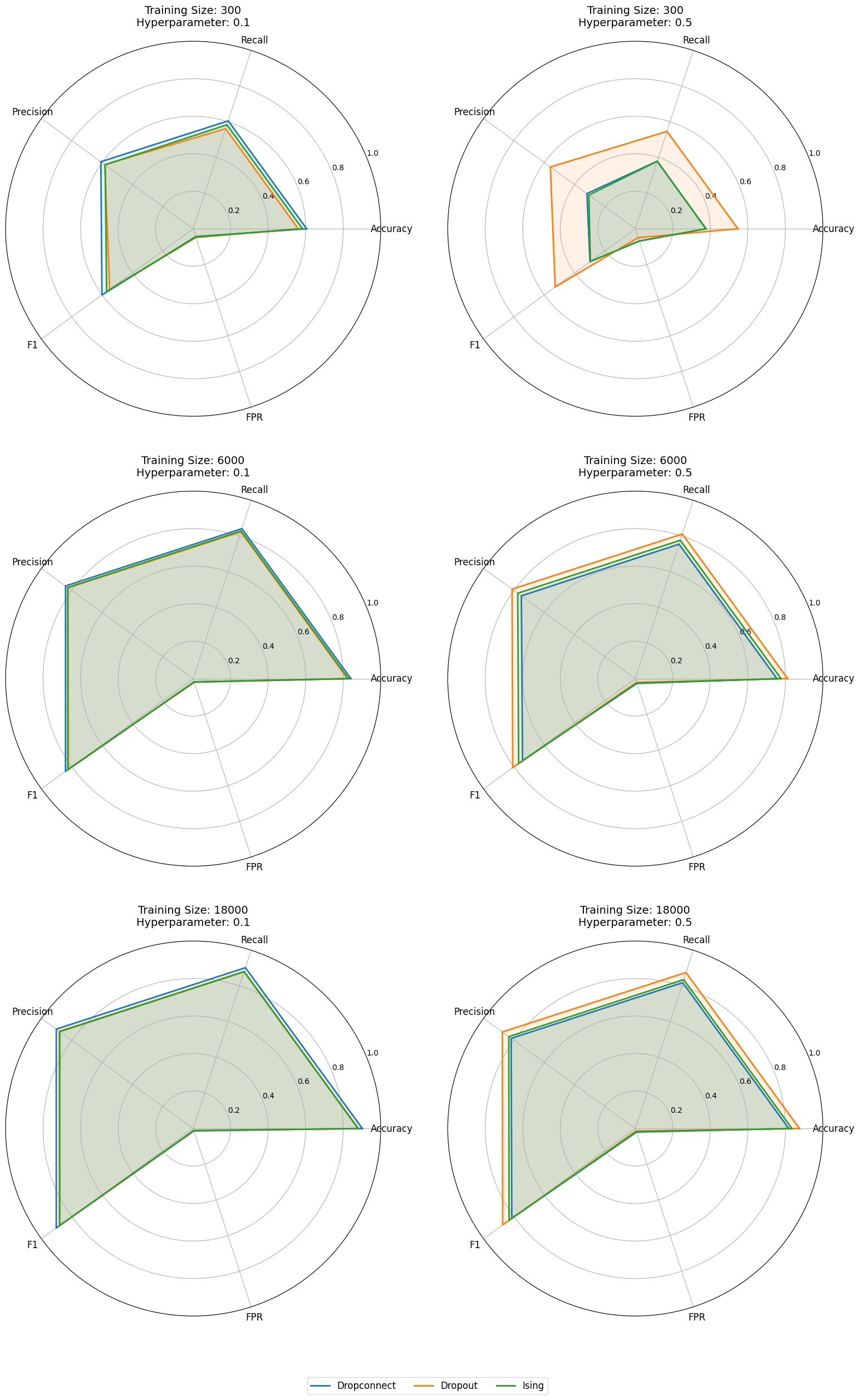}	
\caption{Visualization of accuracy, recall, precision, F1-score, and false positive rate obtained using the Fashion-MNIST dataset.}
\label{tbl_fashionMNIST}
\end{figure}

\begin{figure}[htbp]
\centering
\includegraphics[scale=0.3]{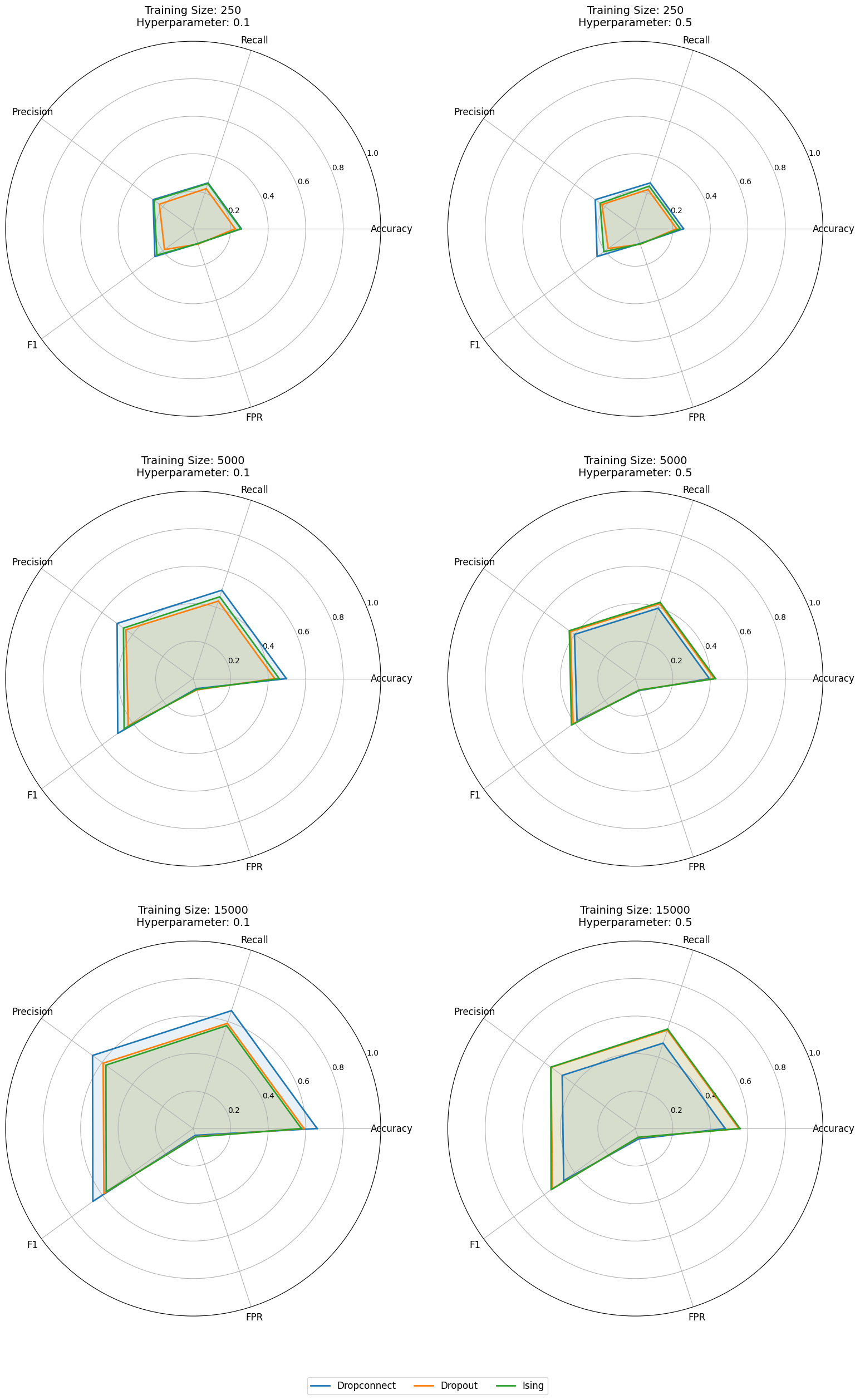}	
\caption{Visualization ofaccuracy, recall, precision, F1-score, and false positive rate obtained using the Cifar10 dataset.}
\label{tbl_cifar10}
\end{figure}

\begin{figure}[htbp]
\centering
\includegraphics[scale=0.3]{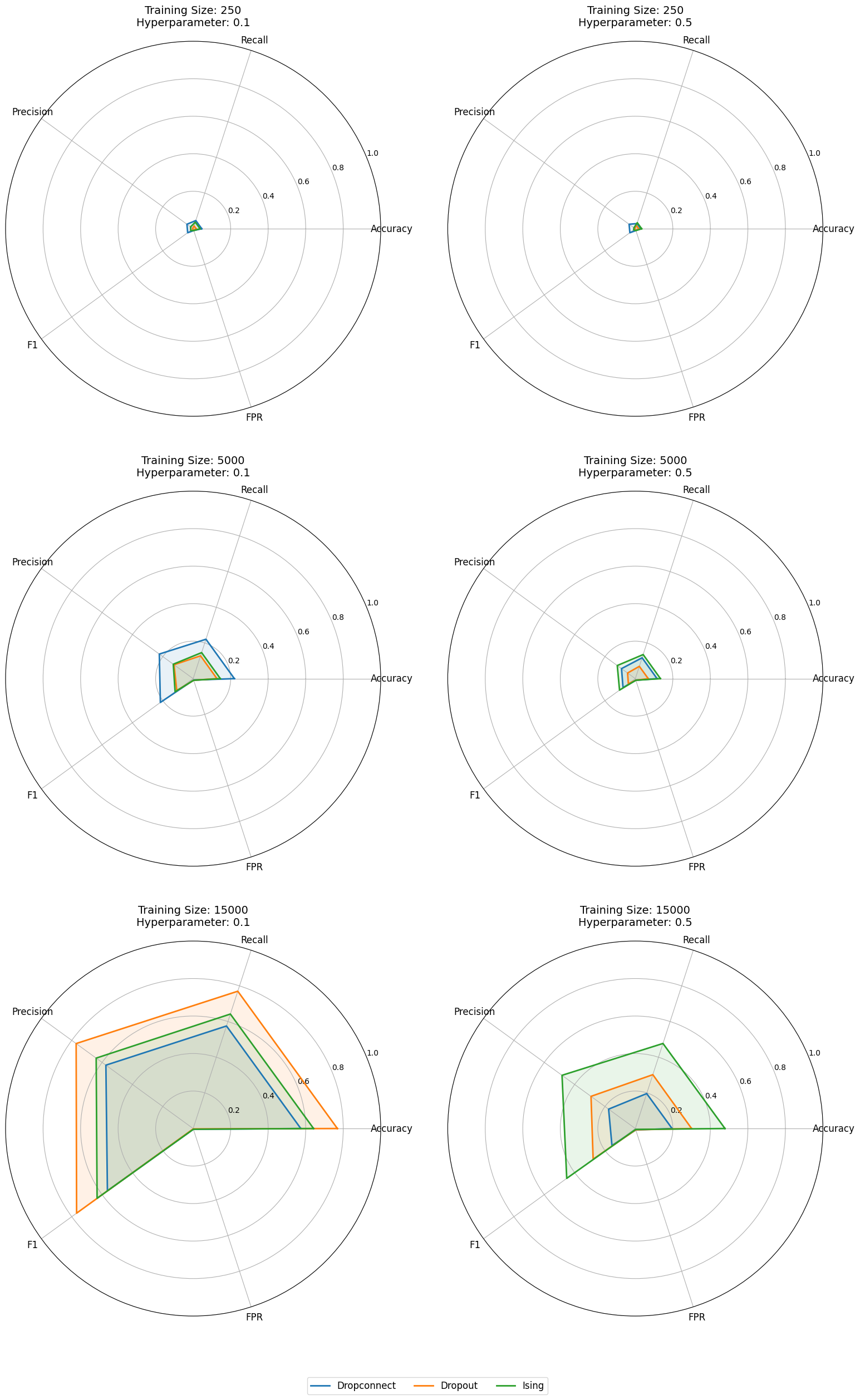}	
\caption{Visualization of accuracy, recall, precision, F1-score, and false positive rate obtained using the Cifar100 dataset.}
\label{tbl_cifar100}
\end{figure}

Experimental results are summarized in Tables \ref{mnist_results_0.1}-\ref{cifar100_results_0.5}, and visualized for clarity in Figures \ref{tbl_MNIST}-\ref{tbl_cifar100}. Each model is trained on the same varying dataset sizes to analyze the effect of regularization across different data regimes. Performance metrics, including classification accuracy, precision, recall, false positive rate, and F1-score, are evaluated on 10 randomly sampled training datasets for each dataset size. {The results presented in Tables \ref{mnist_results_0.1}-\ref{cifar100_results_0.5} showcase the comparative performance of dropconnect, dropout, and Ising regularization methods across our chosen datasets and training sample sizes. The observed trends indicate that performance metrics such as accuracy, macro-averaged recall, precision, F1-score, and false positive rate vary across methods and dataset sizes, highlighting the different strengths of the three methods across different data regimes. For smaller training sample sizes, dropconnect displays comparatively higher performance on accuracy, recall, and precision. However, Ising seems to consistently achieve a more balanced performance across most metrics in the more complex datasets and smaller training sizes. Tables  \ref{cifar100_results_0.1} and \ref{cifar100_results_0.5} highlights the Ising regularization's outperformance in F1-score and metrics stability under smaller data regimes with high task complexity.}

To gauge the performance of the models in terms of probability calibration, we examine the entropy distribution separately for correct and incorrect classifications, shown in Figures \ref{mnist_entropy_0.1_val0.6}-\ref{cifar100_entropy_0.5_val0.895}. Across datasets and hyperparameter settings, the entropy distributions reveal consistent behavioral differences between dropout, dropconnect, and the Ising regularization approach; dropout consistently produces more sharply peaked entropy near zero for both correct and incorrect predictions across all training dataset sizes, reflecting strong confidence but also a clear tendency toward overconfidence, where entropy often remains low even for misclassifications. This highlights dropout’s relative weakness with respect to probability calibration as it struggles to represent meaningful epistemic uncertainty. Dropconnect, by contrast, generates smoother and more interpretable entropy curves, especially as task complexity increases. Dropconnect entropy distributions expand more naturally for incorrect predictions, indicating better uncertainty separation and a stronger capacity to express model doubt. Finally, despite being overly conservative, as evidenced by the lower entropy around zero in correct classifications, the Ising regularization seemingly delivers the most balanced and well-calibrated uncertainty representation overall; its entropy distributions are symmetric and Gaussian-like, slightly moreso than those of the dropconnect setting. In addition to aleatoric uncertainty captured through the Bayesian layers, we expect the likelihood-guided Ising regularization to capture epistemic uncertainty arising from both the architecture and the weights. In practice, Dropout remains suitable for tasks where calibrated uncertainty is secondary. Dropconnect provides a reliable middle ground for data-limited regimes. Ising regularization stands out in settings where uncertainty estimation itself is critical, such as Bayesian inference, active learning, and safety-sensitive decision systems—where understanding why the model is uncertain matters as much as achieving high accuracy, especially considering the posterior on architectures provided by the Ising approach at no additional computational cost.

To complement entropy-based uncertainty profiles, we examine calibration curves across datasets and regularization strategies, shown in Figures \ref{mnist_calibration_0.1_val0.6}-\ref{cifar100_calibration_0.5_val0.895} of the supplementary materials.

The respective plots visualize the alignment between predicted confidence and empirical accuracy where well-calibrated models should closely follow the identity line. The comparative performance of dropout, dropconnect, and Ising regularization aligns with and expands upon the entropy-based interpretations; across regularization levels and training set sizes, dropout performs well for less complicated datasets. Poor dropout calibration manifests mostly under small training sets sizes for complex datasets. Dropconnect calibration depends significantly on the level of the regularization hyperparameter; on simpler datasets, higher levels of regularization lead to underconfidence for dropconnect, but lower levels lead to similar calibration performance to dropout. For more complex training sets, dropconnect tends towards overconfidence for smaller datasets but underconfidence for larger training samples for both regularization levels. The Ising-regularization setting tends to show underconfidence for simpler datasets in general, across samples sizes, more so for higher levels of the regularization offset. The performance of the Ising regularization is comparatively the best in more complex datasets with smaller training data sizes, where calibration curves are closer to the identity than those of dropconnect or dropout. As such, the proposed Ising regularization is well-equipped to express calibrated uncertainty under data sparsity.

Finally, a potentially effective way to mitigate the Ising model’s underconfidence while leaving the probability calibration uncompromised, is to underweight the prior term in the dropout probability computation, allowing the data-driven loss difference to play a more dominant role. The current formulation's prior term backpropagates dropout probabilities according to the strength of outgoing weights. This may lead to over-regularization of weights in bottleneck layers. By scaling this term down, the posterior dropout probabilities become more responsive to the empirical direct impact of removal on empirical risk minimization, sharpening confidence for well-determined weights while maintaining uncertainty elsewhere. This adjustment preserves the Bayesian structure of the model but shifts it toward a more empirical-Bayes behavior, achieving a cleaner balance between confidence and calibration in the Ising-regularized regime.


\begin{table}[htbp]
\centering
\begin{tabular}{ccccccc}
\hline
Training Sample Size & \textbf{Method} & $\mathbf{Accuracy}$ & $\mathbf{Recall}$ & $\mathbf{Precision}$ & $\mathbf{F_{1}}$ & $\mathbf{FPR}$ \\
\hline
300 & Dropconnect & \makecell{0.549 \\ {\scriptsize (0.023)}} 
    & \makecell{0.541 \\ {\scriptsize (0.024)}}  
    & \makecell{0.545 \\ {\scriptsize (0.027)}} 
    & \makecell{0.537 \\ {\scriptsize (0.025)}} 
    & \makecell{0.050 \\ {\scriptsize (0.003)}} \\
    & Dropout & \makecell{0.489 \\ {\scriptsize (0.033)}} 
    & \makecell{0.480 \\ {\scriptsize (0.033)}}  
    & \makecell{0.485 \\ {\scriptsize (0.033)}} 
    & \makecell{0.458 \\ {\scriptsize (0.042)}} 
    & \makecell{0.057 \\ {\scriptsize (0.004)}} \\
    & Ising & \makecell{0.466 \\ {\scriptsize (0.036)}} 
    & \makecell{0.454 \\ {\scriptsize (0.035)}}  
    & \makecell{0.438 \\ {\scriptsize (0.039)}} 
    & \makecell{0.420 \\ {\scriptsize (0.042)}} 
    & \makecell{0.059 \\ {\scriptsize (0.004)}} \\
\hline
6000 & Dropconnect & \makecell{0.923 \\ {\scriptsize (0.004)}} 
    & \makecell{0.922 \\ {\scriptsize (0.004)}}  
    & \makecell{0.922 \\ {\scriptsize (0.004)}} 
    & \makecell{0.922 \\ {\scriptsize (0.004)}} 
    & \makecell{0.009 \\ {\scriptsize (0.000)}} \\
    & Dropout & \makecell{0.907 \\ {\scriptsize (0.006)}} 
    & \makecell{0.905 \\ {\scriptsize (0.006)}}  
    & \makecell{0.906 \\ {\scriptsize (0.006)}} 
    & \makecell{0.905 \\ {\scriptsize (0.006)}} 
    & \makecell{0.010 \\ {\scriptsize (0.001)}} \\
    & Ising & \makecell{0.911 \\ {\scriptsize (0.004)}} 
    & \makecell{0.909 \\ {\scriptsize (0.004)}}  
    & \makecell{0.910 \\ {\scriptsize (0.004)}} 
    & \makecell{0.909 \\ {\scriptsize (0.004)}} 
    & \makecell{0.010 \\ {\scriptsize (0.000)}} \\
\hline
18000 & Dropconnect & \makecell{0.967 \\ {\scriptsize (0.002)}} 
    & \makecell{0.967 \\ {\scriptsize (0.002)}}  
    & \makecell{0.967 \\ {\scriptsize (0.002)}} 
    & \makecell{0.967 \\ {\scriptsize (0.002)}} 
    & \makecell{0.004 \\ {\scriptsize (0.000)}} \\
    & Dropout & \makecell{0.958 \\ {\scriptsize (0.003)}} 
    & \makecell{0.957 \\ {\scriptsize (0.003)}}  
    & \makecell{0.957 \\ {\scriptsize (0.003)}} 
    & \makecell{0.957 \\ {\scriptsize (0.003)}} 
    & \makecell{0.005 \\ {\scriptsize (0.000)}} \\
    & Ising & \makecell{0.955 \\ {\scriptsize (0.002)}} 
    & \makecell{0.955 \\ {\scriptsize (0.002)}}  
    & \makecell{0.955 \\ {\scriptsize (0.002)}} 
    & \makecell{0.955 \\ {\scriptsize (0.002)}} 
    & \makecell{0.005 \\ {\scriptsize (0.000)}} \\
\hline
\end{tabular}
\caption{MNIST - 0.1: Simulation results to investigate the performance of different training set sizes using accuracy, recall, precision, F1-score, and false positive rate}
\label{mnist_results_0.1}
\end{table}

\begin{table}[htbp]
\centering
\begin{tabular}{ccccccc}
\hline
Training Sample Size & \textbf{Method} & $\mathbf{Accuracy}$ & $\mathbf{Recall}$ & $\mathbf{Precision}$ & $\mathbf{F_{1}}$ & $\mathbf{FPR}$ \\
\hline
300 & Dropconnect & \makecell{0.276 \\ {\scriptsize (0.051)}} 
    & \makecell{0.262 \\ {\scriptsize (0.046)}}  
    & \makecell{0.155 \\ {\scriptsize (0.052)}} 
    & \makecell{0.165 \\ {\scriptsize (0.051)}} 
    & \makecell{0.081 \\ {\scriptsize (0.006)}} \\
    & Dropout & \makecell{0.453 \\ {\scriptsize (0.041)}} 
    & \makecell{0.441 \\ {\scriptsize (0.042)}}  
    & \makecell{0.446 \\ {\scriptsize (0.044)}} 
    & \makecell{0.401 \\ {\scriptsize (0.062)}} 
    & \makecell{0.061 \\ {\scriptsize (0.005)}} \\
    & Ising & \makecell{0.289 \\ {\scriptsize (0.050)}} 
    & \makecell{0.275 \\ {\scriptsize (0.045)}}  
    & \makecell{0.178 \\ {\scriptsize (0.064)}} 
    & \makecell{0.185 \\ {\scriptsize (0.055)}} 
    & \makecell{0.079 \\ {\scriptsize (0.006)}} \\
\hline
6000 & Dropconnect & \makecell{0.796 \\ {\scriptsize (0.011)}} 
    & \makecell{0.790 \\ {\scriptsize (0.012)}}  
    & \makecell{0.794 \\ {\scriptsize (0.014)}} 
    & \makecell{0.786 \\ {\scriptsize (0.015)}} 
    & \makecell{0.023 \\ {\scriptsize (0.001)}} \\
    & Dropout & \makecell{0.895 \\ {\scriptsize (0.006)}} 
    & \makecell{0.893 \\ {\scriptsize (0.006)}}  
    & \makecell{0.895 \\ {\scriptsize (0.006)}} 
    & \makecell{0.893 \\ {\scriptsize (0.006)}} 
    & \makecell{0.012 \\ {\scriptsize (0.001)}} \\
    & Ising & \makecell{0.849 \\ {\scriptsize (0.013)}} 
    & \makecell{0.846 \\ {\scriptsize (0.014)}}  
    & \makecell{0.848 \\ {\scriptsize (0.014)}} 
    & \makecell{0.846 \\ {\scriptsize (0.014)}} 
    & \makecell{0.017 \\ {\scriptsize (0.001)}} \\
\hline
18000 & Dropconnect & \makecell{0.906 \\ {\scriptsize (0.005)}} 
    & \makecell{0.905 \\ {\scriptsize (0.005)}}  
    & \makecell{0.906 \\ {\scriptsize (0.004)}} 
    & \makecell{0.905 \\ {\scriptsize (0.005)}} 
    & \makecell{0.010 \\ {\scriptsize (0.001)}} \\
    & Dropout & \makecell{0.954 \\ {\scriptsize (0.003)}} 
    & \makecell{0.954 \\ {\scriptsize (0.003)}}  
    & \makecell{0.954 \\ {\scriptsize (0.003)}} 
    & \makecell{0.953 \\ {\scriptsize (0.003)}} 
    & \makecell{0.005 \\ {\scriptsize (0.000)}} \\
    & Ising & \makecell{0.922 \\ {\scriptsize (0.002)}} 
    & \makecell{0.921 \\ {\scriptsize (0.003)}}  
    & \makecell{0.921 \\ {\scriptsize (0.002)}} 
    & \makecell{0.921 \\ {\scriptsize (0.002)}} 
    & \makecell{0.009 \\ {\scriptsize (0.000)}} \\
\hline
\end{tabular}
\caption{MNIST - 0.5: Simulation results to investigate the performance of different training set sizes using accuracy, recall, precision, F1-score, and false positive rate}
\label{mnist_results_0.5}
\end{table}

\begin{table}[htbp]
\centering
\begin{tabular}{ccccccc}
\hline
Training Sample Size & \textbf{Method} & $\mathbf{Accuracy}$ & $\mathbf{Recall}$ & $\mathbf{Precision}$ & $\mathbf{F_{1}}$ & $\mathbf{FPR}$ \\
\hline
300 & Dropconnect & \makecell{0.605 \\ {\scriptsize (0.013)}} 
    & \makecell{0.605 \\ {\scriptsize (0.013)}}  
    & \makecell{0.608 \\ {\scriptsize (0.010)}} 
    & \makecell{0.601 \\ {\scriptsize (0.010)}} 
    & \makecell{0.044 \\ {\scriptsize (0.001)}} \\
    & Dropout & \makecell{0.560 \\ {\scriptsize (0.021)}} 
    & \makecell{0.561 \\ {\scriptsize (0.021)}}  
    & \makecell{0.582 \\ {\scriptsize (0.027)}} 
    & \makecell{0.550 \\ {\scriptsize (0.023)}} 
    & \makecell{0.049 \\ {\scriptsize (0.002)}} \\
    & Ising & \makecell{0.583 \\ {\scriptsize (0.016)}} 
    & \makecell{0.583 \\ {\scriptsize (0.017)}}  
    & \makecell{0.580 \\ {\scriptsize (0.025)}} 
    & \makecell{0.570 \\ {\scriptsize (0.025)}} 
    & \makecell{0.046 \\ {\scriptsize (0.002)}} \\
\hline
6000 & Dropconnect & \makecell{0.841 \\ {\scriptsize (0.005)}} 
    & \makecell{0.841 \\ {\scriptsize (0.005)}}  
    & \makecell{0.841 \\ {\scriptsize (0.005)}} 
    & \makecell{0.841 \\ {\scriptsize (0.005)}} 
    & \makecell{0.018 \\ {\scriptsize (0.001)}} \\
    & Dropout & \makecell{0.823 \\ {\scriptsize (0.008)}} 
    & \makecell{0.823 \\ {\scriptsize (0.007)}}  
    & \makecell{0.824 \\ {\scriptsize (0.007)}} 
    & \makecell{0.822 \\ {\scriptsize (0.007)}} 
    & \makecell{0.020 \\ {\scriptsize (0.001)}} \\
    & Ising & \makecell{0.828 \\ {\scriptsize (0.005)}} 
    & \makecell{0.828 \\ {\scriptsize (0.005)}}  
    & \makecell{0.827 \\ {\scriptsize (0.005)}} 
    & \makecell{0.826 \\ {\scriptsize (0.005)}} 
    & \makecell{0.019 \\ {\scriptsize (0.001)}} \\
\hline
18000 & Dropconnect & \makecell{0.902 \\ {\scriptsize (0.004)}} 
    & \makecell{0.902 \\ {\scriptsize (0.004)}}  
    & \makecell{0.902 \\ {\scriptsize (0.005)}} 
    & \makecell{0.902 \\ {\scriptsize (0.005)}} 
    & \makecell{0.011 \\ {\scriptsize (0.000)}} \\
    & Dropout & \makecell{0.881 \\ {\scriptsize (0.005)}} 
    & \makecell{0.881 \\ {\scriptsize (0.005)}}  
    & \makecell{0.882 \\ {\scriptsize (0.005)}} 
    & \makecell{0.881 \\ {\scriptsize (0.005)}} 
    & \makecell{0.013 \\ {\scriptsize (0.001)}} \\
    & Ising & \makecell{0.879 \\ {\scriptsize (0.004)}} 
    & \makecell{0.879 \\ {\scriptsize (0.004)}}  
    & \makecell{0.879 \\ {\scriptsize (0.004)}} 
    & \makecell{0.879 \\ {\scriptsize (0.004)}} 
    & \makecell{0.013 \\ {\scriptsize (0.000)}} \\
\hline
\end{tabular}
\caption{FASHION-MNIST - 0.1: Simulation results to investigate the performance of different training set sizes using accuracy, recall, precision, F1-score, and false positive rate}
\label{fashionmnist_results_0.1}
\end{table}

\begin{table}[htbp]
\centering
\begin{tabular}{ccccccc}
\hline
Training Sample Size & \textbf{Method} & $\mathbf{Accuracy}$ & $\mathbf{Recall}$ & $\mathbf{Precision}$ & $\mathbf{F_{1}}$ & $\mathbf{FPR}$ \\
\hline
300 & Dropconnect & \makecell{0.377 \\ {\scriptsize (0.028)}} 
    & \makecell{0.379 \\ {\scriptsize (0.029)}}  
    & \makecell{0.319 \\ {\scriptsize (0.057)}} 
    & \makecell{0.299 \\ {\scriptsize (0.039)}} 
    & \makecell{0.069 \\ {\scriptsize (0.003)}} \\
    & Dropout & \makecell{0.547 \\ {\scriptsize (0.041)}} 
    & \makecell{0.546 \\ {\scriptsize (0.040)}}  
    & \makecell{0.560 \\ {\scriptsize (0.030)}} 
    & \makecell{0.528 \\ {\scriptsize (0.056)}} 
    & \makecell{0.050 \\ {\scriptsize (0.005)}} \\
    & Ising & \makecell{0.376 \\ {\scriptsize (0.030)}} 
    & \makecell{0.378 \\ {\scriptsize (0.030)}}  
    & \makecell{0.306 \\ {\scriptsize (0.048)}} 
    & \makecell{0.296 \\ {\scriptsize (0.038)}} 
    & \makecell{0.069 \\ {\scriptsize (0.003)}} \\
\hline
6000 & Dropconnect & \makecell{0.755 \\ {\scriptsize (0.012)}} 
    & \makecell{0.754 \\ {\scriptsize (0.012)}}  
    & \makecell{0.751 \\ {\scriptsize (0.013)}} 
    & \makecell{0.743 \\ {\scriptsize (0.014)}} 
    & \makecell{0.027 \\ {\scriptsize (0.001)}} \\
    & Dropout & \makecell{0.812 \\ {\scriptsize (0.010)}} 
    & \makecell{0.811 \\ {\scriptsize (0.009)}}  
    & \makecell{0.812 \\ {\scriptsize (0.010)}} 
    & \makecell{0.808 \\ {\scriptsize (0.009)}} 
    & \makecell{0.021 \\ {\scriptsize (0.001)}} \\
    & Ising & \makecell{0.777 \\ {\scriptsize (0.013)}} 
    & \makecell{0.776 \\ {\scriptsize (0.013)}}  
    & \makecell{0.775 \\ {\scriptsize (0.017)}} 
    & \makecell{0.769 \\ {\scriptsize (0.017)}} 
    & \makecell{0.025 \\ {\scriptsize (0.002)}} \\
\hline
18000 & Dropconnect & \makecell{0.818 \\ {\scriptsize (0.005)}} 
    & \makecell{0.818 \\ {\scriptsize (0.005)}}  
    & \makecell{0.818 \\ {\scriptsize (0.005)}} 
    & \makecell{0.814 \\ {\scriptsize (0.005)}} 
    & \makecell{0.020 \\ {\scriptsize (0.001)}} \\
    & Dropout & \makecell{0.875 \\ {\scriptsize (0.006)}} 
    & \makecell{0.875 \\ {\scriptsize (0.006)}}  
    & \makecell{0.876 \\ {\scriptsize (0.007)}} 
    & \makecell{0.874 \\ {\scriptsize (0.007)}} 
    & \makecell{0.014 \\ {\scriptsize (0.001)}} \\
    & Ising & \makecell{0.835 \\ {\scriptsize (0.005)}} 
    & \makecell{0.835 \\ {\scriptsize (0.005)}}  
    & \makecell{0.834 \\ {\scriptsize (0.005)}} 
    & \makecell{0.832 \\ {\scriptsize (0.005)}} 
    & \makecell{0.018 \\ {\scriptsize (0.001)}} \\
\hline
\end{tabular}
\caption{FASHION-MNIST - 0.5: Simulation results to investigate the performance of different training set sizes using accuracy, recall, precision, F1-score, and false positive rate}
\label{fashionmnist_results_0.5}
\end{table}

\begin{table}[htbp]
\centering
\begin{tabular}{ccccccc}
\hline
Training Sample Size & \textbf{Method} & $\mathbf{Accuracy}$ & $\mathbf{Recall}$ & $\mathbf{Precision}$ & $\mathbf{F_{1}}$ & $\mathbf{FPR}$ \\
\hline
250 & Dropconnect & \makecell{0.257 \\ {\scriptsize (0.009)}} 
    & \makecell{0.257 \\ {\scriptsize (0.010)}}  
    & \makecell{0.264 \\ {\scriptsize (0.015)}} 
    & \makecell{0.252 \\ {\scriptsize (0.011)}} 
    & \makecell{0.083 \\ {\scriptsize (0.001)}} \\
    & Dropout & \makecell{0.224 \\ {\scriptsize (0.019)}} 
    & \makecell{0.225 \\ {\scriptsize (0.018)}}  
    & \makecell{0.223 \\ {\scriptsize (0.045)}} 
    & \makecell{0.189 \\ {\scriptsize (0.030)}} 
    & \makecell{0.086 \\ {\scriptsize (0.002)}} \\
    & Ising & \makecell{0.254 \\ {\scriptsize (0.011)}} 
    & \makecell{0.254 \\ {\scriptsize (0.012)}}  
    & \makecell{0.258 \\ {\scriptsize (0.016)}} 
    & \makecell{0.239 \\ {\scriptsize (0.012)}} 
    & \makecell{0.083 \\ {\scriptsize (0.001)}} \\
\hline
5000 & Dropconnect & \makecell{0.497 \\ {\scriptsize (0.007)}} 
    & \makecell{0.496 \\ {\scriptsize (0.007)}}  
    & \makecell{0.501 \\ {\scriptsize (0.007)}} 
    & \makecell{0.497 \\ {\scriptsize (0.007)}} 
    & \makecell{0.056 \\ {\scriptsize (0.001)}} \\
    & Dropout & \makecell{0.435 \\ {\scriptsize (0.016)}} 
    & \makecell{0.435 \\ {\scriptsize (0.016)}}  
    & \makecell{0.443 \\ {\scriptsize (0.017)}} 
    & \makecell{0.428 \\ {\scriptsize (0.019)}} 
    & \makecell{0.063 \\ {\scriptsize (0.002)}} \\
    & Ising & \makecell{0.459 \\ {\scriptsize (0.006)}} 
    & \makecell{0.459 \\ {\scriptsize (0.006)}}  
    & \makecell{0.459 \\ {\scriptsize (0.005)}} 
    & \makecell{0.455 \\ {\scriptsize (0.005)}} 
    & \makecell{0.060 \\ {\scriptsize (0.001)}} \\
\hline
15000 & Dropconnect & \makecell{0.661 \\ {\scriptsize (0.006)}} 
    & \makecell{0.661 \\ {\scriptsize (0.006)}}  
    & \makecell{0.663 \\ {\scriptsize (0.006)}} 
    & \makecell{0.661 \\ {\scriptsize (0.006)}} 
    & \makecell{0.038 \\ {\scriptsize (0.001)}} \\
    & Dropout & \makecell{0.591 \\ {\scriptsize (0.010)}} 
    & \makecell{0.590 \\ {\scriptsize (0.010)}}  
    & \makecell{0.593 \\ {\scriptsize (0.011)}} 
    & \makecell{0.588 \\ {\scriptsize (0.011)}} 
    & \makecell{0.045 \\ {\scriptsize (0.001)}} \\
    & Ising & \makecell{0.577 \\ {\scriptsize (0.008)}} 
    & \makecell{0.577 \\ {\scriptsize (0.008)}}  
    & \makecell{0.575 \\ {\scriptsize (0.008)}} 
    & \makecell{0.573 \\ {\scriptsize (0.008)}} 
    & \makecell{0.047 \\ {\scriptsize (0.001)}} \\
\hline
\end{tabular}
\caption{Cifar10 - 0.1: Simulation results to investigate the performance of different training set sizes using accuracy, recall, precision, F1-score, and false positive rate}
\label{cifar10_results_0.1}
\end{table}

\begin{table}[htbp]
\centering
\begin{tabular}{ccccccc}
\hline
Training Sample Size & \textbf{Method} & $\mathbf{Accuracy}$ & $\mathbf{Recall}$ & $\mathbf{Precision}$ & $\mathbf{F_{1}}$ & $\mathbf{FPR}$ \\
\hline
250 & Dropconnect & \makecell{0.225 \\ {\scriptsize (0.007)}} 
    & \makecell{0.225 \\ {\scriptsize (0.007)}}  
    & \makecell{0.201 \\ {\scriptsize (0.050)}} 
    & \makecell{0.182 \\ {\scriptsize (0.016)}} 
    & \makecell{0.086 \\ {\scriptsize (0.001)}} \\
    & Dropout & \makecell{0.220 \\ {\scriptsize (0.013)}} 
    & \makecell{0.220 \\ {\scriptsize (0.013)}}  
    & \makecell{0.220 \\ {\scriptsize (0.041)}} 
    & \makecell{0.179 \\ {\scriptsize (0.031)}} 
    & \makecell{0.087 \\ {\scriptsize (0.001)}} \\
    & Ising & \makecell{0.239 \\ {\scriptsize (0.014)}} 
    & \makecell{0.239 \\ {\scriptsize (0.013)}}  
    & \makecell{0.232 \\ {\scriptsize (0.019)}} 
    & \makecell{0.208 \\ {\scriptsize (0.018)}} 
    & \makecell{0.085 \\ {\scriptsize (0.002)}} \\
\hline
5000 & Dropconnect & \makecell{0.395 \\ {\scriptsize (0.012)}} 
    & \makecell{0.395 \\ {\scriptsize (0.011)}}  
    & \makecell{0.401 \\ {\scriptsize (0.017)}} 
    & \makecell{0.384 \\ {\scriptsize (0.015)}} 
    & \makecell{0.067 \\ {\scriptsize (0.001)}} \\
    & Dropout & \makecell{0.420 \\ {\scriptsize (0.020)}} 
    & \makecell{0.420 \\ {\scriptsize (0.020)}}  
    & \makecell{0.427 \\ {\scriptsize (0.023)}} 
    & \makecell{0.410 \\ {\scriptsize (0.024)}} 
    & \makecell{0.064 \\ {\scriptsize (0.002)}} \\
    & Ising & \makecell{0.428 \\ {\scriptsize (0.007)}} 
    & \makecell{0.428 \\ {\scriptsize (0.008)}}  
    & \makecell{0.435 \\ {\scriptsize (0.011)}} 
    & \makecell{0.421 \\ {\scriptsize (0.008)}} 
    & \makecell{0.064 \\ {\scriptsize (0.001)}} \\
\hline
15000 & Dropconnect & \makecell{0.479 \\ {\scriptsize (0.007)}} 
    & \makecell{0.479 \\ {\scriptsize (0.007)}}  
    & \makecell{0.482 \\ {\scriptsize (0.006)}} 
    & \makecell{0.472 \\ {\scriptsize (0.007)}} 
    & \makecell{0.058 \\ {\scriptsize (0.001)}} \\
    & Dropout & \makecell{0.552 \\ {\scriptsize (0.008)}} 
    & \makecell{0.552 \\ {\scriptsize (0.008)}}  
    & \makecell{0.555 \\ {\scriptsize (0.009)}} 
    & \makecell{0.546 \\ {\scriptsize (0.009)}} 
    & \makecell{0.050 \\ {\scriptsize (0.001)}} \\
    & Ising & \makecell{0.558 \\ {\scriptsize (0.009)}} 
    & \makecell{0.558 \\ {\scriptsize (0.009)}}  
    & \makecell{0.557 \\ {\scriptsize (0.010)}} 
    & \makecell{0.555 \\ {\scriptsize (0.010)}} 
    & \makecell{0.049 \\ {\scriptsize (0.001)}} \\
\hline
\end{tabular}
\caption{Cifar10 - 0.5: Simulation results to investigate the performance of different training set sizes using accuracy, recall, precision, F1-score, and false positive rate}
\label{cifar10_results_0.5}
\end{table}

\begin{table}[htbp]
\centering
\begin{tabular}{ccccccc}
\hline
Training Sample Size & \textbf{Method} & $\mathbf{Accuracy}$ & $\mathbf{Recall}$ & $\mathbf{Precision}$ & $\mathbf{F_{1}}$ & $\mathbf{FPR}$ \\
\hline
250 & Dropconnect & \makecell{0.046 \\ {\scriptsize (0.006)}} 
    & \makecell{0.046 \\ {\scriptsize (0.006)}}  
    & \makecell{0.041 \\ {\scriptsize (0.005)}} 
    & \makecell{0.036 \\ {\scriptsize (0.004)}} 
    & \makecell{0.010 \\ {\scriptsize (0.000)}} \\
    & Dropout & \makecell{0.016 \\ {\scriptsize (0.005)}} 
    & \makecell{0.016 \\ {\scriptsize (0.005)}}  
    & \makecell{0.003 \\ {\scriptsize (0.002)}} 
    & \makecell{0.003 \\ {\scriptsize (0.001)}} 
    & \makecell{0.010 \\ {\scriptsize (0.000)}} \\
    & Ising & \makecell{0.037 \\ {\scriptsize (0.004)}} 
    & \makecell{0.037 \\ {\scriptsize (0.004)}}  
    & \makecell{0.017 \\ {\scriptsize (0.003)}} 
    & \makecell{0.016 \\ {\scriptsize (0.002)}} 
    & \makecell{0.010 \\ {\scriptsize (0.000)}} \\
\hline
5000 & Dropconnect & \makecell{0.220 \\ {\scriptsize (0.008)}} 
    & \makecell{0.221 \\ {\scriptsize (0.008)}}  
    & \makecell{0.223 \\ {\scriptsize (0.007)}} 
    & \makecell{0.216 \\ {\scriptsize (0.007)}} 
    & \makecell{0.008 \\ {\scriptsize (0.000)}} \\
    & Dropout & \makecell{0.127 \\ {\scriptsize (0.005)}} 
    & \makecell{0.128 \\ {\scriptsize (0.005)}}  
    & \makecell{0.126 \\ {\scriptsize (0.010)}} 
    & \makecell{0.108 \\ {\scriptsize (0.007)}} 
    & \makecell{0.009 \\ {\scriptsize (0.000)}} \\
    & Ising & \makecell{0.145 \\ {\scriptsize (0.005)}} 
    & \makecell{0.146 \\ {\scriptsize (0.005)}}  
    & \makecell{0.131 \\ {\scriptsize (0.007)}} 
    & \makecell{0.119 \\ {\scriptsize (0.006)}} 
    & \makecell{0.009 \\ {\scriptsize (0.000)}} \\
\hline
15000 & Dropconnect & \makecell{0.574 \\ {\scriptsize (0.021)}} 
    & \makecell{0.575 \\ {\scriptsize (0.023)}}  
    & \makecell{0.575 \\ {\scriptsize (0.021)}} 
    & \makecell{0.565 \\ {\scriptsize (0.023)}} 
    & \makecell{0.004 \\ {\scriptsize (0.000)}} \\
    & Dropout & \makecell{0.769 \\ {\scriptsize (0.014)}} 
    & \makecell{0.770 \\ {\scriptsize (0.015)}}  
    & \makecell{0.771 \\ {\scriptsize (0.015)}} 
    & \makecell{0.768 \\ {\scriptsize (0.015)}} 
    & \makecell{0.002 \\ {\scriptsize (0.000)}} \\
    & Ising & \makecell{0.642 \\ {\scriptsize (0.114)}} 
    & \makecell{0.642 \\ {\scriptsize (0.112)}}  
    & \makecell{0.639 \\ {\scriptsize (0.119)}} 
    & \makecell{0.633 \\ {\scriptsize (0.120)}} 
    & \makecell{0.004 \\ {\scriptsize (0.001)}} \\
\hline
\end{tabular}
\caption{Cifar100 - 0.1: Simulation results to investigate the performance of different training set sizes using accuracy, recall, precision, F1-score, and false positive rate}
\label{cifar100_results_0.1}
\end{table}

\begin{table}[htbp]
\centering
\begin{tabular}{ccccccc}
\hline
Training Sample Size & \textbf{Method} & $\mathbf{Accuracy}$ & $\mathbf{Recall}$ & $\mathbf{Precision}$ & $\mathbf{F_{1}}$ & $\mathbf{FPR}$ \\
\hline
250 & Dropconnect & \makecell{0.029 \\ {\scriptsize (0.002)}} 
    & \makecell{0.029 \\ {\scriptsize (0.003)}}  
    & \makecell{0.006 \\ {\scriptsize (0.002)}} 
    & \makecell{0.007 \\ {\scriptsize (0.002)}} 
    & \makecell{0.010 \\ {\scriptsize (0.000)}} \\
    
    & Dropout & \makecell{0.018 \\ {\scriptsize (0.004)}} 
    & \makecell{0.018 \\ {\scriptsize (0.004)}}  
    & \makecell{0.004 \\ {\scriptsize (0.003)}} 
    & \makecell{0.004 \\ {\scriptsize (0.001)}} 
    & \makecell{0.010 \\ {\scriptsize (0.000)}} \\
    
    & Ising & \makecell{0.034 \\ {\scriptsize (0.003)}} 
    & \makecell{0.034 \\ {\scriptsize (0.003)}}  
    & \makecell{0.009 \\ {\scriptsize (0.003)}} 
    & \makecell{0.011 \\ {\scriptsize (0.002)}} 
    & \makecell{0.010 \\ {\scriptsize (0.000)}} \\
\hline
5000 & Dropconnect & \makecell{0.116 \\ {\scriptsize (0.003)}} 
    & \makecell{0.117 \\ {\scriptsize (0.002)}}  
    & \makecell{0.091 \\ {\scriptsize (0.009)}} 
    & \makecell{0.081 \\ {\scriptsize (0.003)}} 
    & \makecell{0.009 \\ {\scriptsize (0.000)}} \\
    & Dropout & \makecell{0.069 \\ {\scriptsize (0.033)}} 
    & \makecell{0.069 \\ {\scriptsize (0.033)}}  
    & \makecell{0.051 \\ {\scriptsize (0.029)}} 
    & \makecell{0.046 \\ {\scriptsize (0.027)}} 
    & \makecell{0.009 \\ {\scriptsize (0.000)}} \\
    & Ising & \makecell{0.134 \\ {\scriptsize (0.004)}} 
    & \makecell{0.135 \\ {\scriptsize (0.005)}}  
    & \makecell{0.119 \\ {\scriptsize (0.008)}} 
    & \makecell{0.104 \\ {\scriptsize (0.006)}} 
    & \makecell{0.009 \\ {\scriptsize (0.000)}} \\
\hline
15000 & Dropconnect & \makecell{0.196 \\ {\scriptsize (0.008)}} 
    & \makecell{0.196 \\ {\scriptsize (0.006)}}  
    & \makecell{0.176 \\ {\scriptsize (0.014)}} 
    & \makecell{0.154 \\ {\scriptsize (0.006)}} 
    & \makecell{0.008 \\ {\scriptsize (0.000)}} \\
    & Dropout & \makecell{0.300 \\ {\scriptsize (0.065)}} 
    & \makecell{0.302 \\ {\scriptsize (0.063)}}  
    & \makecell{0.292 \\ {\scriptsize (0.068)}} 
    & \makecell{0.278 \\ {\scriptsize (0.068)}} 
    & \makecell{0.007 \\ {\scriptsize (0.001)}} \\
    & Ising & \makecell{0.478 \\ {\scriptsize (0.023)}} 
    & \makecell{0.477 \\ {\scriptsize (0.027)}}  
    & \makecell{0.483 \\ {\scriptsize (0.025)}} 
    & \makecell{0.452 \\ {\scriptsize (0.026)}} 
    & \makecell{0.005 \\ {\scriptsize (0.000)}} \\
\hline
\end{tabular}
\caption{Cifar100 - 0.5: Simulation results to investigate the performance of different training set sizes using accuracy, recall, precision, F1-score, and false positive rate}
\label{cifar100_results_0.5}
\end{table}

\section{Conclusion and discussion}\label{conc}

This paper introduces a novel likelihood-guided variational Ising-based regularization framework for attention-based architectures that unifies model sparsification, uncertainty estimation, and calibration within a single probabilistic formulation. The proposed approach derives posterior dropout probabilities directly from data by marginalizing the effect of individual weights on the predictive loss, thereby eliminating the need for manually tuned stochasticity parameters and enabling the derivation of a posterior over network connections as well as traceback of the activating input. This likelihood-guided approach extends Bayesian inference principles to structured dropout and provides a theoretically grounded mechanism for adaptive regularization in deep neural networks.

Experimental evaluations on MNIST, FashionMNIST, CIFAR-10, and CIFAR-100 demonstrate that the Ising model achieves comparable performance to established regularization methods while producing well structured entropy distributions, indicating good probability calibration. These results indicate that the method effectively captures both epistemic and aleatoric sources of uncertainty while preserving representational flexibility. Moreover, the posterior over dropout masks induces structured sparsity, allowing efficient pruning and interpretability without auxiliary regularization or post hoc thresholding.

From a theoretical perspective, the proposed framework links local sensitivity analysis to variational Bayesian learning by quantifying the predictive impact of weight perturbations through loss differentials. This connection grounds model regularization in the geometry of the likelihood landscape rather than simple heuristic noise injection. In conclusion, the Ising regularization framework represents a principled step toward data-adaptive and interpretable uncertainty modeling in deep networks. Future research will focus on mitigating the observed conservatism of the Ising formulation by exploring alternative prior structures and weighting schemes that allow the posterior to express sharper yet well-calibrated confidence under varying data regimes.

\bibliographystyle{plainnat}
\bibliography{references}

\clearpage
\appendix

\setcounter{figure}{0}            
\renewcommand{\thefigure}{S\arabic{figure}}
\section*{Supplementary Materials}
The supplementary material is organized as follows. Section S1 presents the visualizations of four benchmark datasets, Section S2 reports the calibration curves, and Section S3 summarizes the entropy distributions.

\subsection*{S1. Four benchmark datasets: MNIST, FashionMNIST, CIFAR-10, and CIFAR-100}

The visualizations of MNIST, FashionMNIST, CIFAR-10, and CIFAR-100 are displayed in \ref{MNIST_viz}-\ref{Cifar100_viz}.

\begin{center}
\hspace*{-1cm}
\includegraphics[scale=0.4]{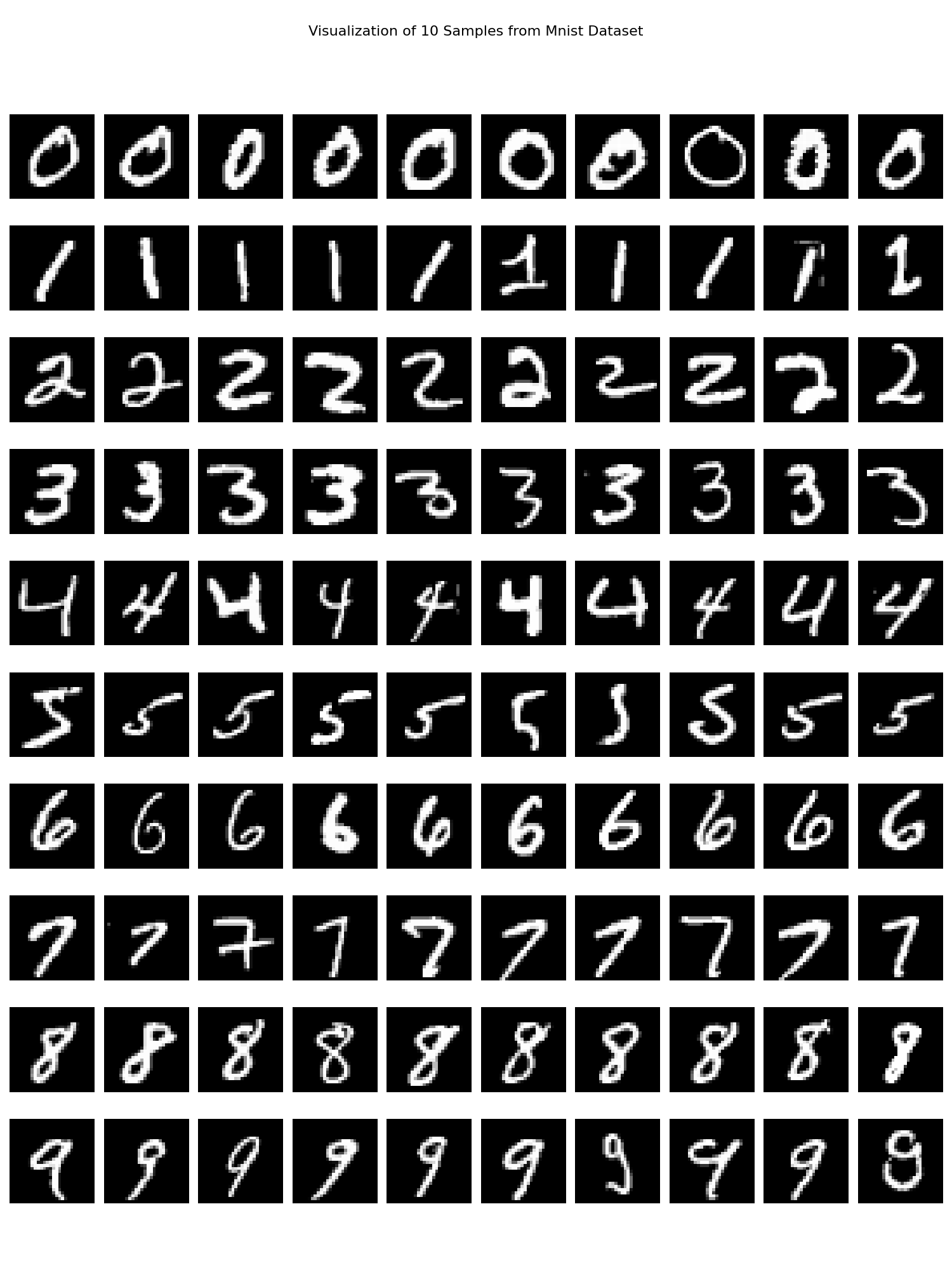}	
\captionof{figure}{Visualization of random samples from the MNIST dataset.}
\label{MNIST_viz}
\end{center}

\begin{center}
\hspace*{-1cm}
\includegraphics[scale=0.4]{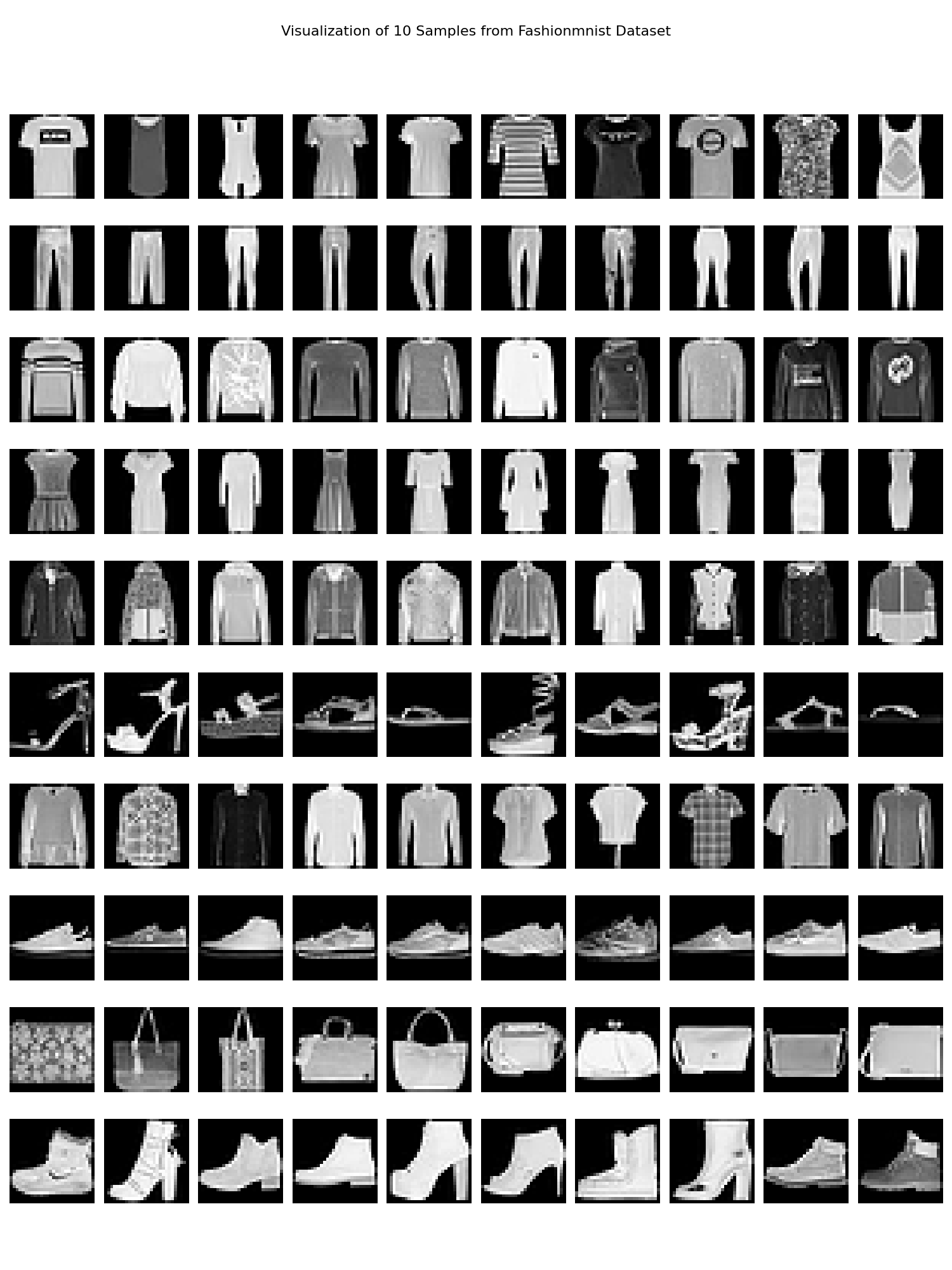}	
\captionof{figure}{Visualization of random samples from the Fashion-MNIST dataset.}
\label{FashionMNIST_viz}
\end{center}

\begin{center}
\hspace*{-1cm}
\includegraphics[scale=0.4]{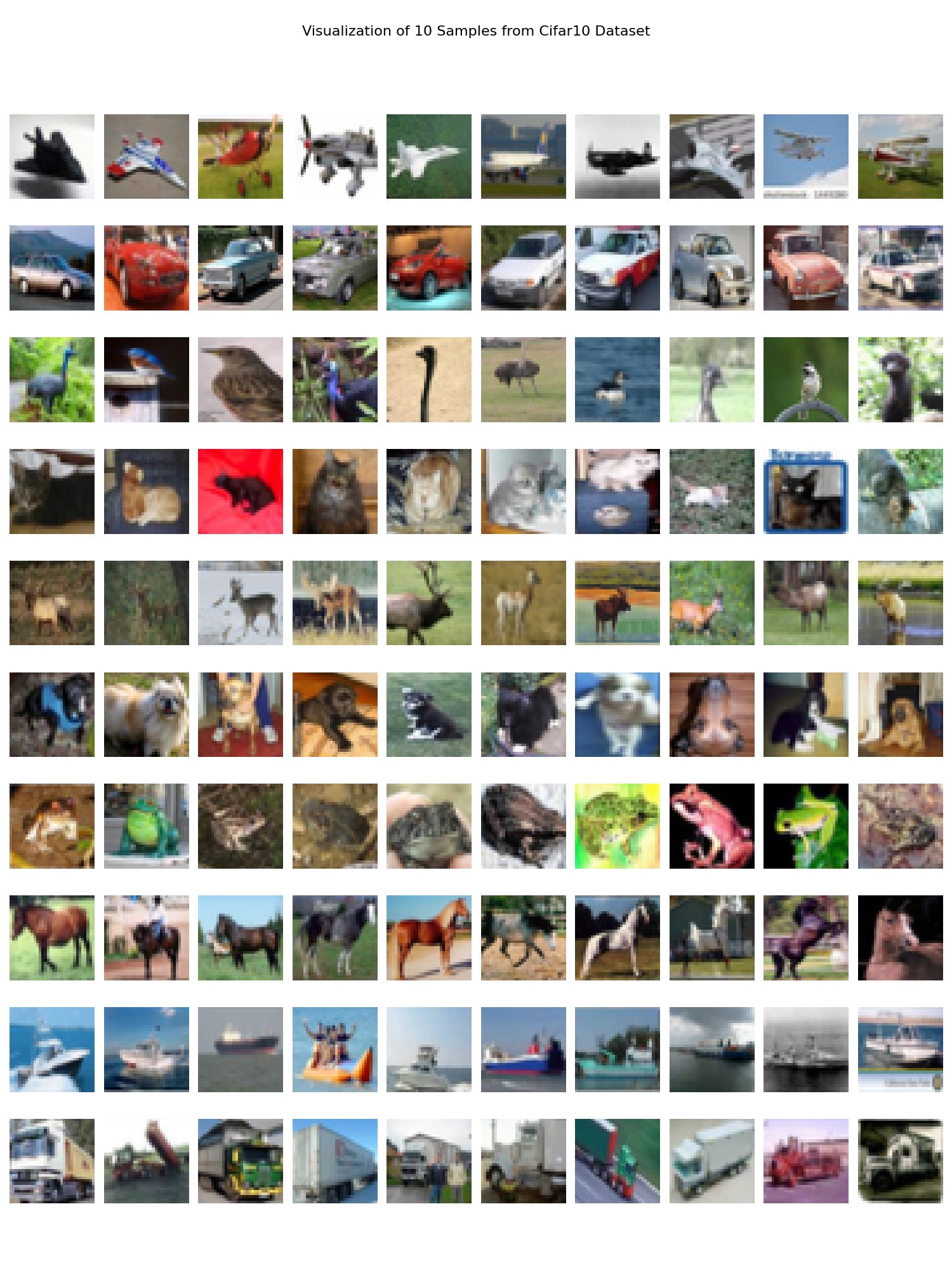}	
\captionof{figure}{Visualization of random samples from the Cifar10 dataset.}
\label{Cifar10_viz}
\end{center}

\begin{center}
\hspace*{-1cm}
\includegraphics[scale=0.4]{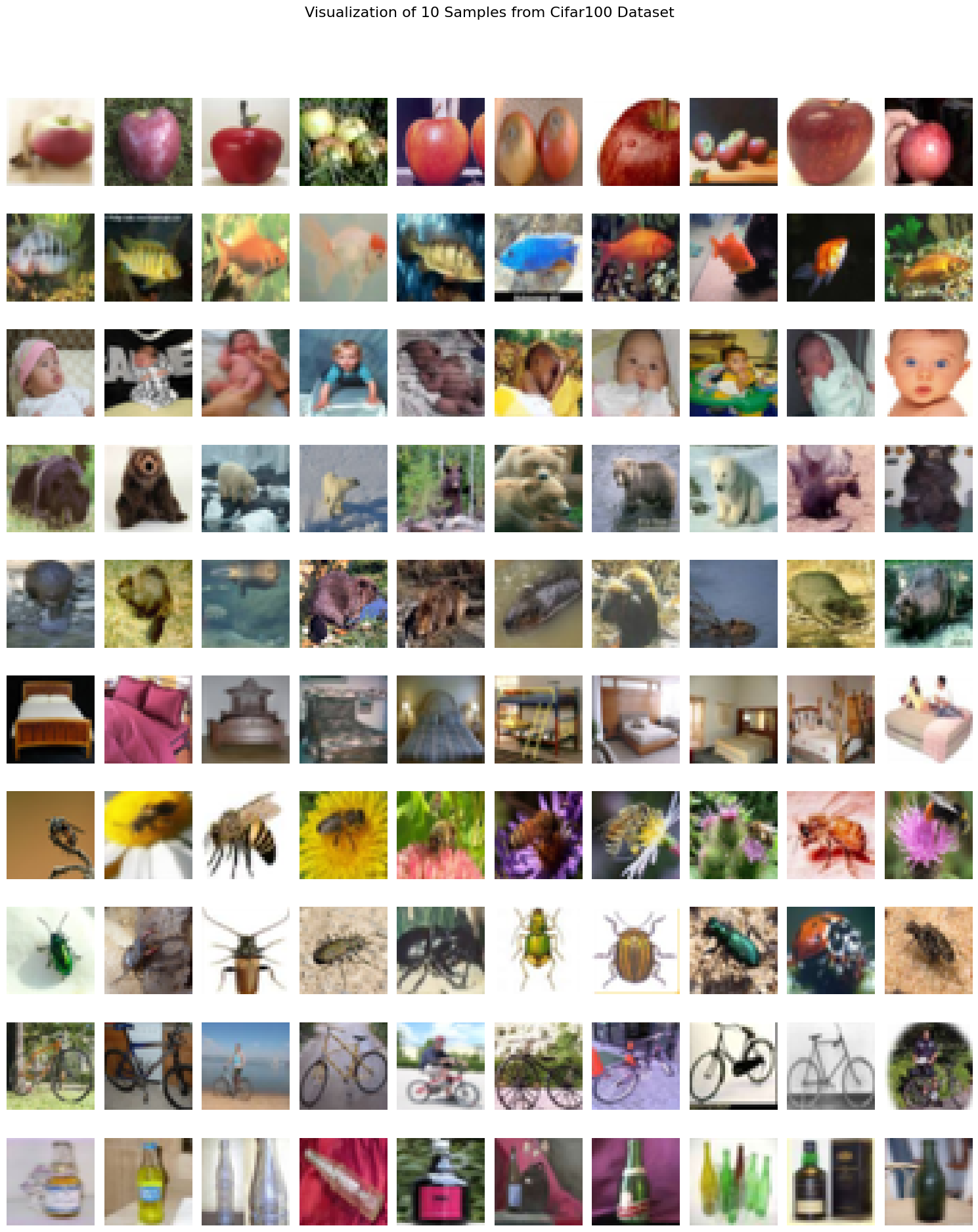}	
\captionof{figure}{Visualization of random samples from the Cifar100 dataset.}
\label{Cifar100_viz}
\end{center}

\subsection*{S2. Calibration Curves}
Calibration curves across datasets and regularization strategies, complementing entropy-based uncertainty profiles, are shown in Figures \ref{mnist_calibration_0.1_val0.6}-\ref{cifar100_calibration_0.5_val0.895}.


\begin{center}
\hspace*{-1cm}
\includegraphics[scale=0.6]{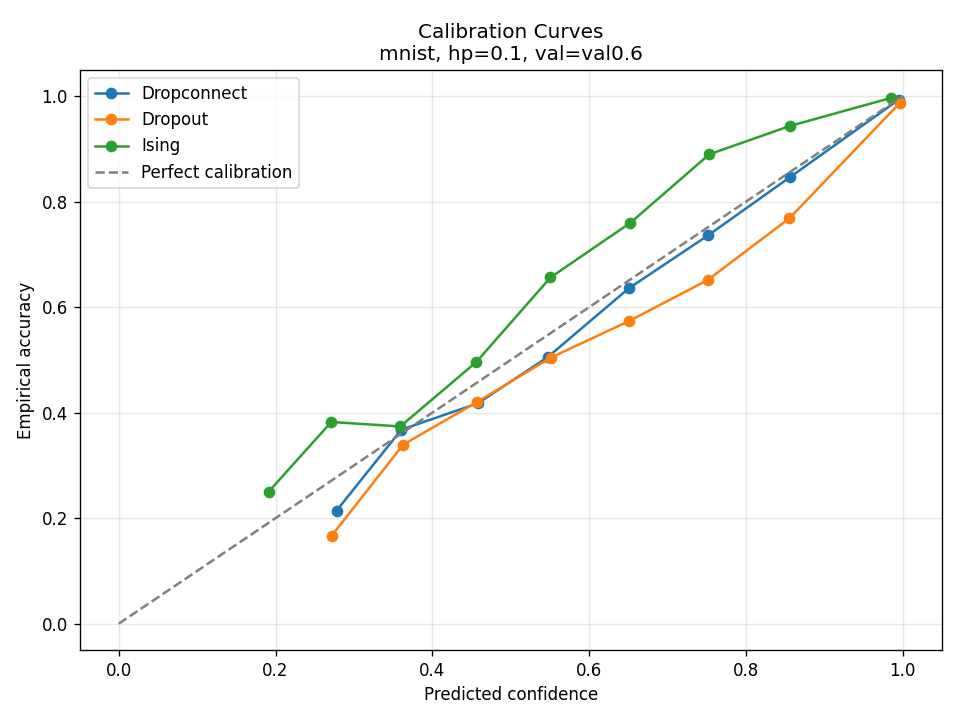}	
\captionof{figure}{Comparison of probability calibration for the three methods examined in this paper. The diagonal line represents perfect calibration; being under the line indicates overconfidence, while being above the line indicates an overly conservative model. The setting for this experiment is 18000 training samples, a regularization hyperparameter of 0.1, and a fixed test dataset size of 6000 from the MNIST dataset.}
\label{mnist_calibration_0.1_val0.6}
\end{center}

\begin{center}
\hspace*{-1cm}
\includegraphics[scale=0.6]{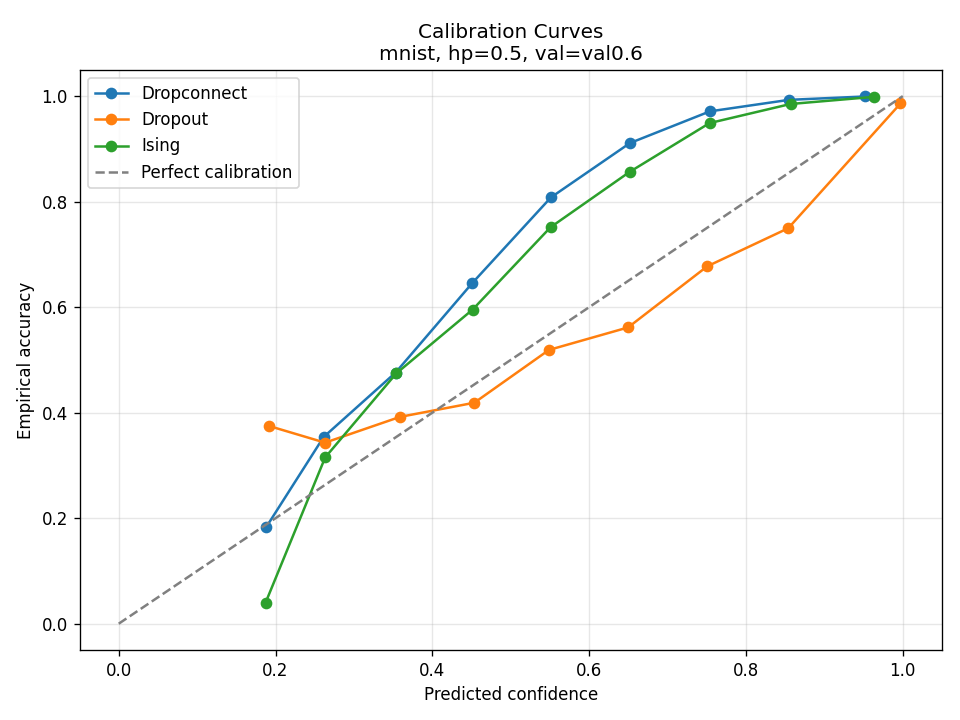}	
\captionof{figure}{Comparison of probability calibration for the three methods examined in this paper. The diagonal line represents perfect calibration; being under the line indicates overconfidence, while being above the line indicates an overly conservative model. The setting for this experiment is 18000 training samples, a regularization hyperparameter of 0.5, and a fixed test dataset size of 6000 from the MNIST dataset.}
\label{mnist_calibration_0.5_val0.6}
\end{center}

\begin{center}
\hspace*{-1cm}
\includegraphics[scale=0.6]{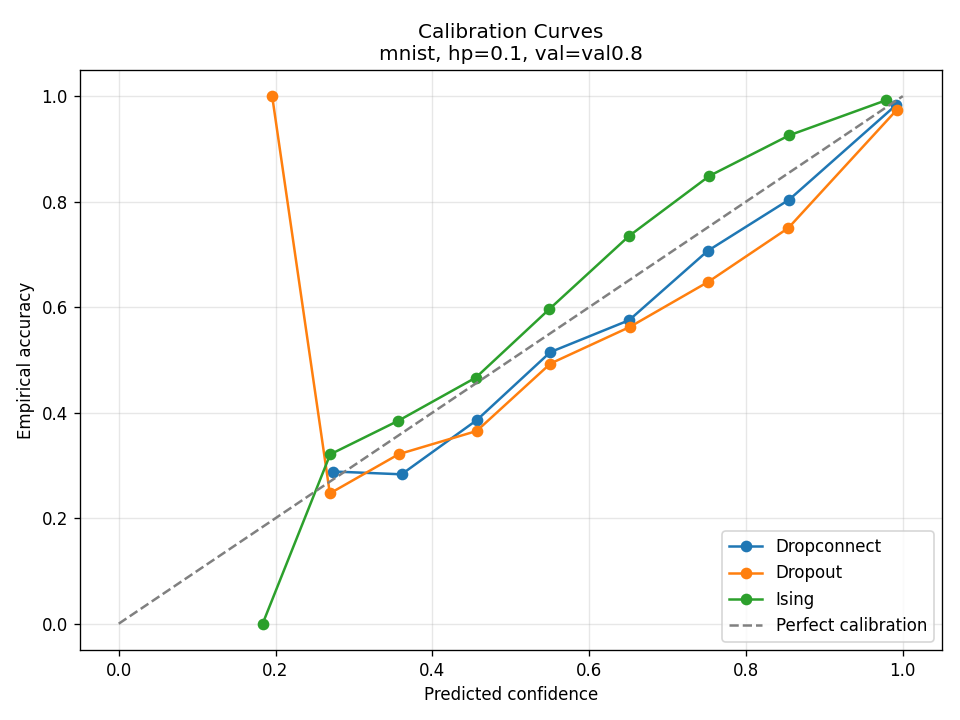}	
\captionof{figure}{Comparison of probability calibration for the three methods examined in this paper. The diagonal line represents perfect calibration; being under the line indicates overconfidence, while being above the line indicates an overly conservative model. The setting for this experiment is 6000 training samples, a 0.1 regularization hyperparameter, and a fixed testing dataset size of 6000 from the MNIST dataset.}
\label{mnist_calibration_0.1_val0.8}
\end{center}

\begin{center}
\hspace*{-1cm}
\includegraphics[scale=0.6]{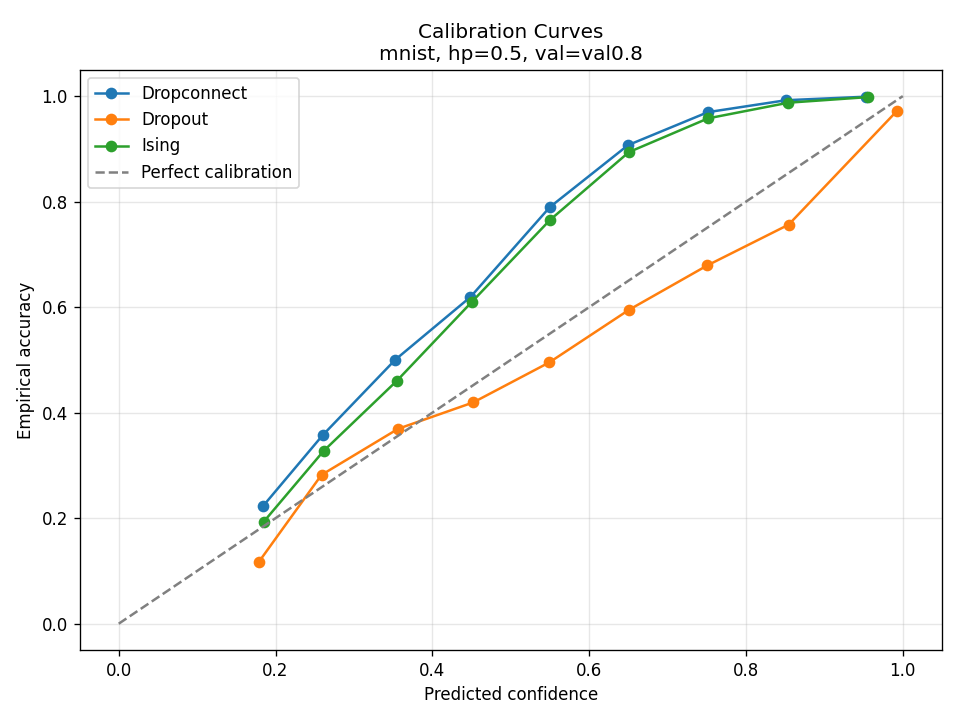}	
\captionof{figure}{Comparison of probability calibration for the three methods examined in this paper. The diagonal line represents perfect calibration; being under the line indicates overconfidence, while being above the line indicates an overly conservative model. The setting for this experiment is 6000 training samples, a regularization hyperparameter of 0.5, and a fixed test dataset size of 6000 from the MNIST dataset.}
\label{mnist_calibration_0.5_val0.8}
\end{center}

\begin{center}
\hspace*{-1cm}
\includegraphics[scale=0.6]{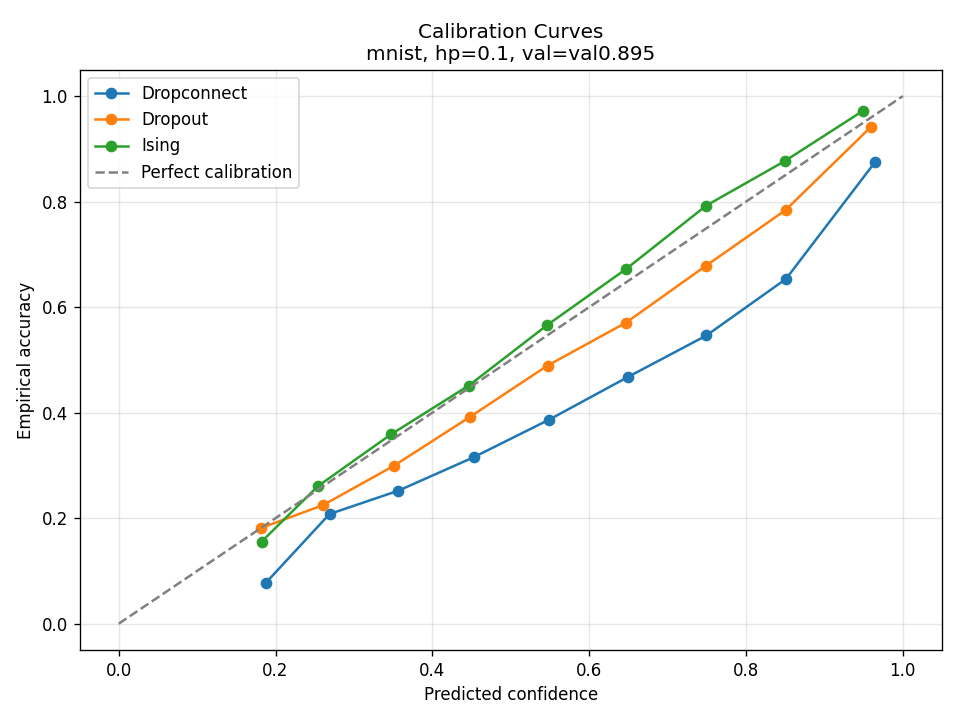}	
\captionof{figure}{Comparison of probability calibration for the three methods examined in this paper. The diagonal line represents perfect calibration; being under the line indicates overconfidence, while being above the line indicates an overly conservative model. The setting for this experiment is 300 training samples, a regularization hyperparameter of 0.1, and a fixed test dataset size of 6000 from the MNIST dataset.}
\label{mnist_calibration_0.1_val0.895}
\end{center}

\begin{center}
\hspace*{-1cm}
\includegraphics[scale=0.6]{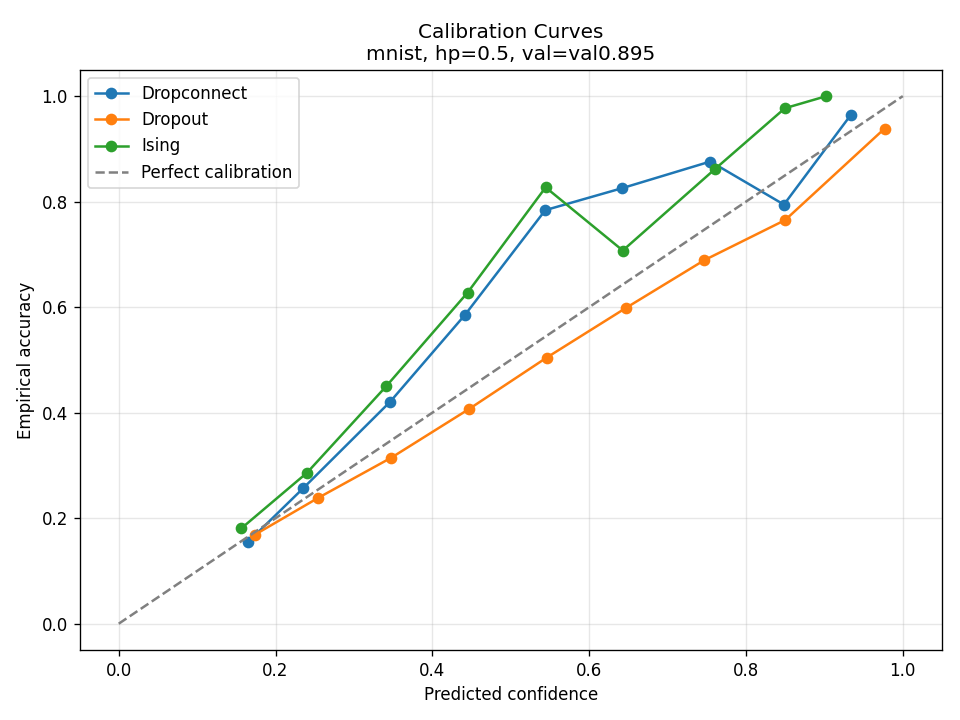}	
\captionof{figure}{Comparison of probability calibration for the three methods examined in this paper. The diagonal line represents perfect calibration; being under the line indicates overconfidence, while being above the line indicates an overly conservative model. The setting for this experiment is 300 training samples, a regularization hyperparameter of 0.5, and a fixed test dataset size of 6000 from the MNIST dataset.}
\label{mnist_calibration_0.5_val0.895}
\end{center}


\begin{center}
\hspace*{-1cm}
\includegraphics[scale=0.6]{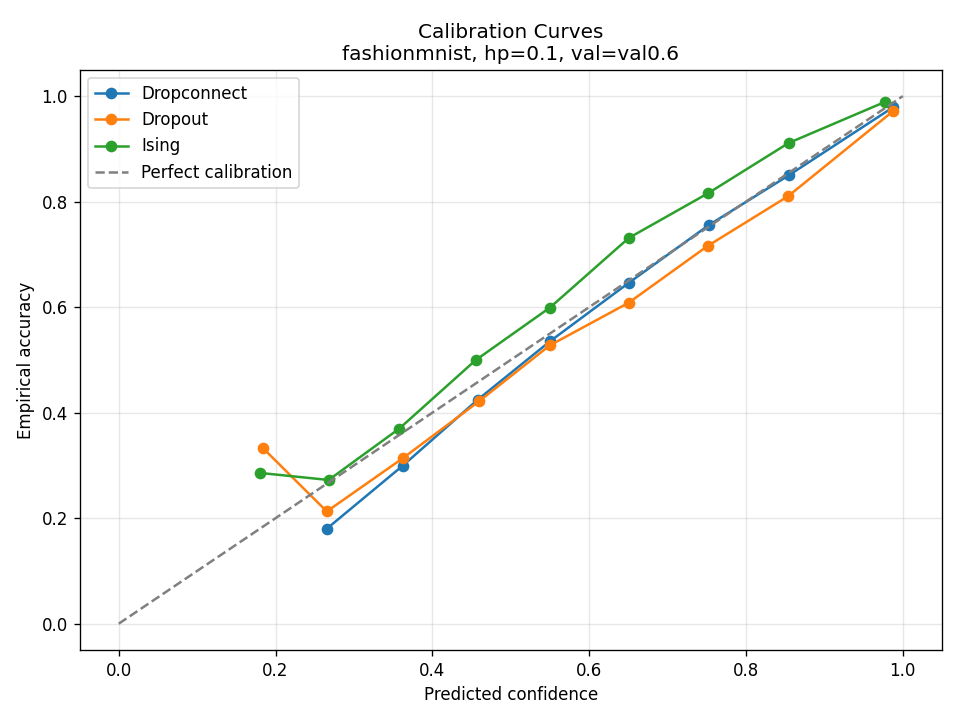}	
\captionof{figure}{Comparison of probability calibration for the three methods examined in this paper. The diagonal line represents perfect calibration; being under the line indicates overconfidence, while being above the line indicates an overly conservative model. The setting for this experiment is 18000 training samples, 0.1 regularization hyperparameter, fixed testing dataset size of 6000 from the Fashion-MNIST dataset.}
\label{fashionmnist_calibration_0.1_val0.6}
\end{center}

\begin{center}
\hspace*{-1cm}
\includegraphics[scale=0.6]{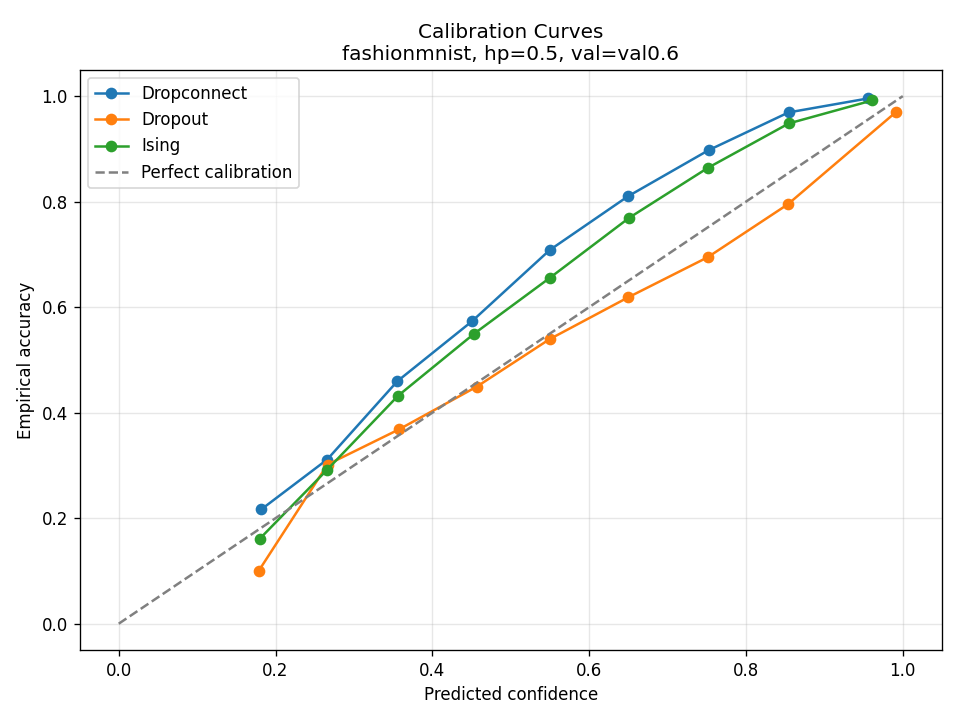}	
\captionof{figure}{Comparison of probability calibration for the three methods examined in this paper. The diagonal line represents perfect calibration; being under the line indicates overconfidence, while being above the line indicates an overly conservative model. The setting for this experiment is 18000 training samples, 0.5 regularization hyperparameter, fixed testing dataset size of 6000 from the Fashion-MNIST dataset.}
\label{fashionmnist_calibration_0.5_val0.6}
\end{center}

\begin{center}
\hspace*{-1cm}
\includegraphics[scale=0.6]{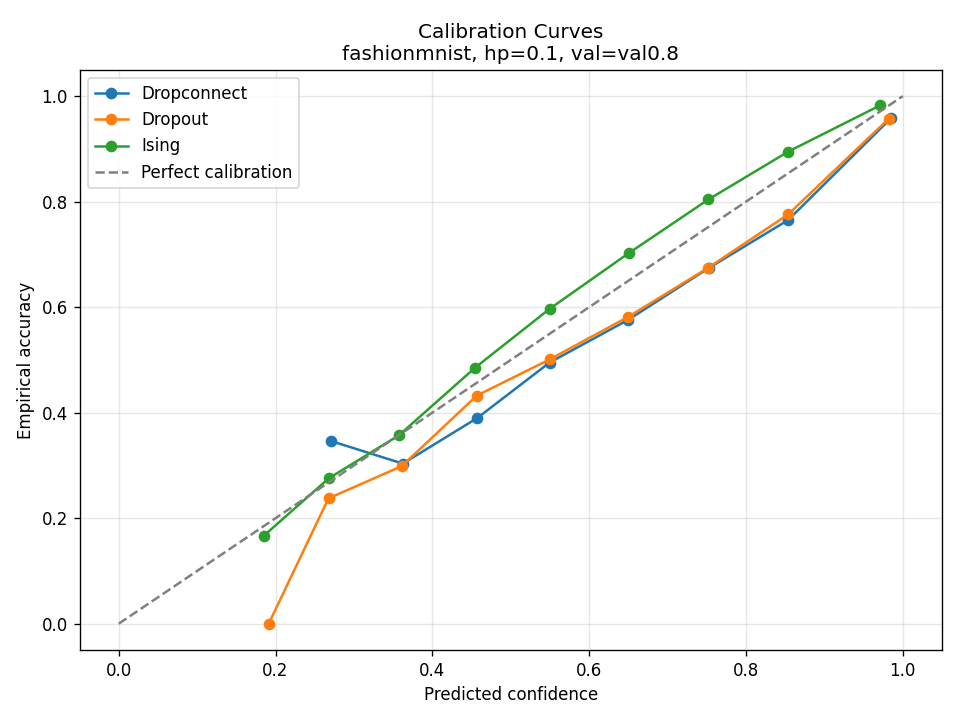}	
\captionof{figure}{Comparison of probability calibration for the three methods examined in this paper. The diagonal line represents perfect calibration; being under the line indicates overconfidence, while being above the line indicates an overly conservative model. The setting for this experiment is 6000 training samples, a 0.1 regularization hyperparameter, and a fixed testing dataset size of 6000 from the Fashion-MNIST dataset.}
\label{fashionmnist_calibration_0.1_val0.8}
\end{center}

\begin{center}
\hspace*{-1cm}
\includegraphics[scale=0.6]{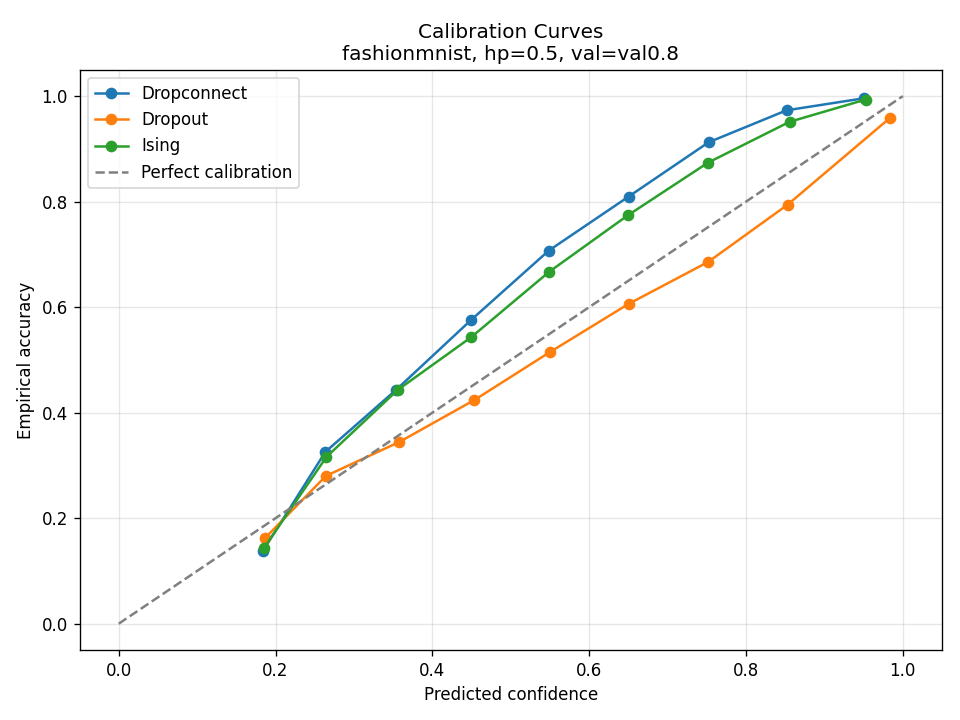}
\captionof{figure}{Comparison of probability calibration for the three methods examined in this paper. The diagonal line represents perfect calibration; being under the line indicates overconfidence, while being above the line indicates an overly conservative model. The setting for this experiment is 6000 training samples, a 0.5 regularization hyperparameter, and a fixed testing dataset size of 6000 from the Fashion-MNIST dataset.}
\label{fashionmnist_calibration_0.5_val0.8}
\end{center}

\begin{center}
\hspace*{-1cm}
\includegraphics[scale=0.6]{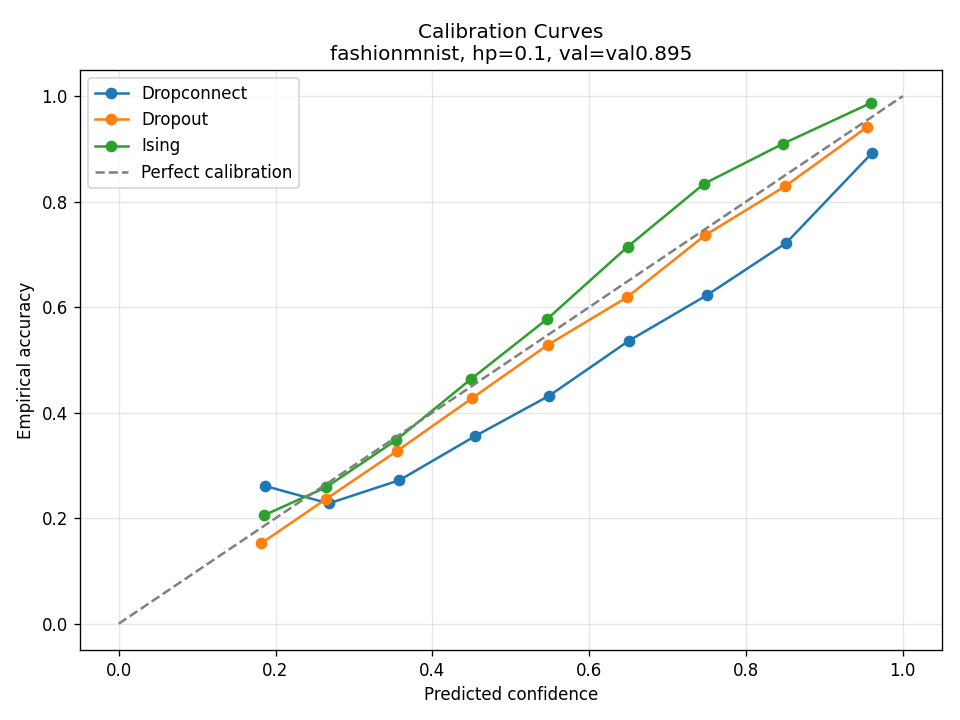}	
\captionof{figure}{Comparison of probability calibration for the three methods examined in this paper. The diagonal line represents perfect calibration; being under the line indicates overconfidence, while being above the line indicates an overly conservative model. The setting for this experiment is 300 training samples, a 0.1 regularization hyperparameter, and a fixed testing dataset size of 6000 from the Fashion-MNIST dataset.}
\label{fashionmnist_calibration_0.1_val0.895}
\end{center}

\begin{center}
\hspace*{-1cm}
\includegraphics[scale=0.6]{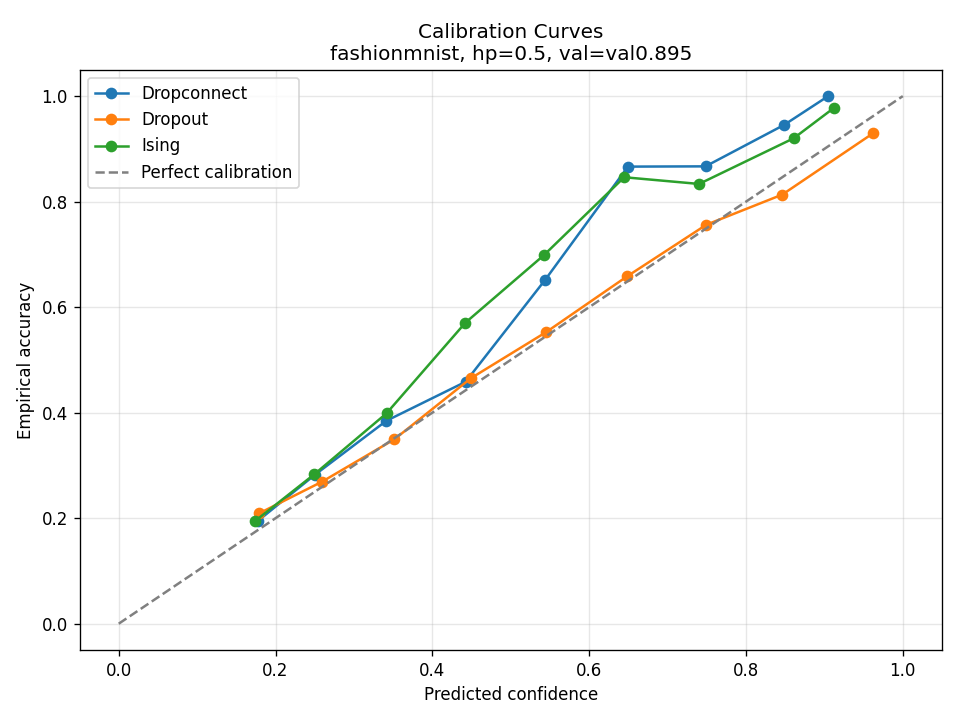}	
\captionof{figure}{Comparison of probability calibration for the three methods examined in this paper. The diagonal line represents perfect calibration; being under the line indicates overconfidence, while being above the line indicates an overly conservative model. The setting for this experiment is 300 training samples, a 0.5 regularization hyperparameter, and a fixed testing dataset size of 6000 from the Fashion-MNIST dataset.}
\label{fashionmnist_calibration_0.5_val0.895}
\end{center}


\begin{center}
\hspace*{-1cm}
\includegraphics[scale=0.6]{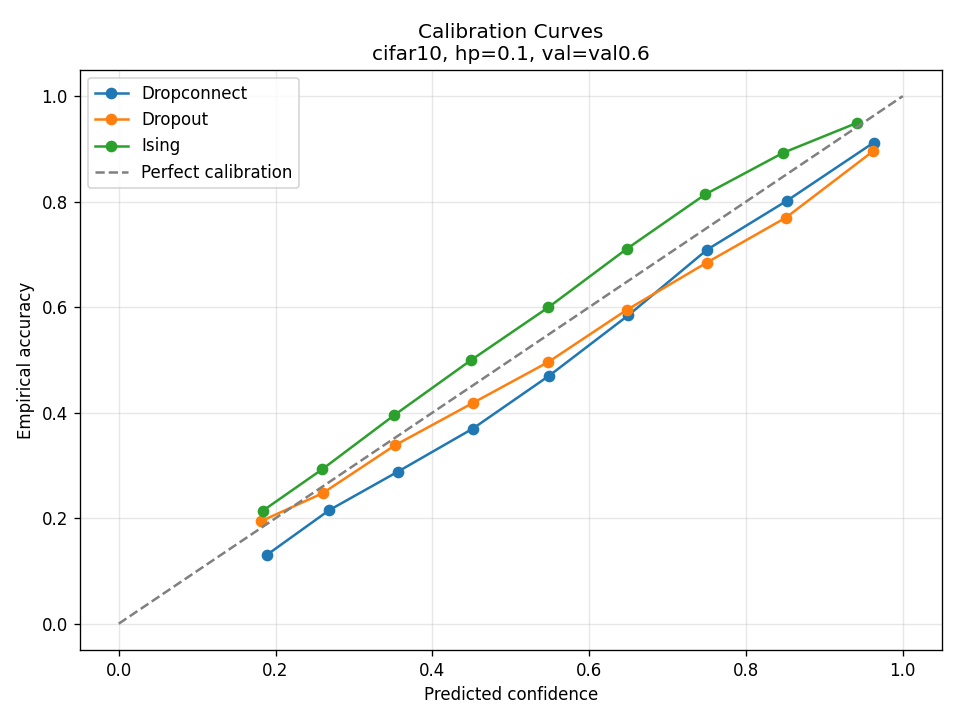}
\captionof{figure}{Comparison of probability calibration for the three methods examined in this paper. The diagonal line represents perfect calibration; being under the line indicates overconfidence, while being above the line indicates an overly conservative model. The setting for this experiment is 15000 training samples, a regularization hyperparameter of 0.1, and a fixed test dataset size of 5000 from the CIFAR-10 dataset.}
\label{cifar10_calibration_0.1_val0.6}
\end{center}

\begin{center}
\hspace*{-1cm}
\includegraphics[scale=0.6]{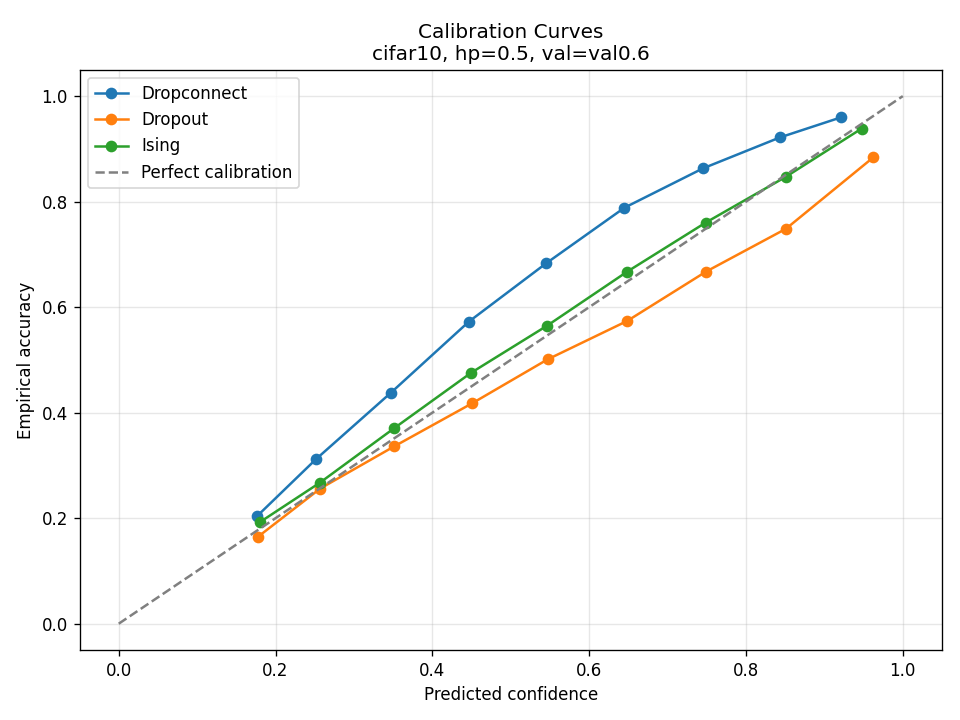}	
\captionof{figure}{Comparison of probability calibration for the three methods examined in this paper. The diagonal line represents perfect calibration; being under the line indicates overconfidence, while being above the line indicates an overly conservative model. The setting for this experiment is 15000 training samples, a regularization hyperparameter of 0.5, and a fixed test dataset size of 5000 from the CIFAR-10 dataset.}
\label{cifar10_calibration_0.5_val0.6}
\end{center}

\begin{center}
\hspace*{-1cm}
\includegraphics[scale=0.6]{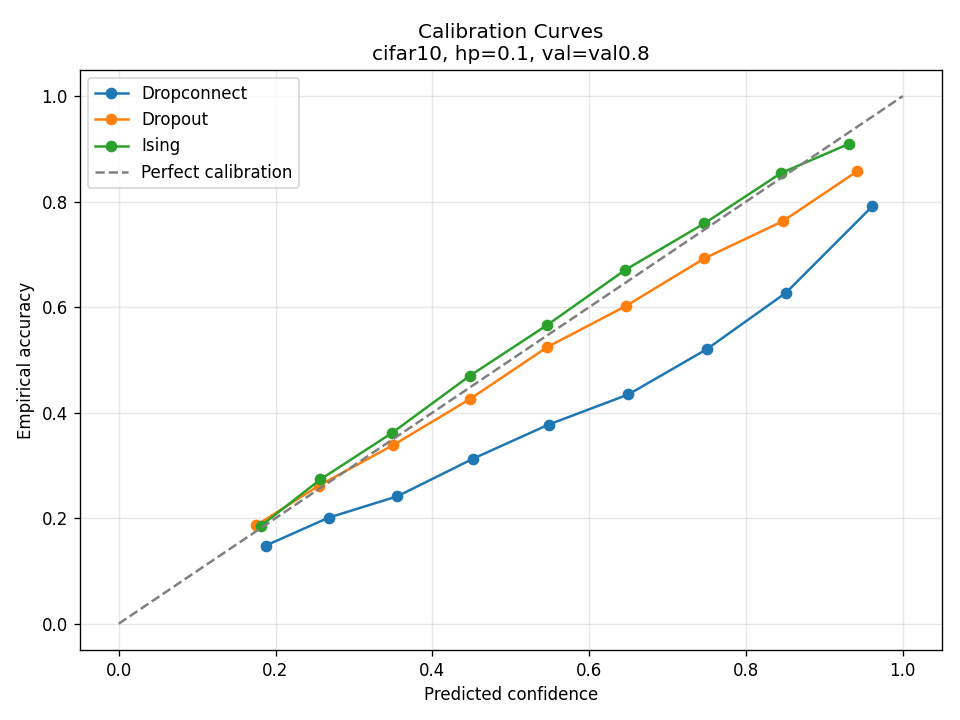}
\captionof{figure}{Comparison of probability calibration for the three methods examined in this paper. The diagonal line represents perfect calibration; being under the line indicates overconfidence, while being above the line indicates an overly conservative model. The setting for this experiment is 5000 training samples, 0.1 regularization hyperparameter, fixed testing dataset size of 5000 from the Cifar10 dataset.}
\label{cifar10_calibration_0.1_val0.8}
\end{center}

\begin{center}
\hspace*{-1cm}
\includegraphics[scale=0.6]{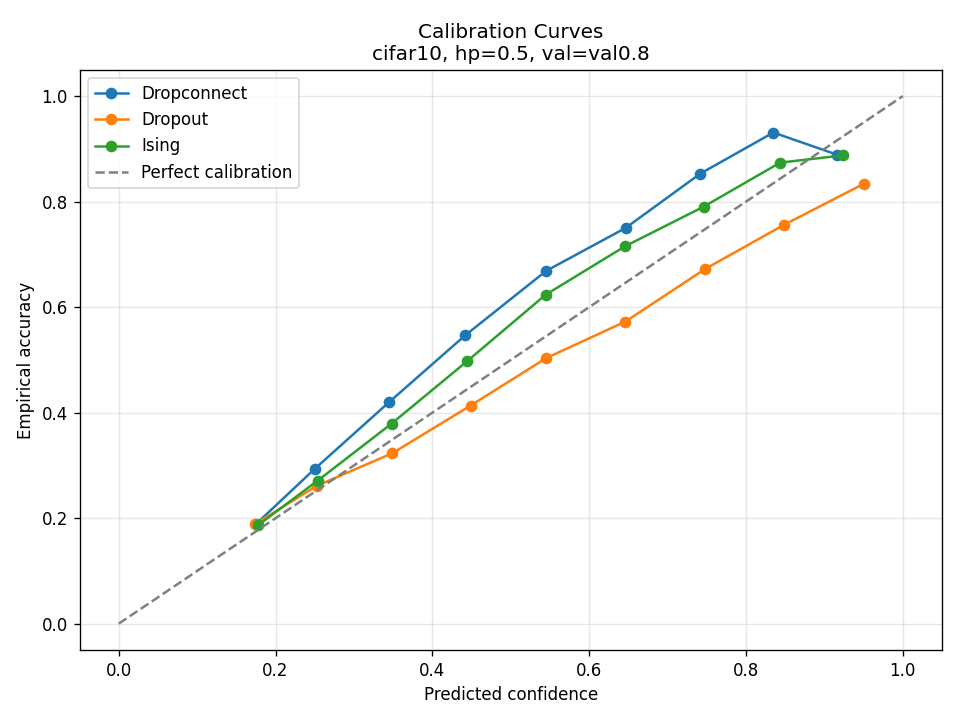}	
\captionof{figure}{Comparison of probability calibration for the three methods examined in this paper. The diagonal line represents perfect calibration; being under the line indicates overconfidence, while being above the line indicates an overly conservative model. The setting for this experiment is 5000 training samples, 0.5 regularization hyperparameter, fixed testing dataset size of 5000 from the Cifar10 dataset.}
\label{cifar10_calibration_0.5_val0.8}
\end{center}

\begin{center}
\hspace*{-1cm}
\includegraphics[scale=0.6]{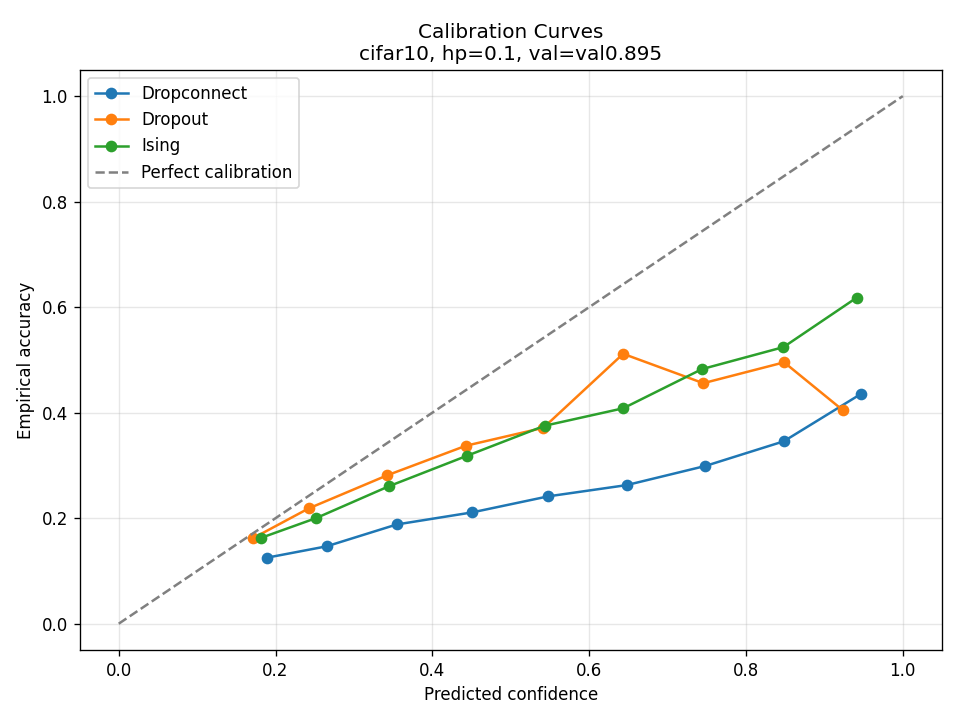}
\captionof{figure}{Comparison of probability calibration for the three methods examined in this paper. The diagonal line represents perfect calibration; being under the line indicates overconfidence, while being above the line indicates an overly conservative model. The setting for this experiment is 250 training samples, 0.1 regularization hyperparameter, fixed testing dataset size of 5000 from the Cifar10 dataset.}
\label{cifar10_calibration_0.1_val0.895}
\end{center}

\begin{center}
\hspace*{-1cm}
\includegraphics[scale=0.6]{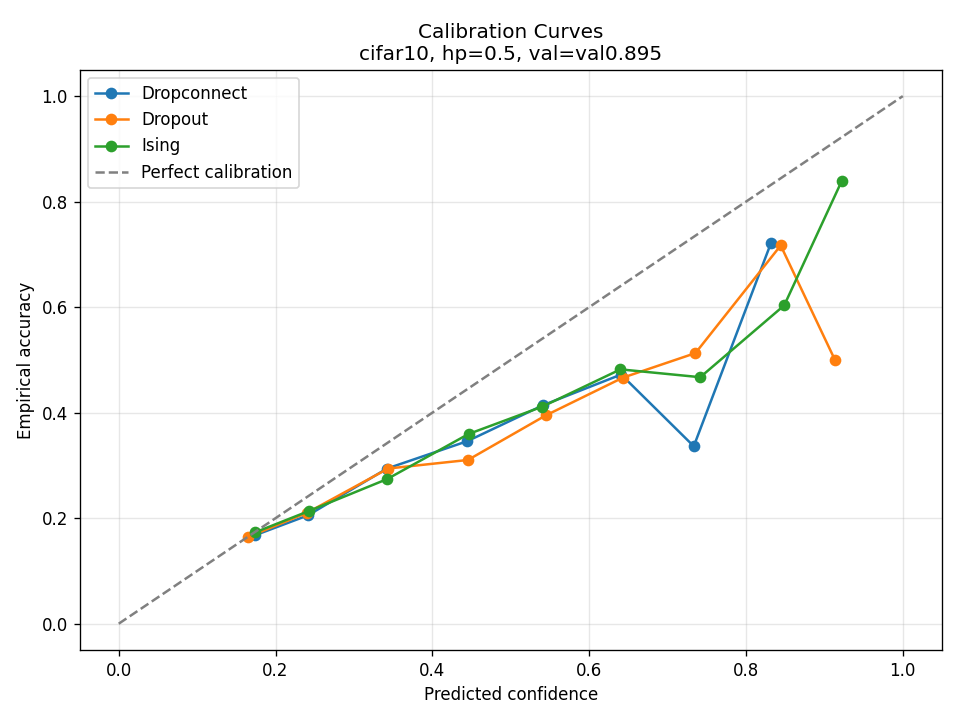}	
\captionof{figure}{Comparison of probability calibration for the three methods examined in this paper. The diagonal line represents perfect calibration; being under the line indicates overconfidence, while being above the line indicates an overly conservative model. The setting for this experiment is 250 training samples, a 0.5 regularization hyperparameter, and a fixed testing dataset size of 5000 from the Cifar10 dataset.}
\label{cifar10_calibration_0.5_val0.895}
\end{center}


\begin{center}
\hspace*{-1cm}
\includegraphics[scale=0.6]{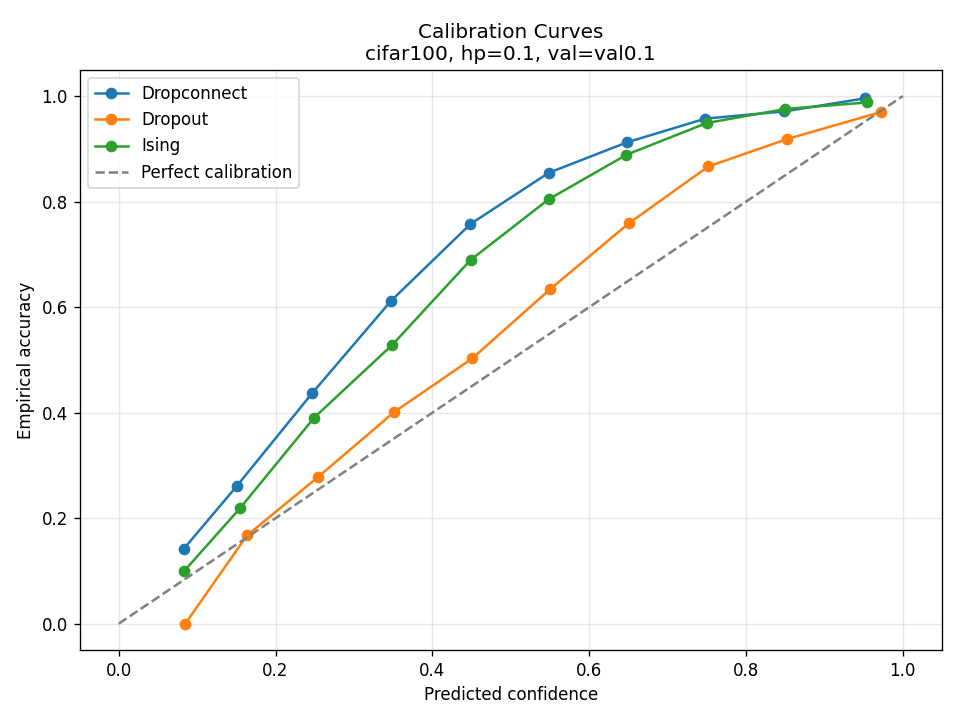}
\captionof{figure}{Comparison of probability calibration for the three methods examined in this paper. The diagonal line represents perfect calibration; being under the line indicates overconfidence, while being above the line indicates an overly conservative model. The setting for this experiment is 40000 training samples, 0.1 regularization hyperparameter, fixed testing dataset size of 5000 from the Cifar100 dataset.}
\label{cifar100_calibration_0.1_val0.1}
\end{center}

\begin{center}
\hspace*{-1cm}
\includegraphics[scale=0.6]{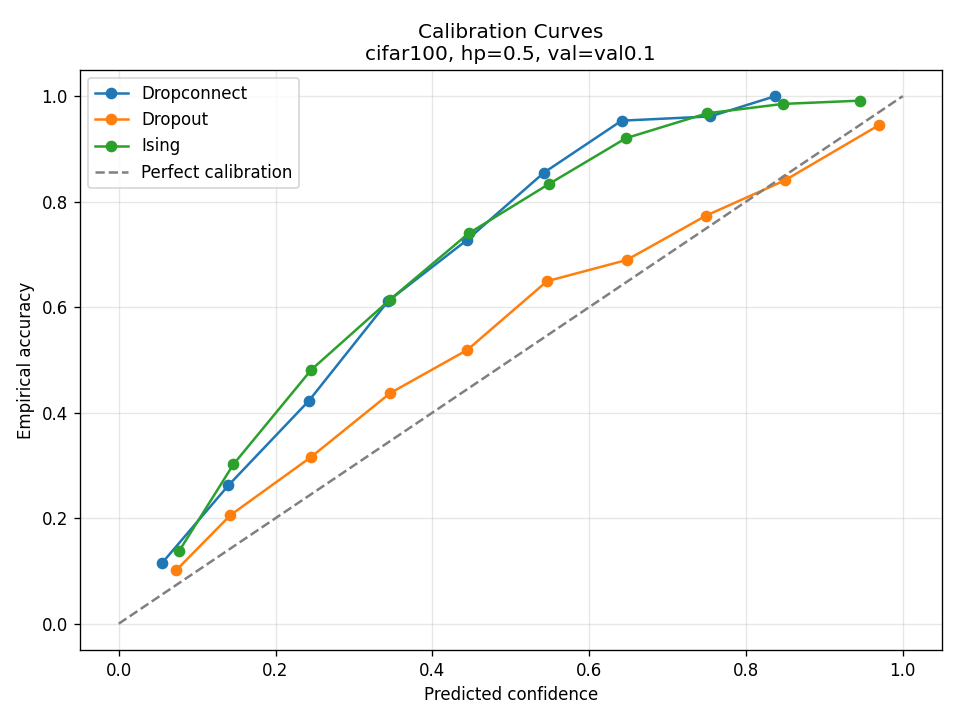}	
\captionof{figure}{Comparison of probability calibration for the three methods examined in this paper. The diagonal line represents perfect calibration; being under the line indicates overconfidence, while being above the line indicates an overly conservative model. The setting for this experiment is 40000 training samples, 0.5 regularization hyperparameter, fixed testing dataset size of 5000 from the Cifar100 dataset.}
\label{cifar100_calibration_0.5_val0.1}
\end{center}

\begin{center}
\hspace*{-1cm}
\includegraphics[scale=0.6]{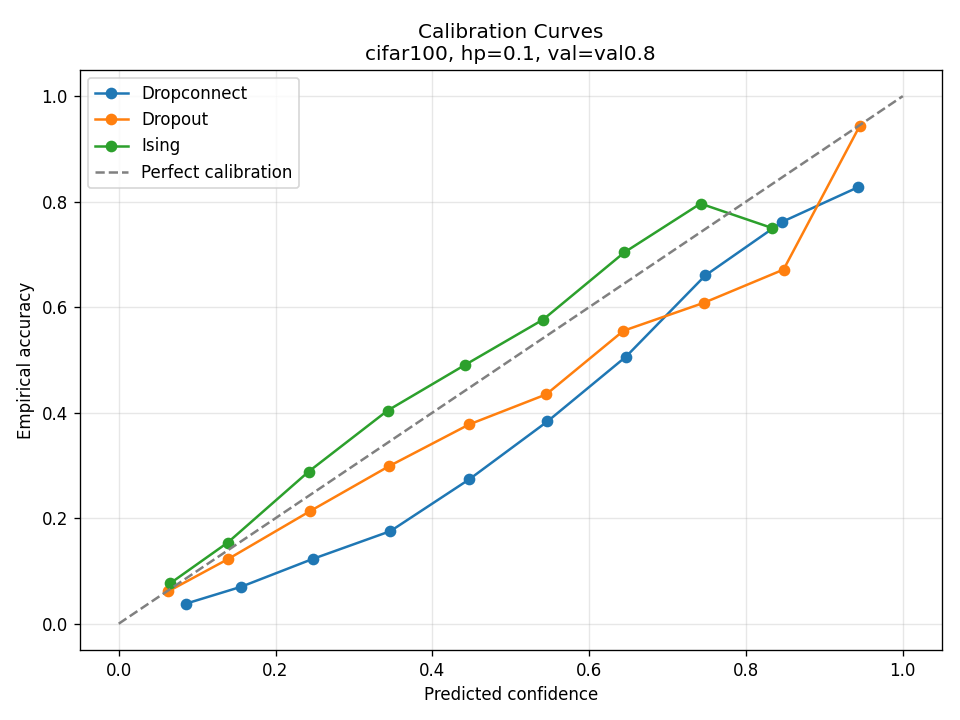}
\captionof{figure}{Comparison of probability calibration for the three methods examined in this paper. The diagonal line represents perfect calibration; being under the line indicates overconfidence, while being above the line indicates an overly conservative model. The setting for this experiment is 15000 training samples, 0.1 regularization hyperparameter, fixed testing dataset size of 5000 from the Cifar100 dataset.}
\label{cifar100_calibration_0.1_val0.8}
\end{center}

\begin{center}
\hspace*{-1cm}
\includegraphics[scale=0.6]{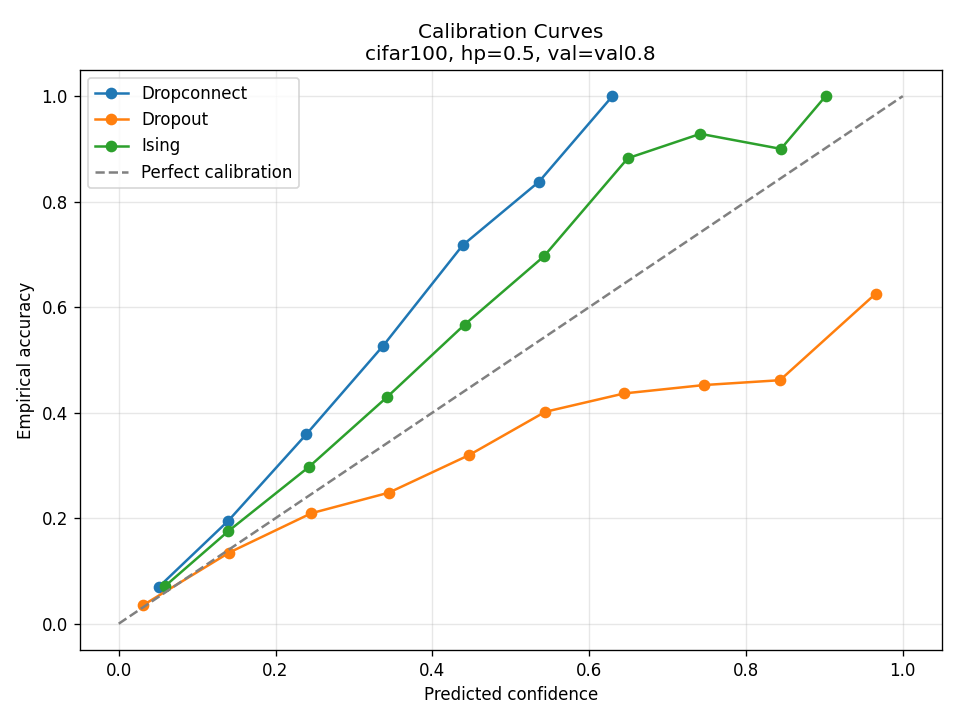}	
\captionof{figure}{Comparison of probability calibration for the three methods examined in this paper. The diagonal line represents perfect calibration; being under the line indicates overconfidence, while being above the line indicates an overly conservative model. The setting for this experiment is 15000 training samples, 0.5 regularization hyperparameter, fixed testing dataset size of 5000 from the Cifar100 dataset.}
\label{cifar100_calibration_0.5_val0.8}
\end{center}

\begin{center}
\hspace*{-1cm}
\includegraphics[scale=0.6]{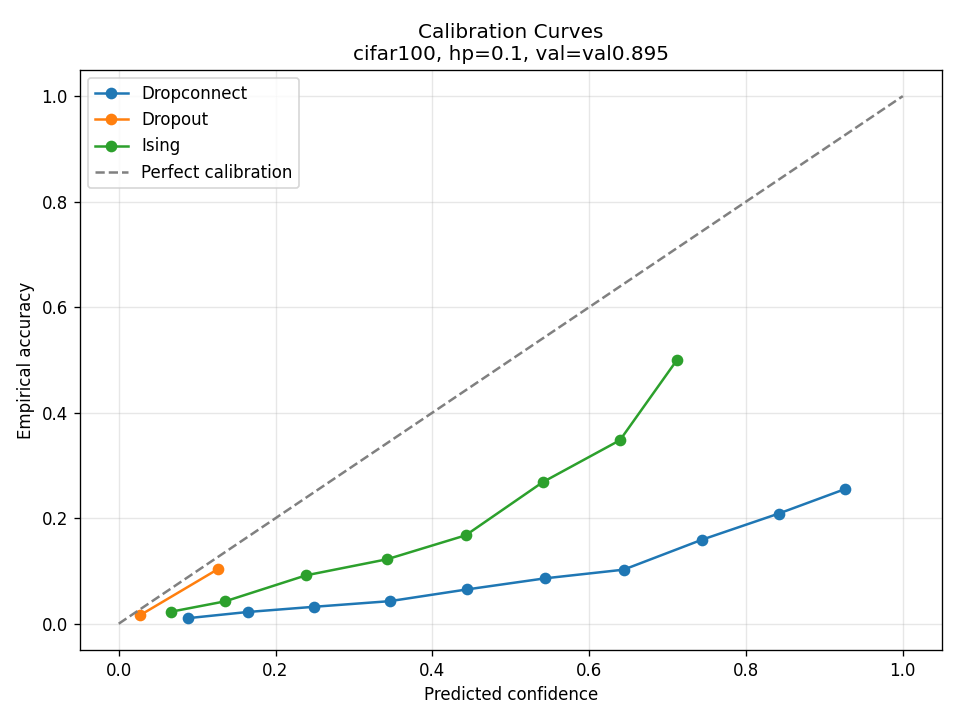}	
\captionof{figure}{Comparison of probability calibration for the three methods examined in this paper. The diagonal line represents perfect calibration; being under the line indicates overconfidence, while being above the line indicates an overly conservative model. The setting for this experiment is 250 training samples, 0.1 regularization hyperparameter, fixed testing dataset size of 5000 from the Cifar100 dataset.}
\label{cifar100_calibration_0.1_val0.895}
\end{center}

\begin{center}
\hspace*{-1cm}
\includegraphics[scale=0.6]{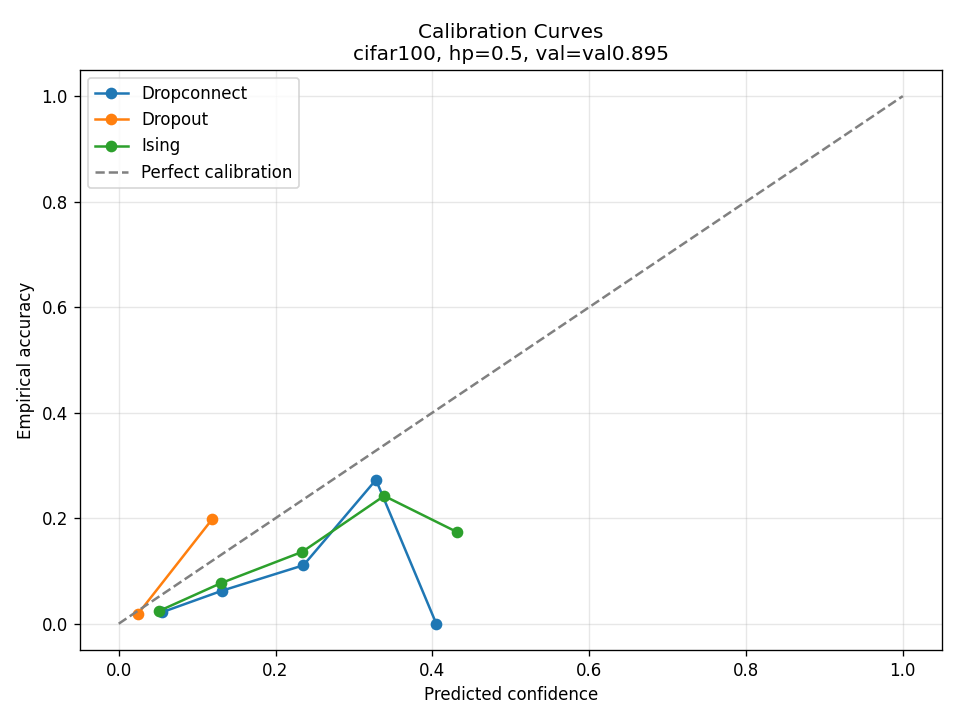}	
\captionof{figure}{Comparison of probability calibration for the three methods examined in this paper. The diagonal line represents perfect calibration; being under the line indicates overconfidence, while being above the line indicates an overly conservative model. The setting for this experiment is 250 training samples, 0.5 regularization hyperparameter, fixed testing dataset size of 5000 from the Cifar100 dataset.}
\label{cifar100_calibration_0.5_val0.895}
\end{center}


\subsection*{S3. Entropy distribution for correct and incorrect classifications}

The entropy distributions among methods are displayed in Figures \ref{mnist_entropy_0.1_val0.6}-\ref{cifar100_entropy_0.5_val0.895}. We display separate plots for correct and incorrect classifications.

\begin{center}
\hspace*{-1cm}
\includegraphics[scale=0.6]{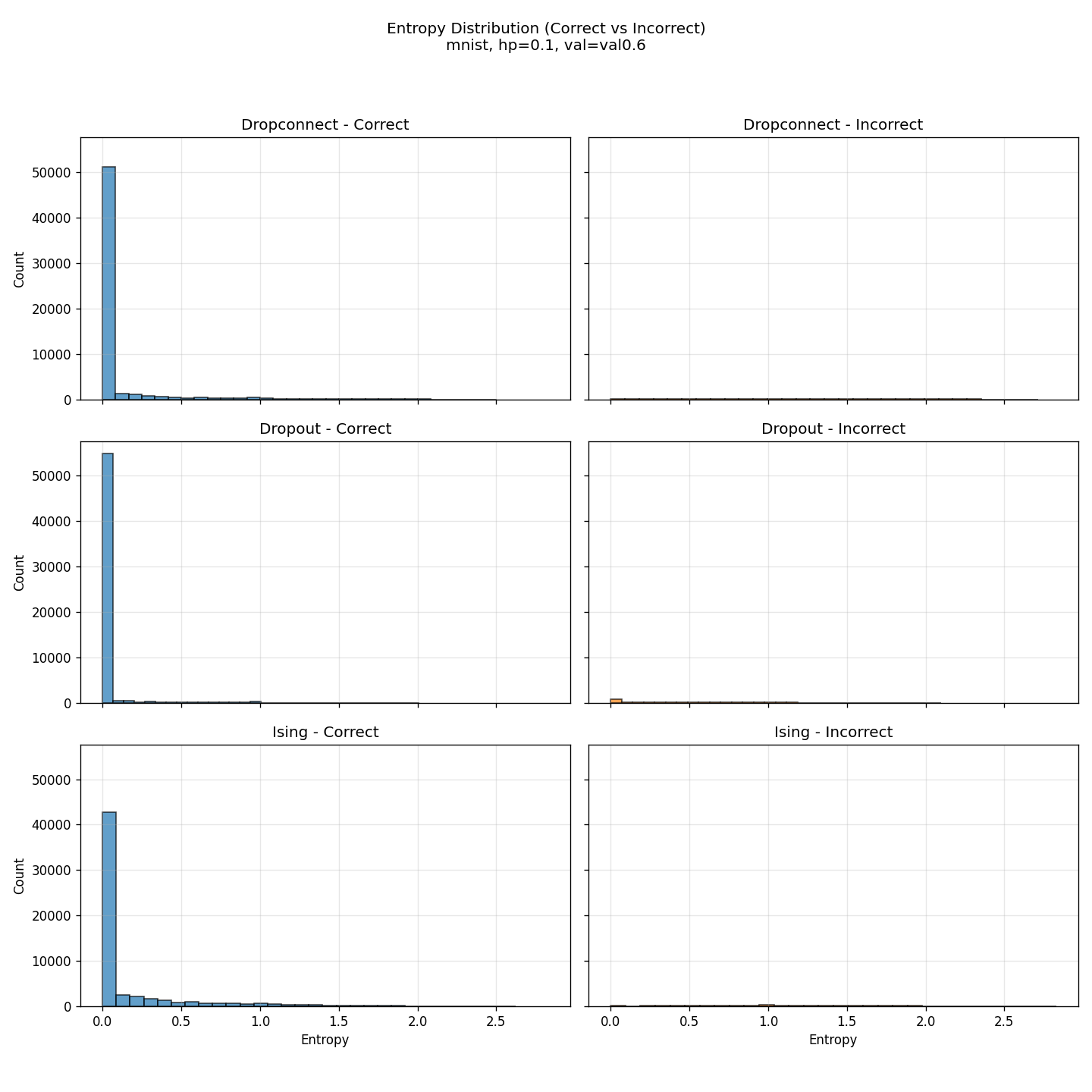}	
\captionof{figure}{Comparison of entropy distribution separated by correct and incorrect classifications for the three methods examined in this paper. The setting for this experiment is 18000 training samples, 0.1 regularization hyperparameter, fixed testing dataset size of 6000 from the MNIST dataset.}
\label{mnist_entropy_0.1_val0.6}
\end{center}

\begin{center}
\hspace*{-1cm}
\includegraphics[scale=0.6]{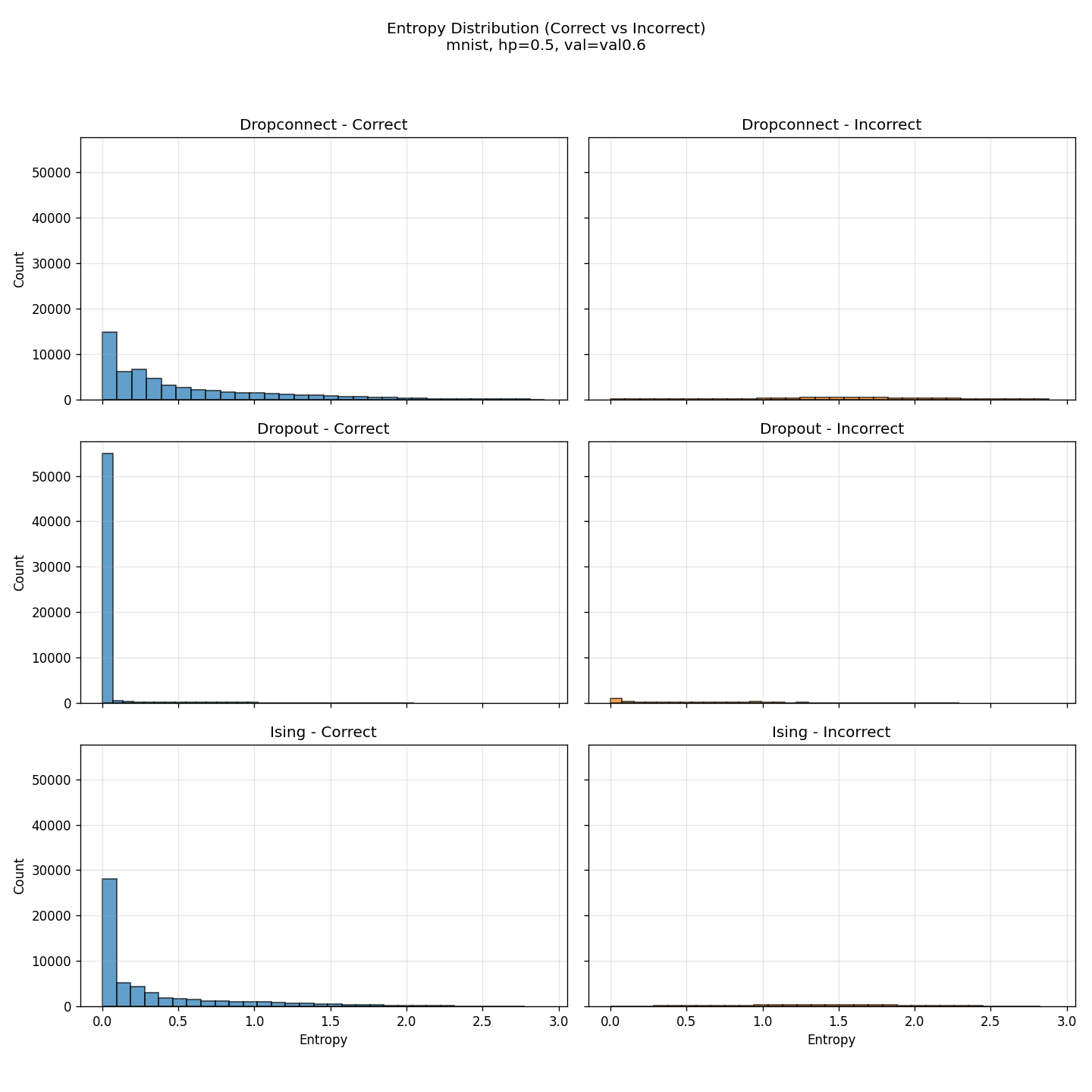}	
\captionof{figure}{Comparison of entropy distribution separated by correct and incorrect classifications for the three methods examined in this paper. The setting for this experiment is 18000 training samples, 0.5 regularization hyperparameter, fixed testing dataset size of 6000 from the MNIST dataset.}
\label{mnist_entropy_0.5_val0.6}
\end{center}

\begin{center}
\hspace*{-1cm}
\includegraphics[scale=0.6]{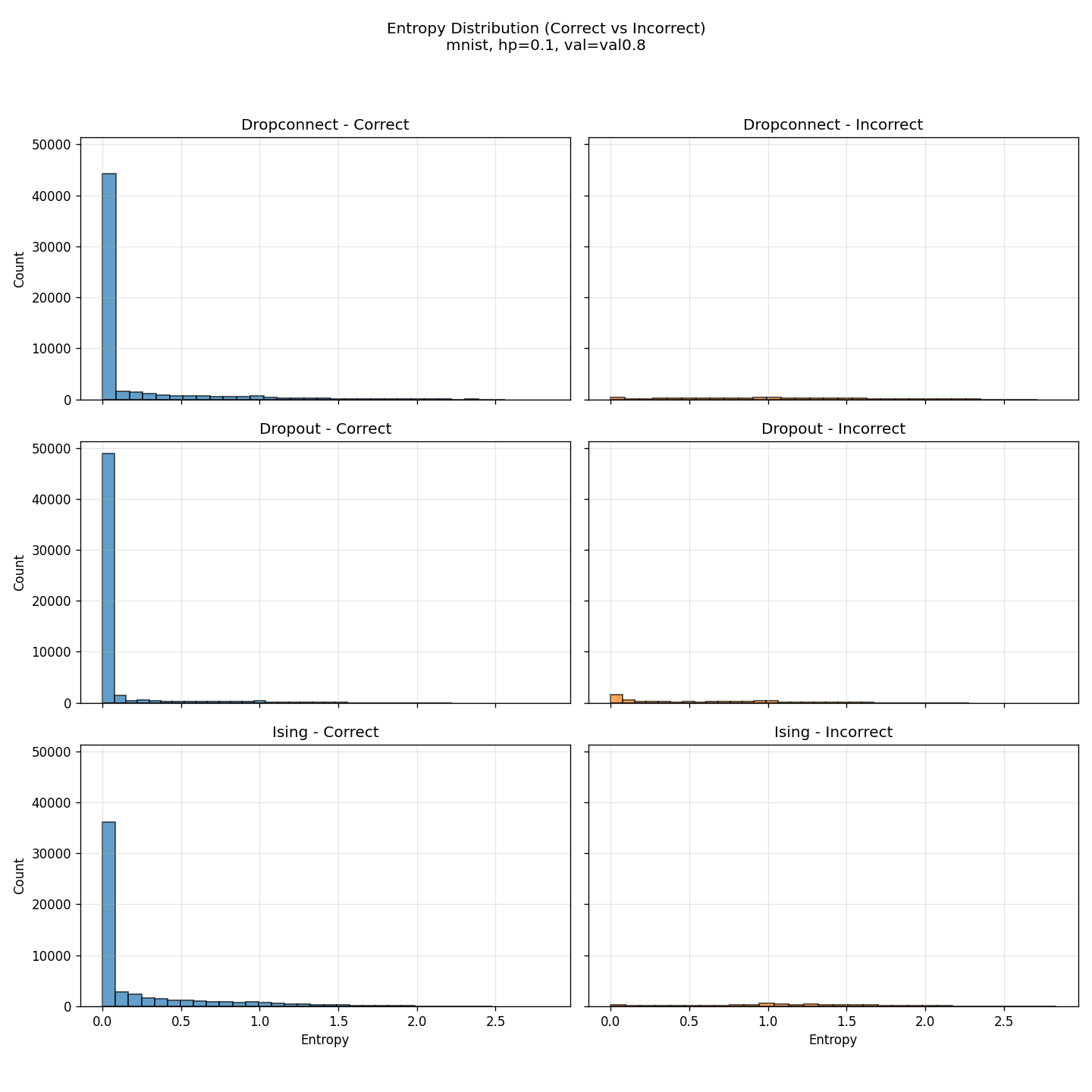}	
\captionof{figure}{Comparison of entropy distribution separated by correct and incorrect classifications for the three methods examined in this paper. The setting for this experiment is 6000 training samples, 0.1 regularization hyperparameter, fixed testing dataset size of 6000 from the MNIST dataset.}
\label{mnist_entropy_0.1_val0.8}
\end{center}

\begin{center}
\hspace*{-1cm}
\includegraphics[scale=0.6]{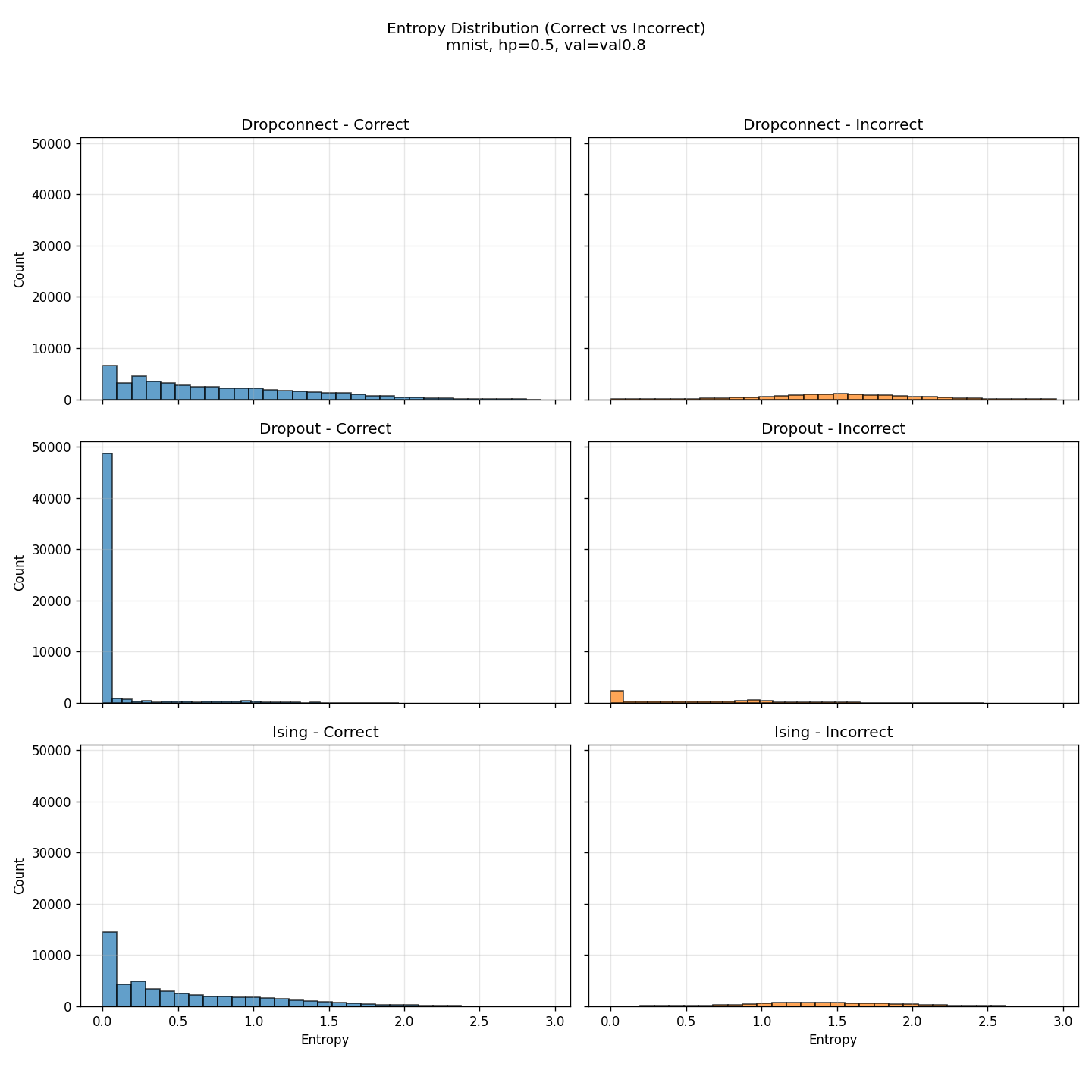}	
\captionof{figure}{Comparison of entropy distribution separated by correct and incorrect classifications for the three methods examined in this paper. The setting for this experiment is 6000 training samples, 0.5 regularization hyperparameter, fixed testing dataset size of 6000 from the MNIST dataset.}
\label{mnist_entropy_0.5_val0.8}
\end{center}

\begin{center}
\hspace*{-1cm}
\includegraphics[scale=0.6]{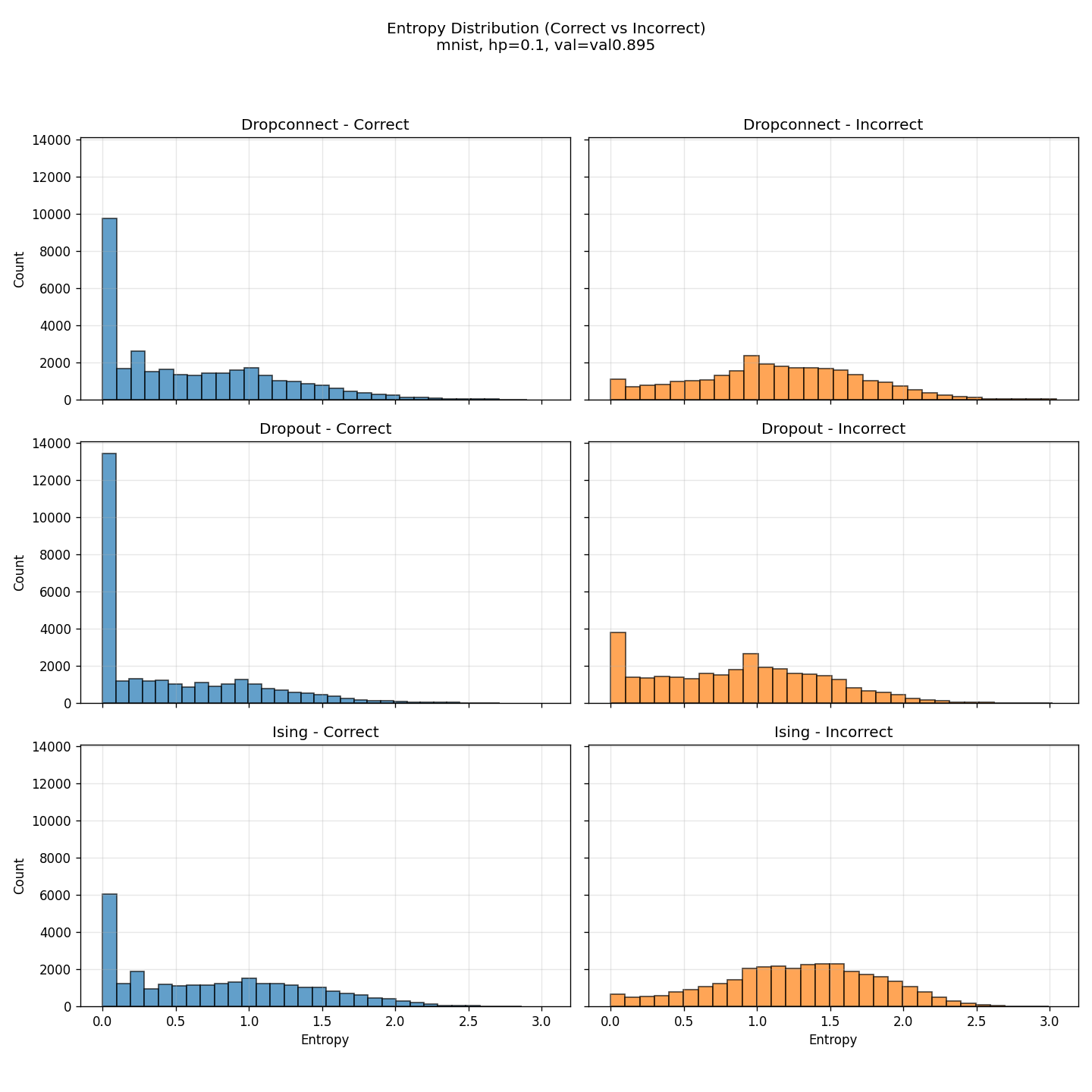}
\captionof{figure}{Comparison of entropy distribution separated by correct and incorrect classifications for the three methods examined in this paper. The setting for this experiment is 300 training samples, 0.1 regularization hyperparameter, fixed testing dataset size of 6000 from the MNIST dataset.}
\label{mnist_entropy_0.1_val0.895}
\end{center}

\begin{center}
\hspace*{-1cm}
\includegraphics[scale=0.6]{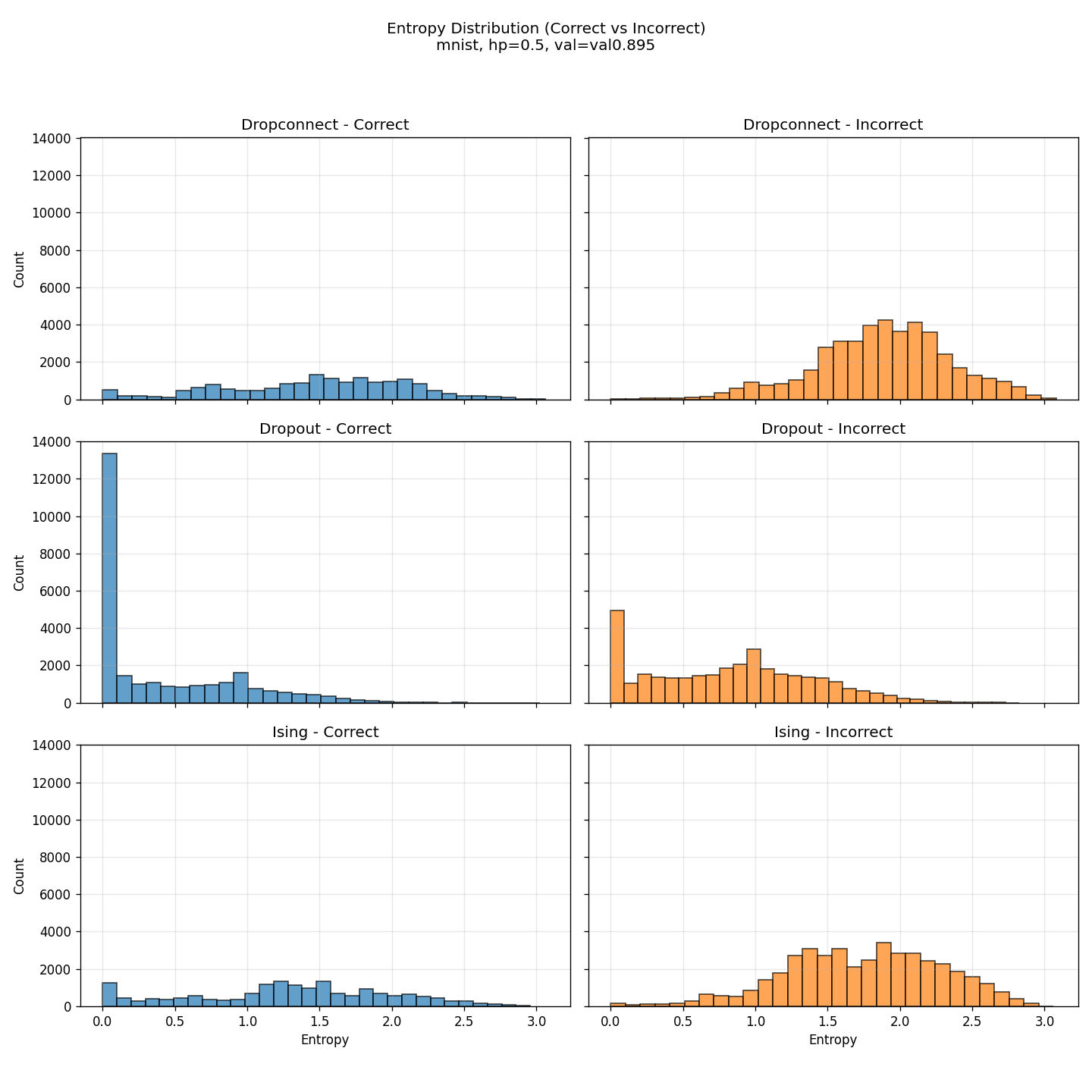}
\captionof{figure}{Comparison of entropy distribution separated by correct and incorrect classifications for the three methods examined in this paper. The setting for this experiment is 300 training samples, 0.5 regularization hyperparameter, fixed testing dataset size of 6000 from the MNIST dataset.}
\label{mnist_entropy_0.5_val0.895}
\end{center}


\begin{center}
\hspace*{-1cm}
\includegraphics[scale=0.6]{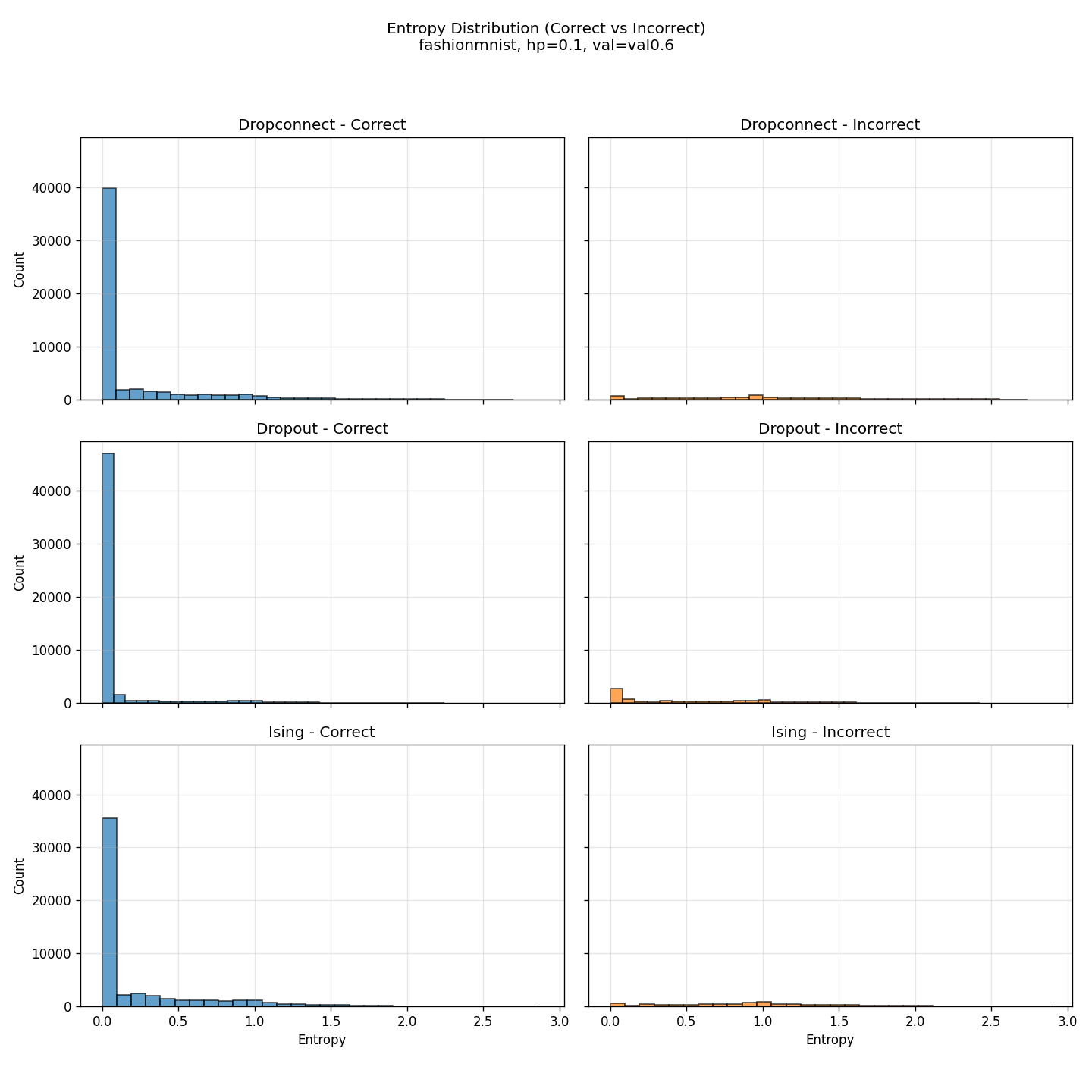}
\captionof{figure}{Comparison of entropy distribution separated by correct and incorrect classifications for the three methods examined in this paper. The setting for this experiment is 18000 training samples, 0.1 regularization hyperparameter, fixed testing dataset size of 6000 from the Fashion-MNIST dataset.}
\label{fashionmnist_entropy_0.1_val0.6}
\end{center}

\begin{center}
\hspace*{-1cm}
\includegraphics[scale=0.6]{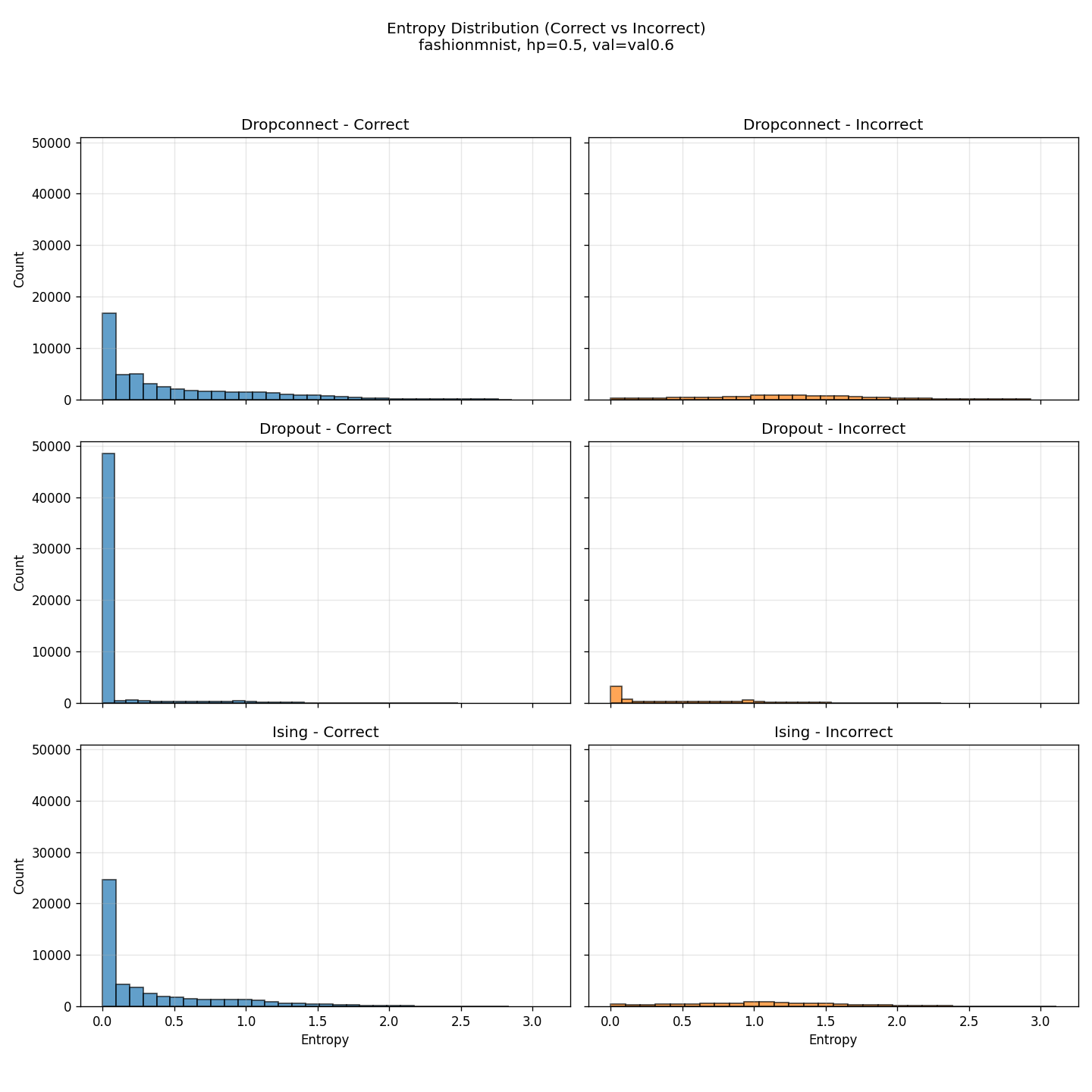}	
\captionof{figure}{Comparison of entropy distribution separated by correct and incorrect classifications for the three methods examined in this paper. The setting for this experiment is 18000 training samples, 0.5 regularization hyperparameter, fixed testing dataset size of 6000 from the Fashion-MNIST dataset.}
\label{fashionmnist_entropy_0.5_val0.6}
\end{center}

\begin{center}
\hspace*{-1cm}
\includegraphics[scale=0.6]{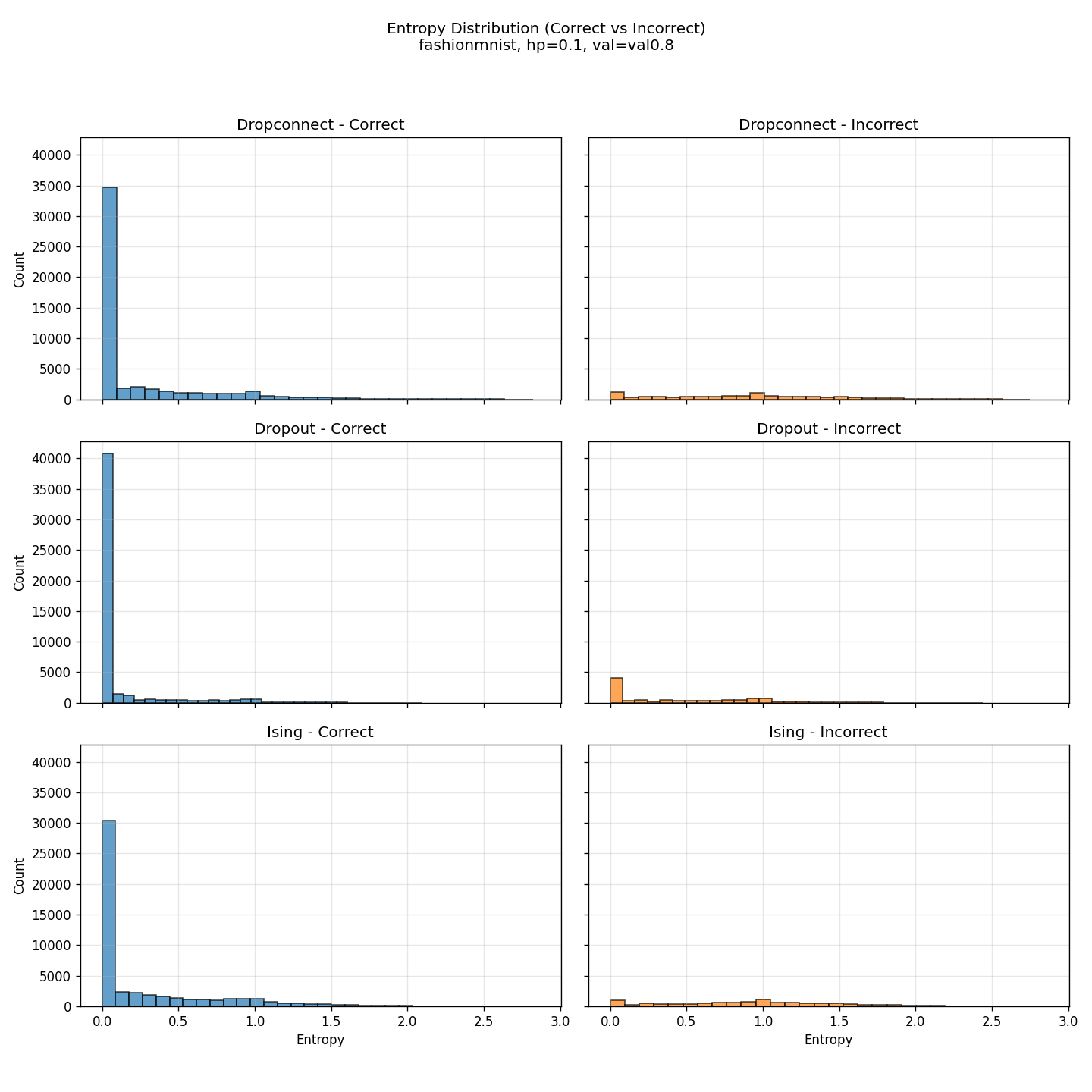}	
\captionof{figure}{Comparison of entropy distribution separated by correct and incorrect classifications for the three methods examined in this paper. The setting for this experiment is 6000 training samples, 0.1 regularization hyperparameter, fixed testing dataset size of 6000 from the Fashion-MNIST dataset.}
\label{fashionmnist_entropy_0.1_val0.8}
\end{center}

\begin{center}
\hspace*{-1cm}
\includegraphics[scale=0.6]{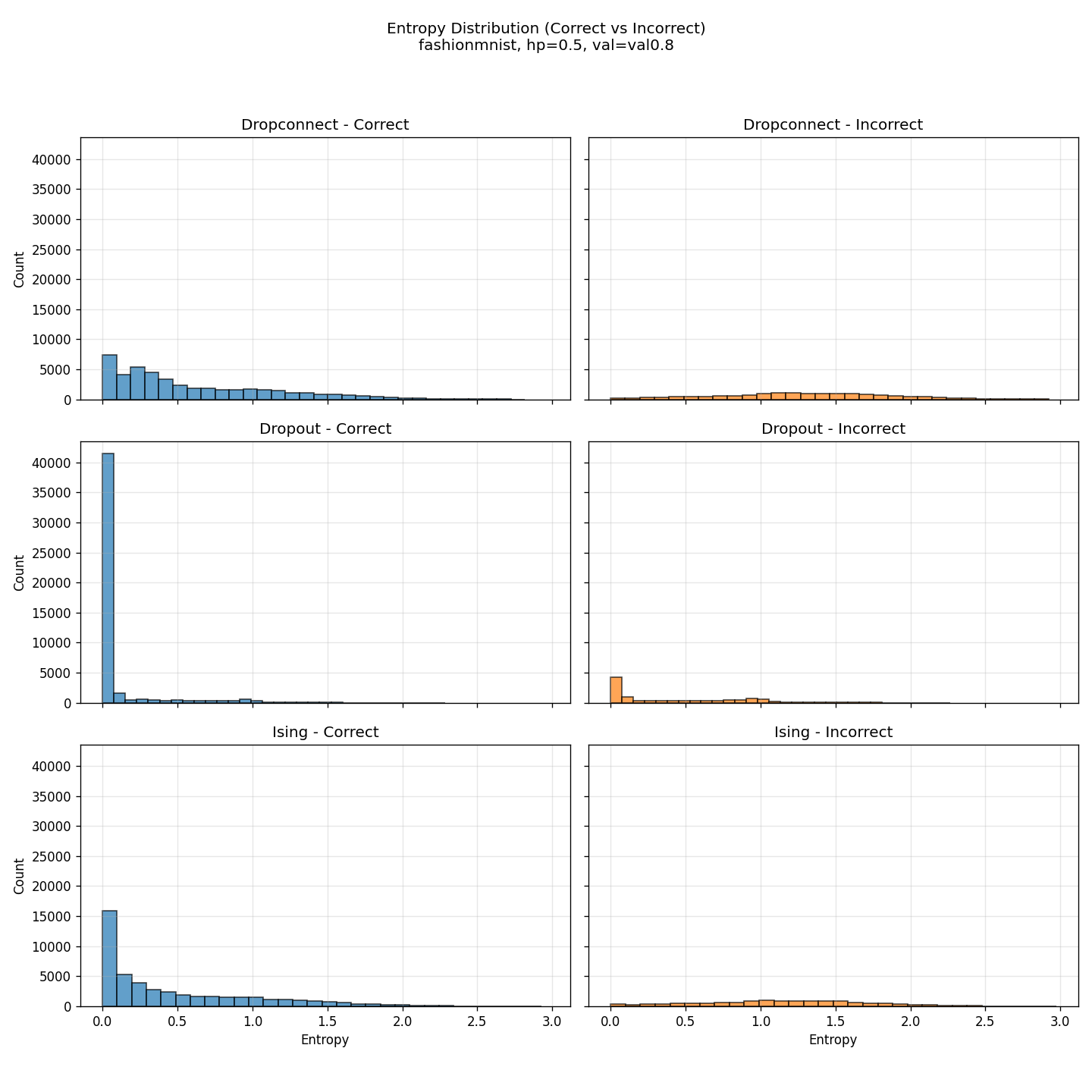}	
\captionof{figure}{Comparison of entropy distribution separated by correct and incorrect classifications for the three methods examined in this paper. The setting for this experiment is 6000 training samples, 0.5 regularization hyperparameter, fixed testing dataset size of 6000 from the Fashion-MNIST dataset.}
\label{fashionmnist_entropy_0.5_val0.8}
\end{center}

\begin{center}
\hspace*{-1cm}
\includegraphics[scale=0.6]{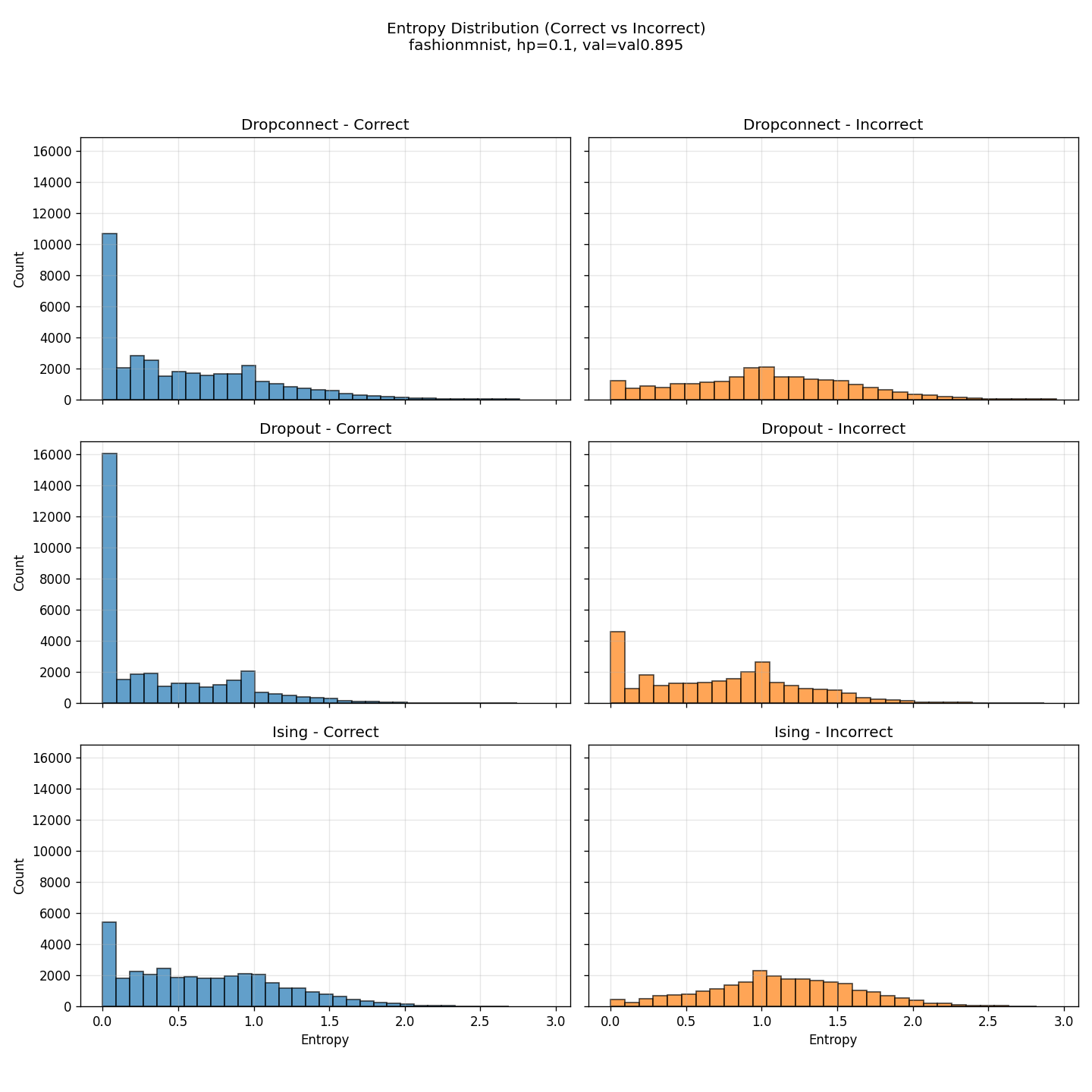}	
\captionof{figure}{Comparison of entropy distribution separated by correct and incorrect classifications for the three methods examined in this paper. The setting for this experiment is 300 training samples, 0.1 regularization hyperparameter, fixed testing dataset size of 6000 from the Fashion-MNIST dataset.}
\label{fashionmnist_entropy_0.1_val0.895}
\end{center}

\begin{center}
\hspace*{-1cm}
\includegraphics[scale=0.6]{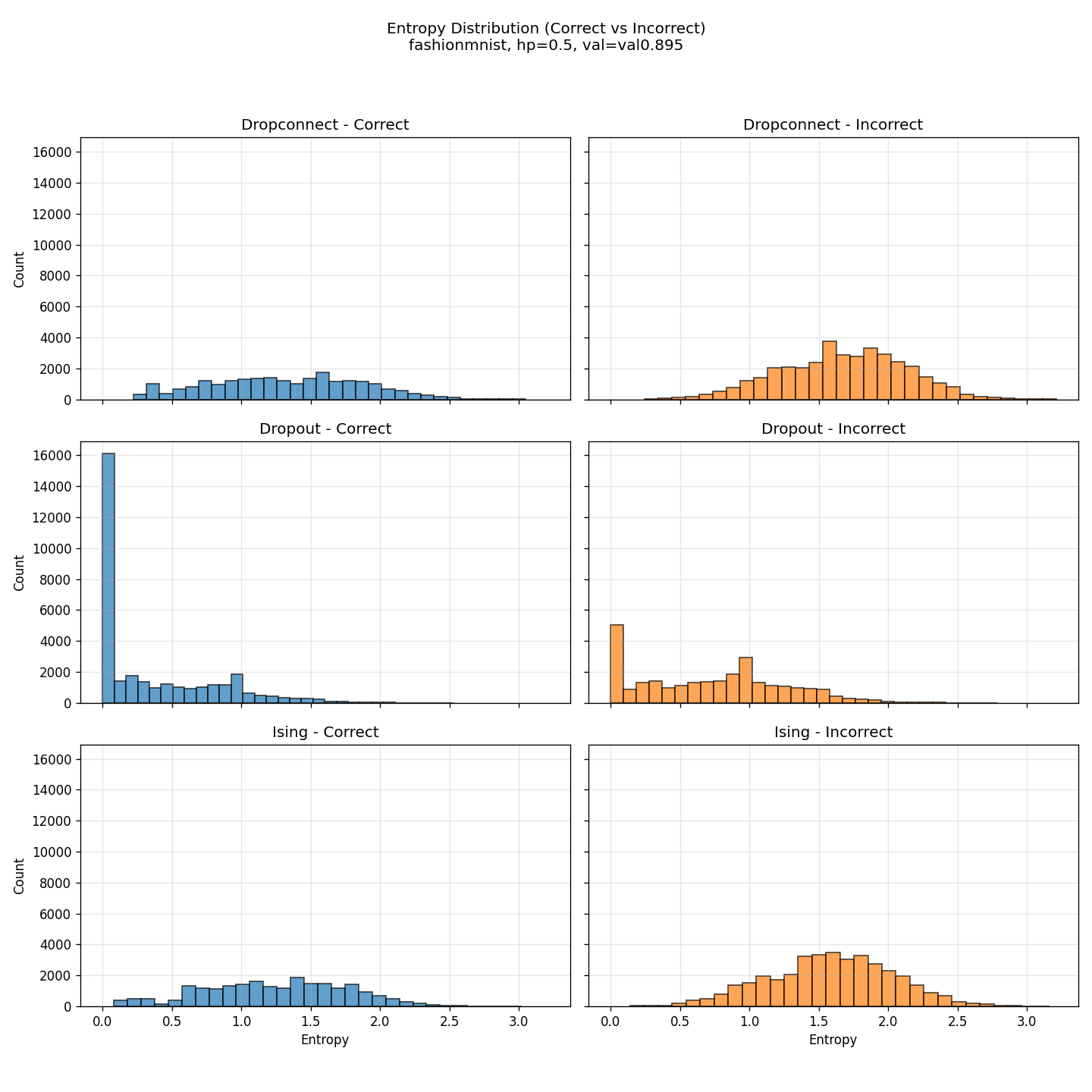}	
\captionof{figure}{Comparison of entropy distribution separated by correct and incorrect classifications for the three methods examined in this paper. The setting for this experiment is 300 training samples, 0.5 regularization hyperparameter, fixed testing dataset size of 6000 from the Fashion-MNIST dataset.}
\label{fashionmnist_entropy_0.5_val0.895}
\end{center}


\begin{center}
\hspace*{-1cm}
\includegraphics[scale=0.6]{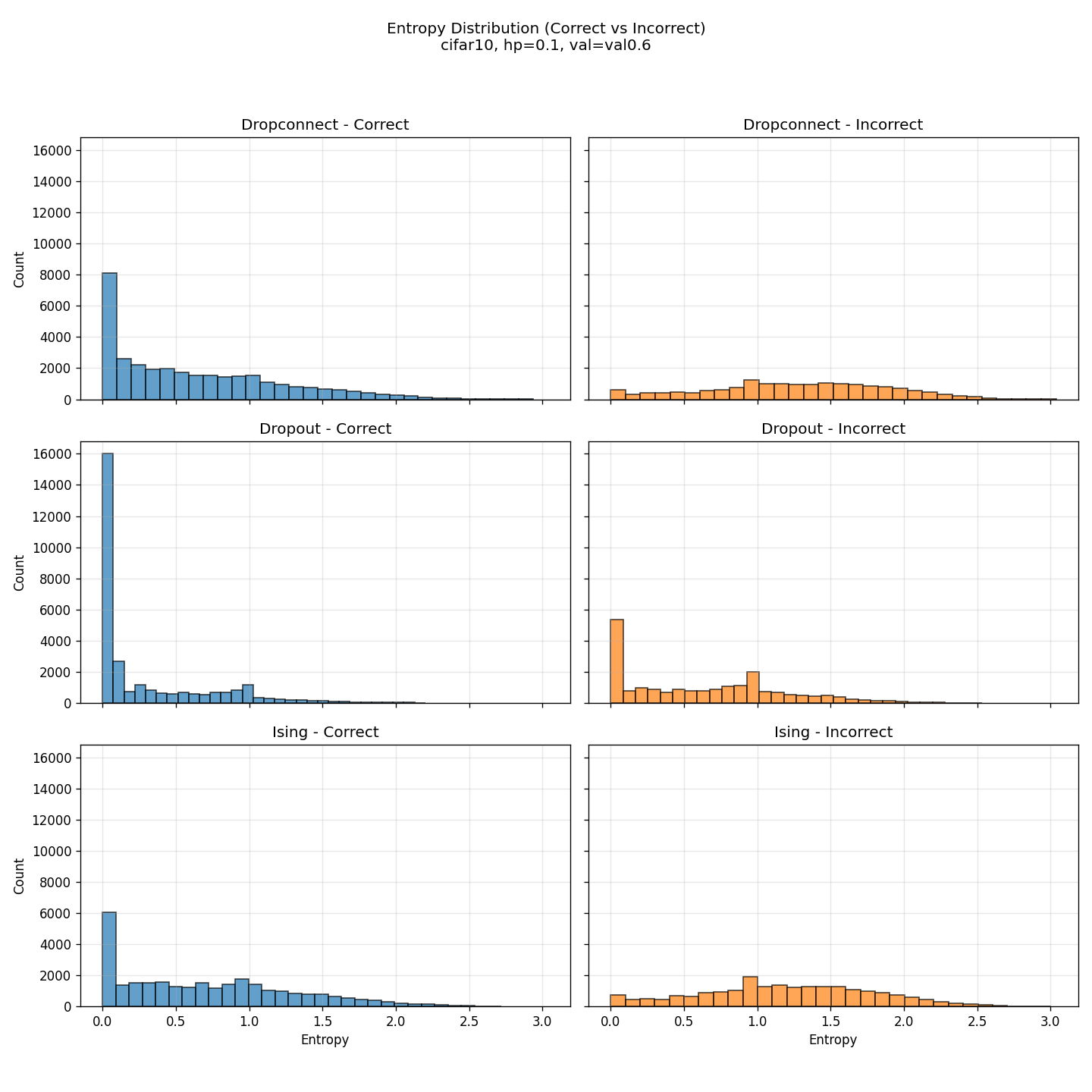}	
\captionof{figure}{Comparison of entropy distribution separated by correct and incorrect classifications for the three methods examined in this paper. The setting for this experiment is 15000 training samples, 0.1 regularization hyperparameter, fixed testing dataset size of 6000 from the Cifar10 dataset.}
\label{cifar10_entropy_0.1_val0.6}
\end{center}

\begin{center}
\hspace*{-1cm}
\includegraphics[scale=0.6]{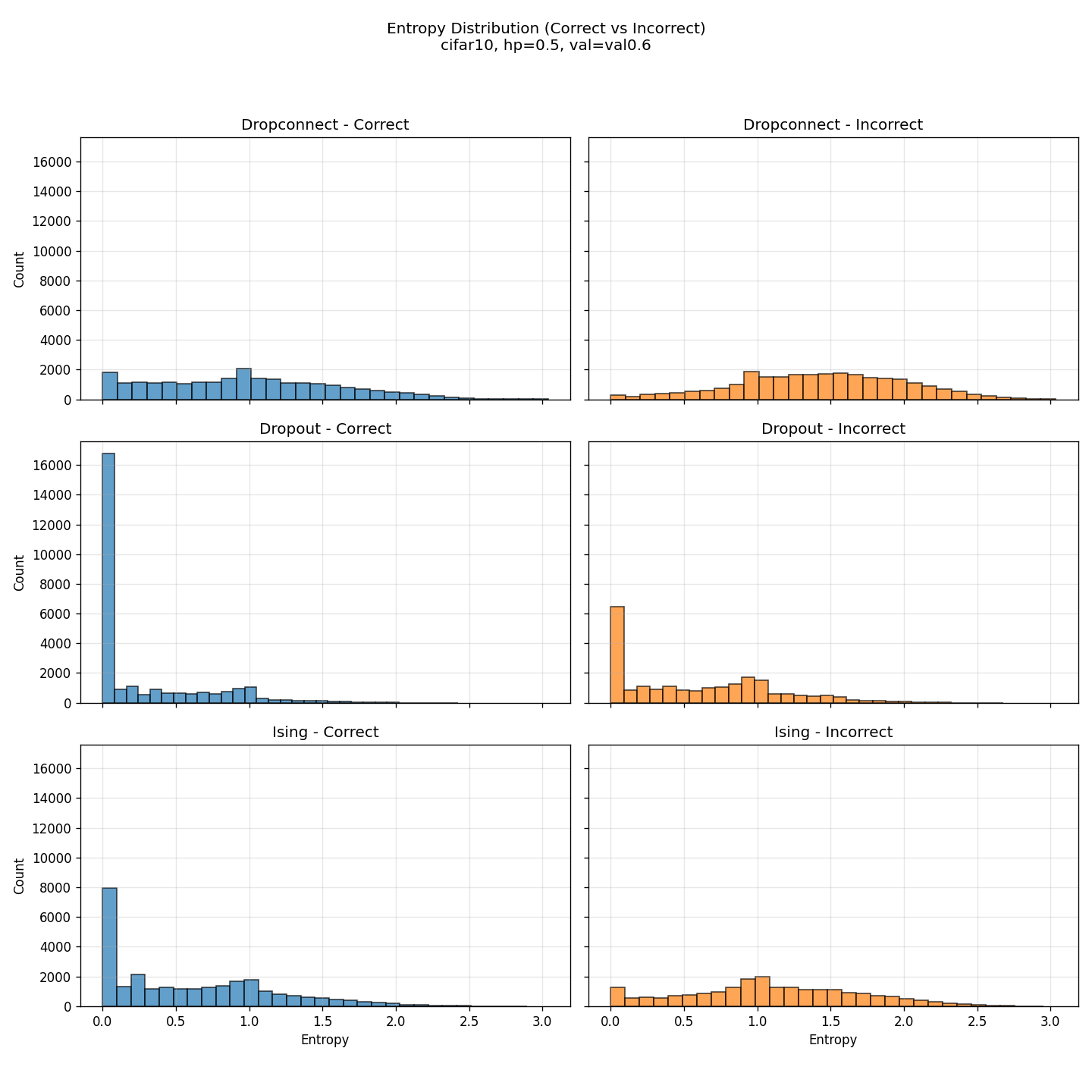}	
\captionof{figure}{Comparison of entropy distribution separated by correct and incorrect classifications for the three methods examined in this paper. The setting for this experiment is 15000 training samples, 0.5 regularization hyperparameter, fixed testing dataset size of 6000 from the Cifar10 dataset.}
\label{cifar10_entropy_0.5_val0.6}
\end{center}

\begin{center}
\hspace*{-1cm}
\includegraphics[scale=0.6]{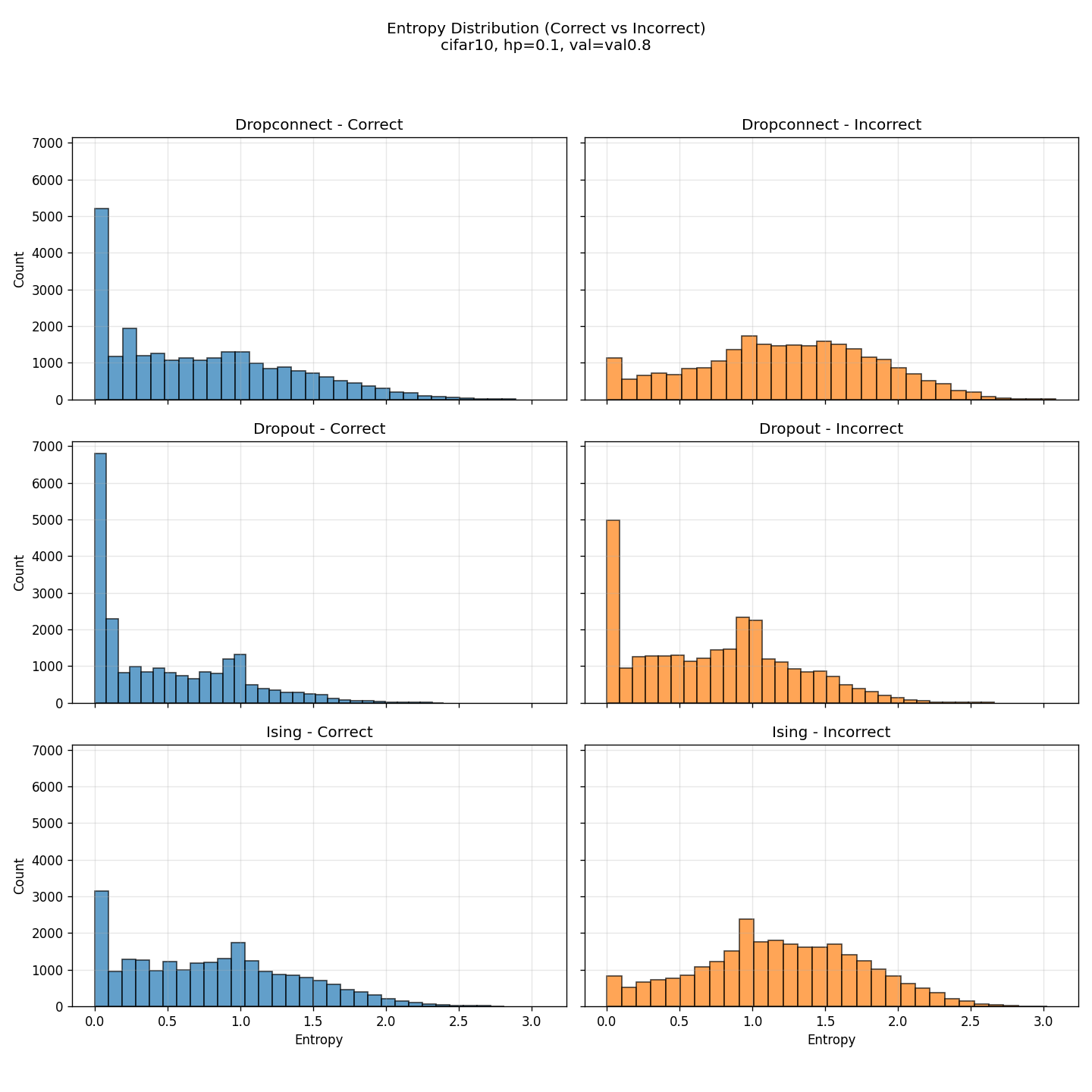}
\captionof{figure}{Comparison of entropy distribution separated by correct and incorrect classifications for the three methods examined in this paper. The setting for this experiment is 5000 training samples, 0.1 regularization hyperparameter, fixed testing dataset size of 5000 from the Cifar10 dataset.}
\label{cifar10_entropy_0.1_val0.8}
\end{center}
	
\begin{center}
\hspace*{-1cm}
\includegraphics[scale=0.6]{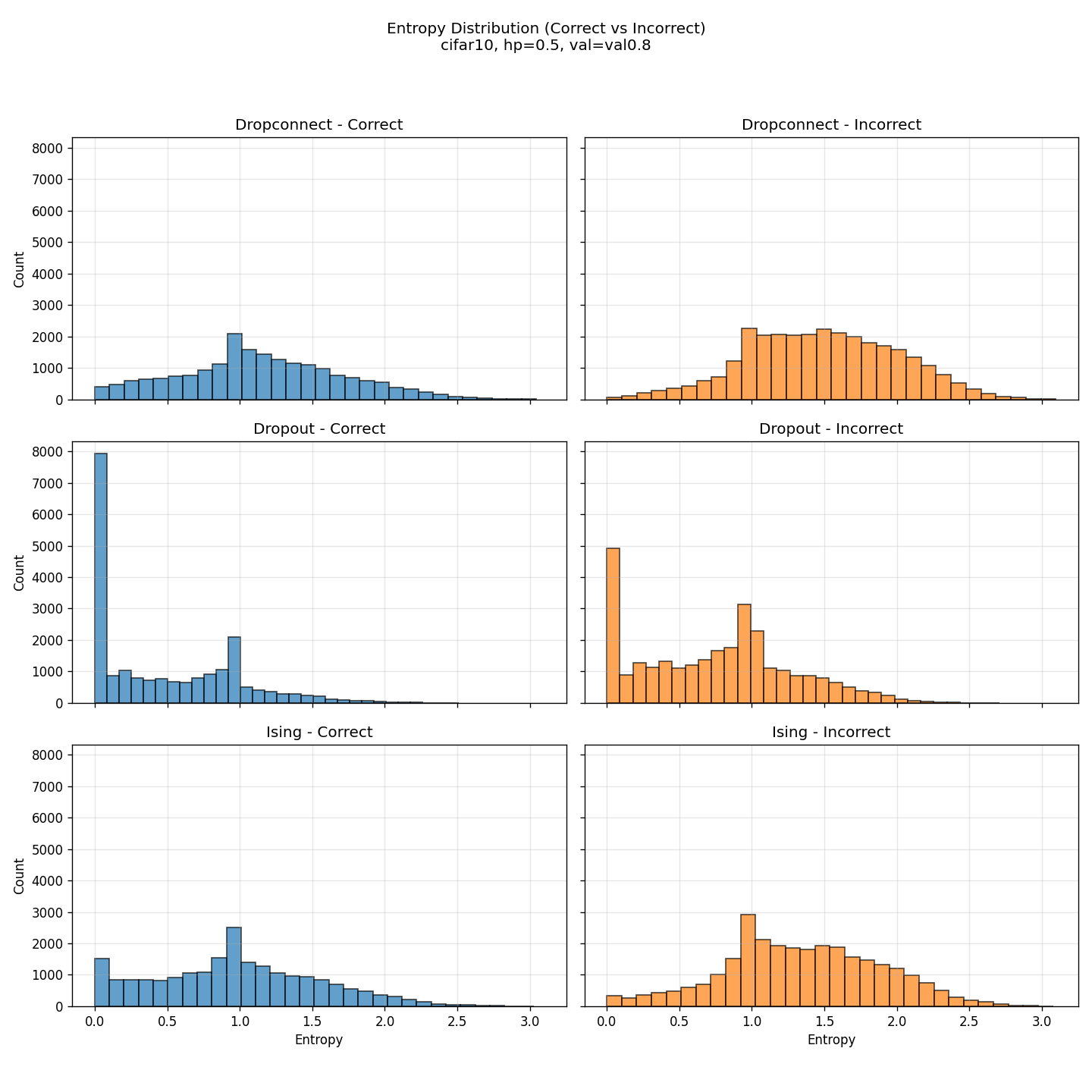}
\captionof{figure}{Comparison of entropy distribution separated by correct and incorrect classifications for the three methods examined in this paper. The setting for this experiment is 5000 training samples, 0.5 regularization hyperparameter, fixed testing dataset size of 5000 from the Cifar10 dataset.}
\label{cifar10_entropy_0.5_val0.8}
\end{center}

\begin{center}
\hspace*{-1cm}
\includegraphics[scale=0.6]{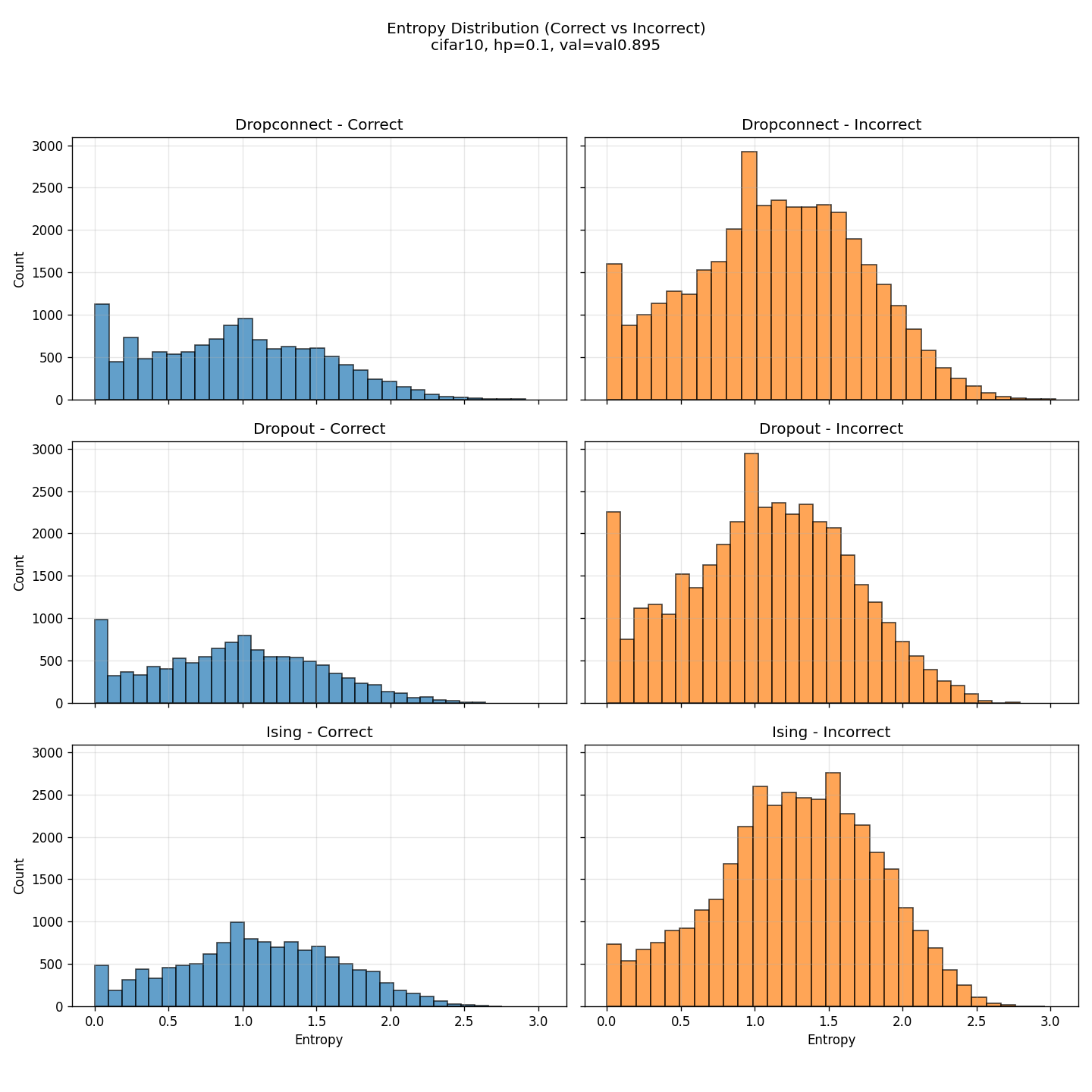}	
\captionof{figure}{Comparison of entropy distribution separated by correct and incorrect classifications for the three methods examined in this paper. The setting for this experiment is 250 training samples, 0.1 regularization hyperparameter, fixed testing dataset size of 5000 from the Cifar10 dataset.}
\label{cifar10_entropy_0.1_val0.895}
\end{center}

\begin{center}
\hspace*{-1cm}
\includegraphics[scale=0.6]{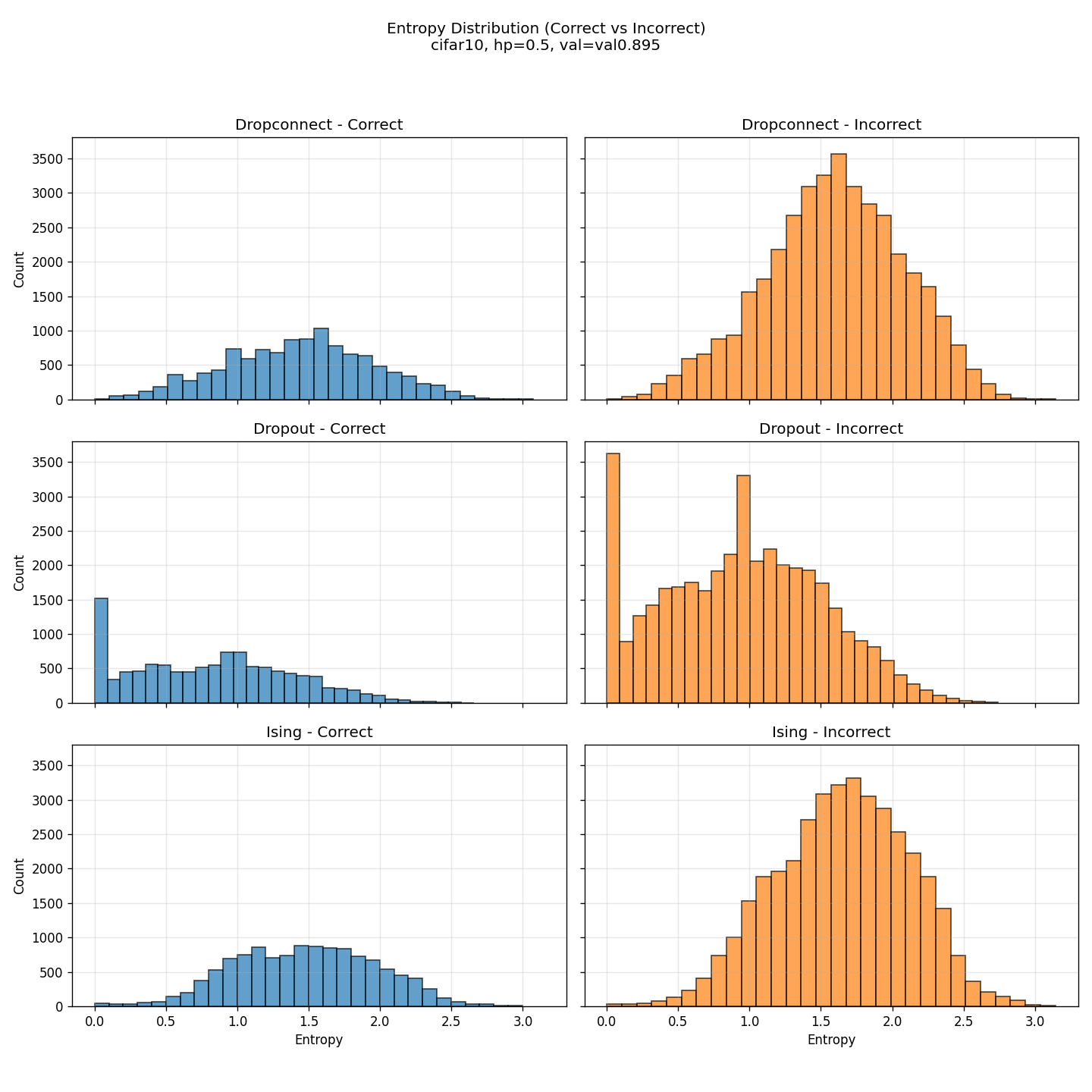}	
\captionof{figure}{Comparison of entropy distribution separated by correct and incorrect classifications for the three methods examined in this paper. The setting for this experiment is 250 training samples, 0.5 regularization hyperparameter, fixed testing dataset size of 5000 from the Cifar10 dataset.}
\label{cifar10_entropy_0.5_val0.895}
\end{center}


\begin{center}
\hspace*{-1cm}
\includegraphics[scale=0.6]{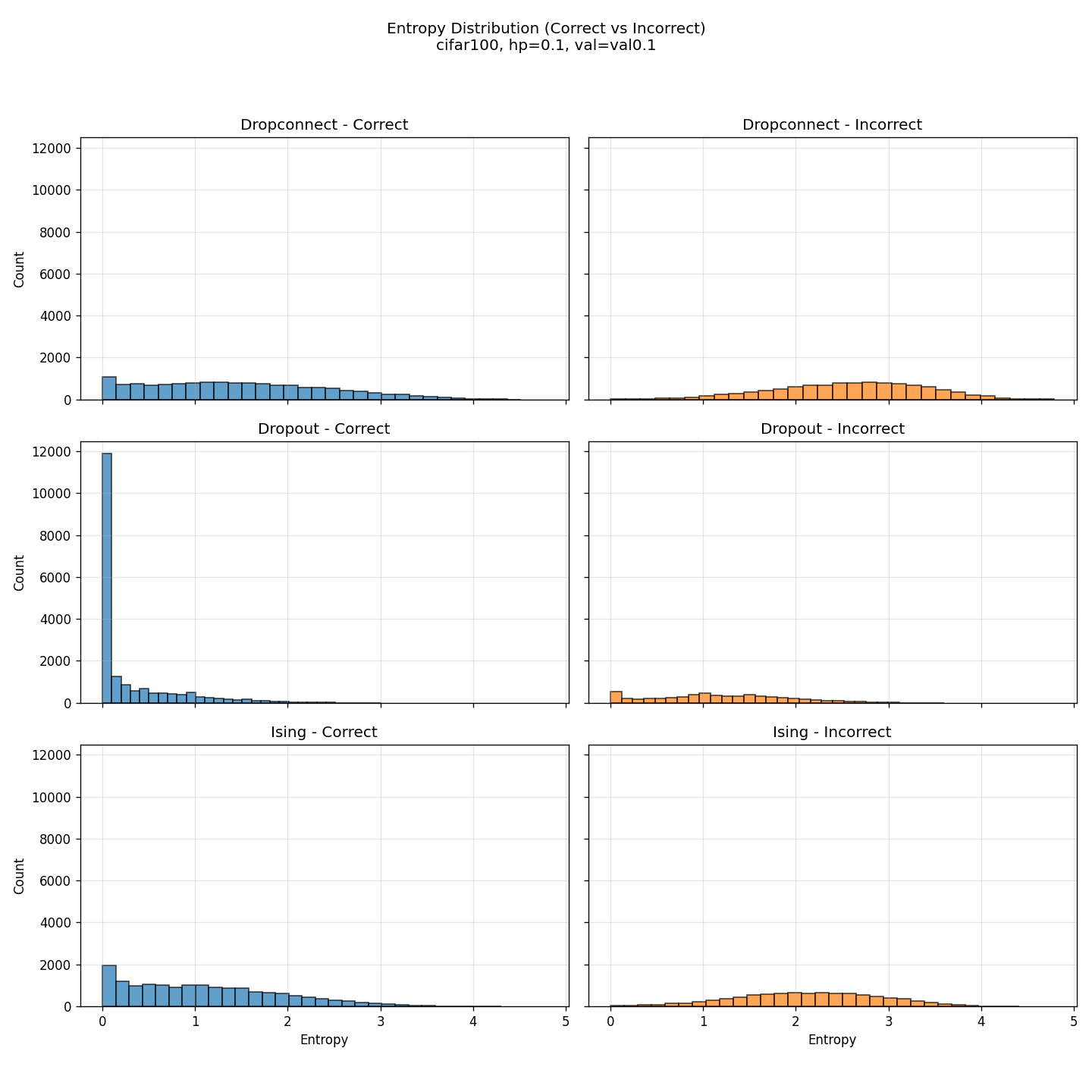}
\captionof{figure}{Comparison of entropy distribution separated by correct and incorrect classifications for the three methods examined in this paper. The setting for this experiment is 40000 training samples, 0.1 regularization hyperparameter, fixed testing dataset size of 5000 from the Cifar100 dataset.}
\label{cifar100_entropy_0.1_val0.1}
\end{center}

\begin{center}
\hspace*{-1cm}
\includegraphics[scale=0.6]{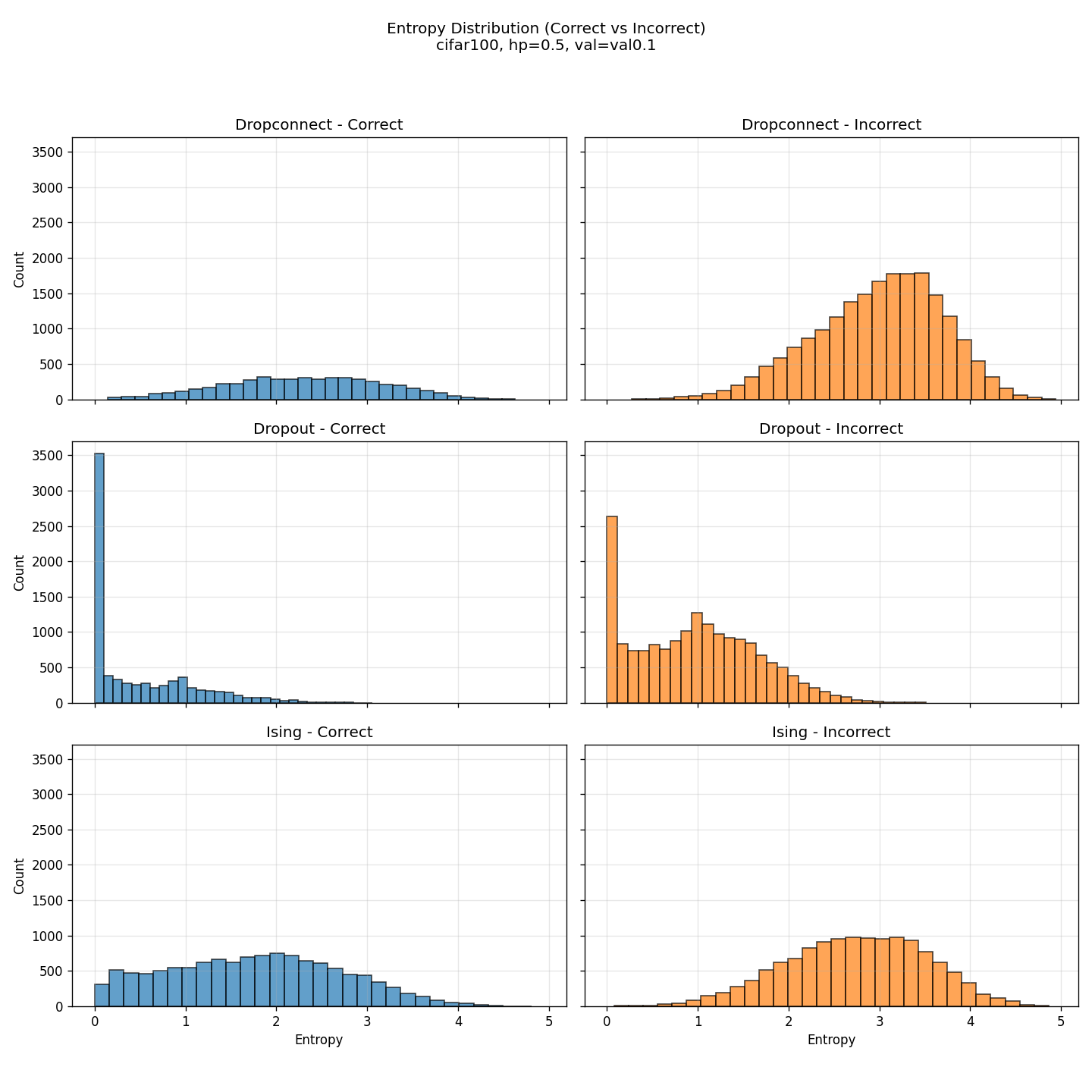}	
\captionof{figure}{Comparison of entropy distribution separated by correct and incorrect classifications for the three methods examined in this paper. The setting for this experiment is 40000 training samples, 0.5 regularization hyperparameter, fixed testing dataset size of 5000 from the Cifar100 dataset.}
\label{cifar100_entropy_0.5_val0.1}
\end{center}

\begin{center}
\hspace*{-1cm}
\includegraphics[scale=0.6]{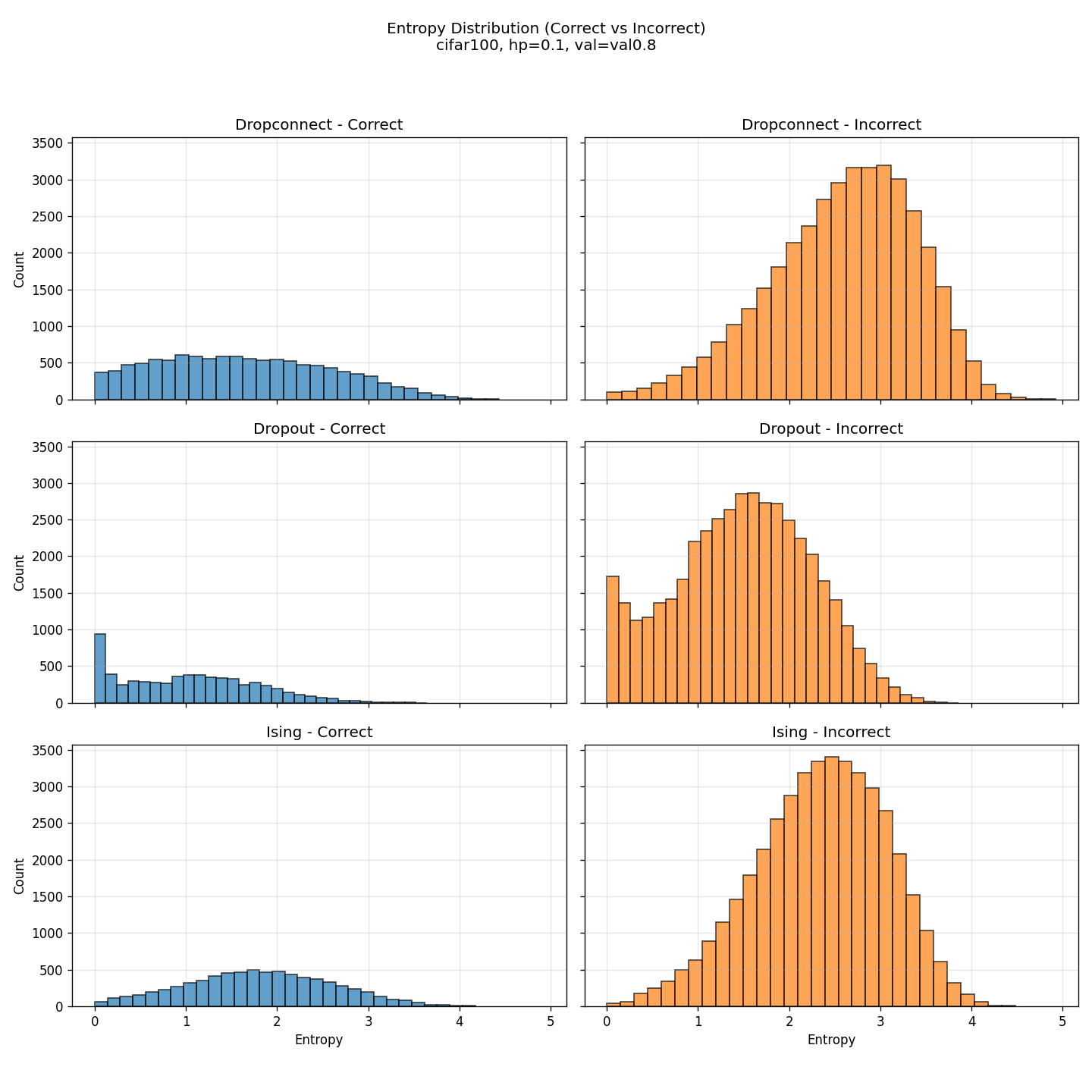}	
\captionof{figure}{Comparison of entropy distribution separated by correct and incorrect classifications for the three methods examined in this paper. The setting for this experiment is 5000 training samples, 0.1 regularization hyperparameter, fixed testing dataset size of 5000 from the Cifar100 dataset.}
\label{cifar100_entropy_0.1_val0.8}
\end{center}

\begin{center}
\hspace*{-1cm}
\includegraphics[scale=0.6]{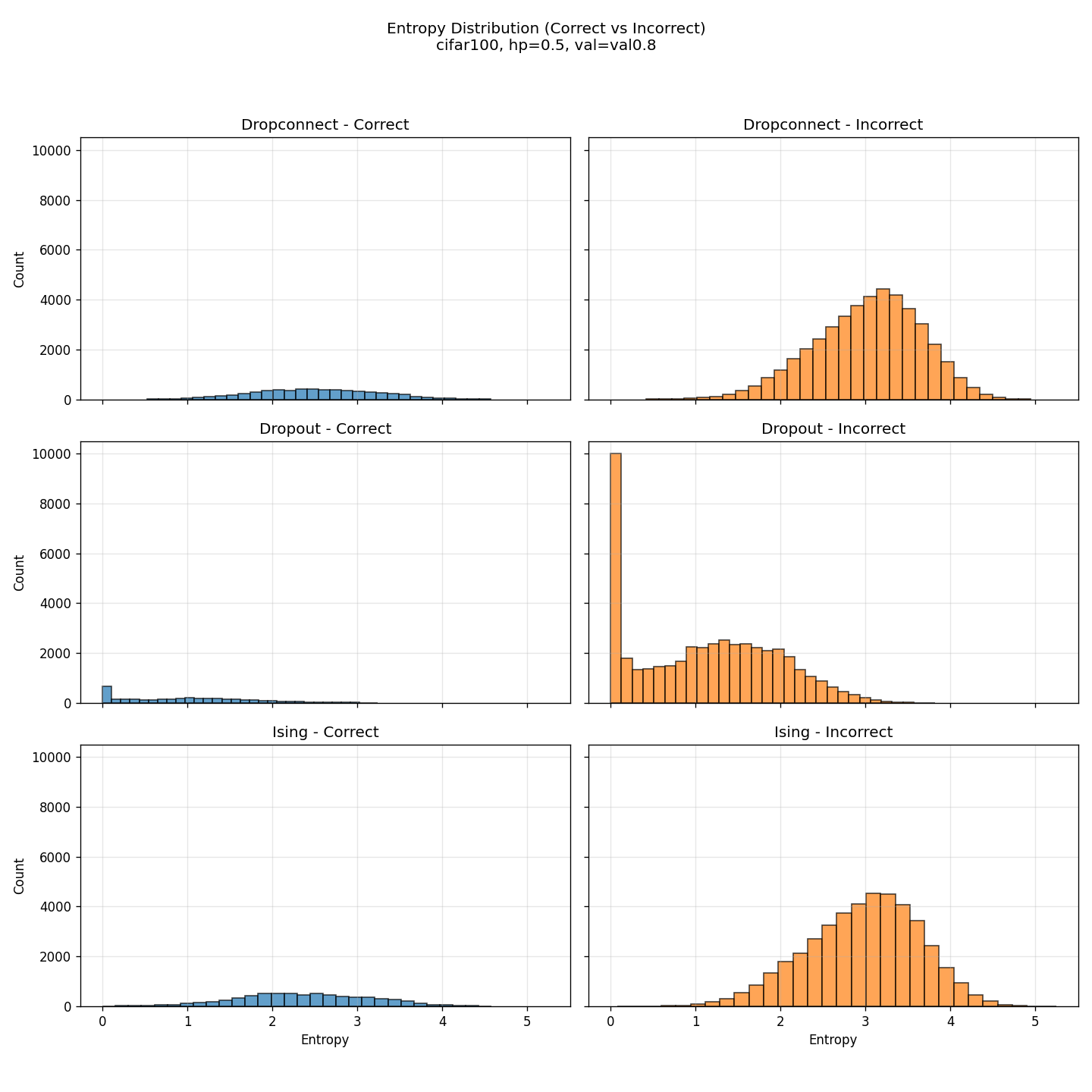}	
\captionof{figure}{Comparison of entropy distribution separated by correct and incorrect classifications for the three methods examined in this paper. The setting for this experiment is 5000 training samples, 0.5 regularization hyperparameter, fixed testing dataset size of 5000 from the Cifar100 dataset.}
\label{cifar100_entropy_0.5_val0.8}
\end{center}

\begin{center}
\hspace*{-1cm}
\includegraphics[scale=0.6]{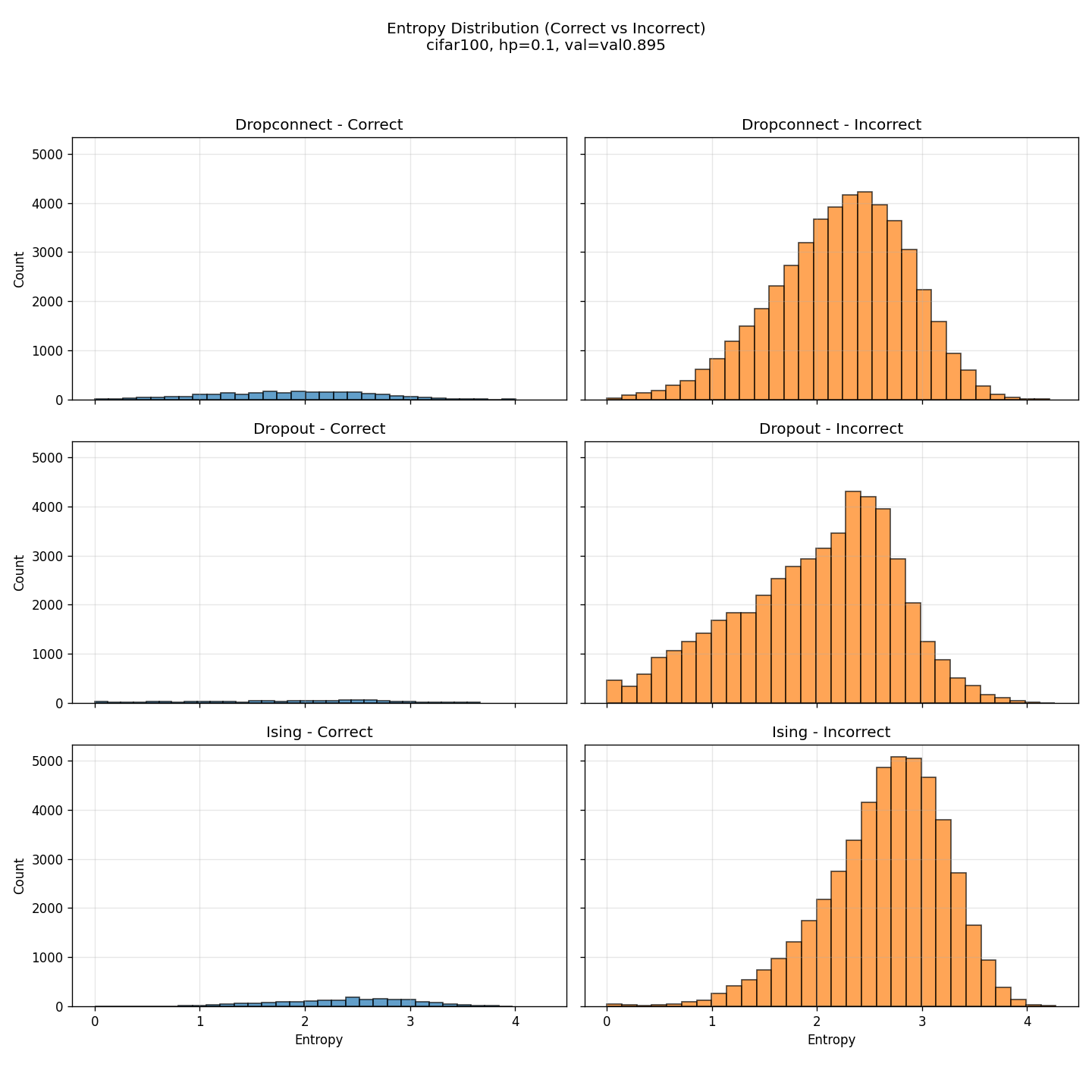}	
\captionof{figure}{Comparison of entropy distribution separated by correct and incorrect classifications for the three methods examined in this paper. The setting for this experiment is 250 training samples, 0.1 regularization hyperparameter, fixed testing dataset size of 5000 from the Cifar100 dataset.}
\label{cifar100_entropy_0.1_val0.895}
\end{center}

\begin{center}
\hspace*{-1cm}
\includegraphics[scale=0.6]{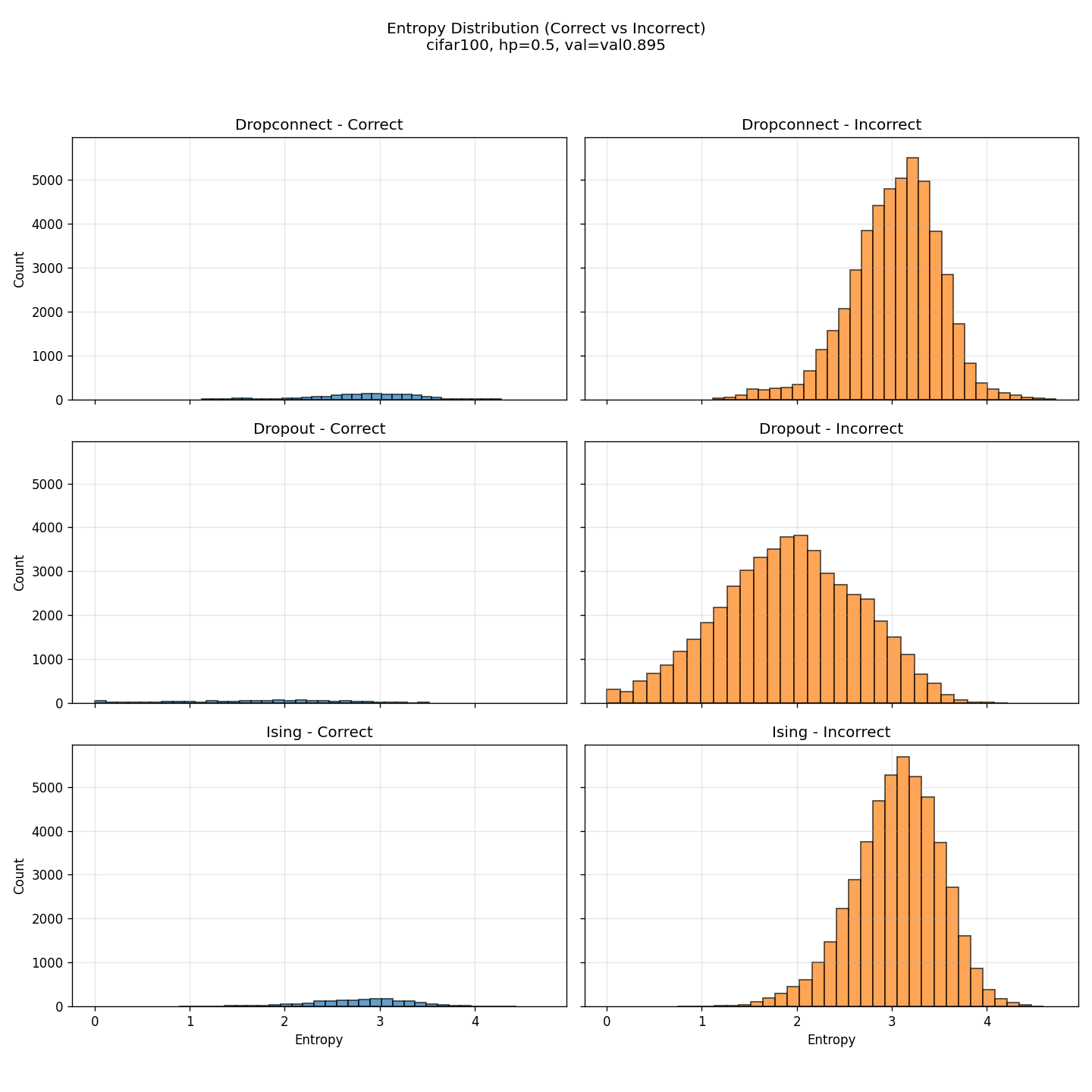}	
\captionof{figure}{Comparison of entropy distribution separated by correct and incorrect classifications for the three methods examined in this paper. The setting for this experiment is 250 training samples, 0.5 regularization hyperparameter, fixed testing dataset size of 5000 from the Cifar100 dataset.}
\label{cifar100_entropy_0.5_val0.895}
\end{center}

\mbox{}


\end{document}